%% file: main.tex
\documentclass{article}

\usepackage[preprint, nonatbib]{neurips_2024}

\usepackage[utf8]{inputenc} 
\usepackage[T1]{fontenc}    
\usepackage{hyperref}       
\usepackage{url}            
\usepackage{booktabs}       

\usepackage{amsmath} 
\usepackage{amsfonts}       
\usepackage{nicefrac}       
\usepackage{microtype}      
\usepackage{xcolor, colortbl}         
\usepackage{graphicx}
\usepackage[font=footnotesize, labelformat=simple]{subcaption}

\usepackage{amssymb}
\usepackage{tabularx}
\usepackage{booktabs, multirow}
\usepackage{changepage,threeparttable}
\usepackage{tabularray}
\usepackage[pro]{fontawesome5}
\usepackage{bm}
\usepackage{bbding}
\usepackage{wrapfig}
\usepackage[font=small]{caption}
\usepackage[export]{adjustbox}

\usepackage[capitalize]{cleveref}
\crefname{section}{Sec.}{Secs.}
\Crefname{section}{Section}{Sections}
\Crefname{table}{Table}{Tables}
\crefname{table}{Tab.}{Tabs.}

\newcommand{\ToggleComments}[1]{} 

\newcommand{\hb}[1]{\ToggleComments{\textcolor{blue}{[Hakan: #1]}}}

\usepackage{xspace}
\makeatletter
\DeclareRobustCommand\onedot{\futurelet\@let@token\@onedot}
\def\@onedot{\ifx\@let@token.\else.\null\fi\xspace}
 
\def\ie{\emph{i.e}\onedot,\xspace}

\makeatother

\title{Articulate your NeRF: Unsupervised articulated object modeling via conditional view synthesis}

\author{%
Jianning Deng \quad Kartic Subr \quad Hakan Bilen \\
University of Edinburgh\\
\texttt{\{jianning.deng, K.Subr, hbilen\}@ed.ac.uk}
}

\begin{document}

\maketitle

\begin{abstract}
    We propose a novel unsupervised method to learn pose and part-segmentation of articulated objects with rigid parts.
    Given two observations of an object in different articulation states, our method learns the geometry and appearance of object parts by using an implicit model from the first observation, distills the part segmentation and articulation from the second observation while rendering the latter observation.
    Additionally, to tackle the complexities in the joint optimization of part segmentation and articulation, we propose a voxel grid based initialization strategy and a decoupled optimization procedure.
    Compared to the prior unsupervised work, our model obtains significantly better performance, generalizes to objects with multiple parts while it can be efficiently from few views for the latter observation.
\end{abstract}

\input{sections/01introduction}
\input{sections/02relwork}
\input{sections/03method}
\input{sections/04experiments}
\input{sections/05conclusion}

\bibliographystyle{plain}
\bibliography{reference}

\newpage
\input{sections/08supp}

\end{document}

%% file: sections/01introduction.tex
\section{Introduction}

Articulated objects, composed of multiple rigid parts connected by joints allowing rotational or translational motion, such as doors, cupboards and spectacles are ubiquitous in our daily lives. 
Automatically understanding the shape, structure and motion of these objects is crucial for numerous applications in robotic manipulation \cite{katz2008manipulating,Xiang_2020_SAPIEN} and character animation \cite{badler1990making,siarohin2021motion}.
Many works \cite{jiang2022opd, qian2022understanding, sun2023opdmulti} focused on this problem use groundtruth 3D shape, articulation information, and/or part segmentation to learn articulated object models but acquiring accurate 3D observations and manual annotations is typically complex and too expensive for building real large-scale datasets. 

In this paper, we introduce a novel unsupervised technique that learns part segmentation and articulation (\ie axis of movement, and translation/rotation of each movable part) from two sets of observations without requiring groundtruth shape, part segmentation or articulation.
Each set contains images of an object from multiple viewpoints in different articulations.
Our key idea is that articulation changes only the poses of the object parts, not their  geometry or texture.
Hence, once the geometry and appearance are learned, one can transform to another articulation state given the part locations and target articulation parameters.

Building on this idea, we frame the learning problem as a conditional novel articulation (and view) synthesis task (see~\cref{fig:overview}(a)).
Given a source observation with multiple views of an object in one articulation state, we first learn the object's shape and appearance by using an implicit model~\cite{mildenhall2021nerf} and freeze its weights.
Next we pass the target observation -- multi-view images of the same object in a different articulation state -- to a tight bottleneck that distills part locations and articulations.
We constrain our model to assign each 3D coordinate that is occupied by the object to a part and to apply a valid geometric transformation to the 3D coordinates of each part through ray geometry.
The predictions of part segmentation and articulation, along with the target camera viewpoint, are passed to the implicit function and its differential renderer to reproduce the target observations (see~\cref{fig:overview}(b)).
Minimizing the photometric error between the rendered and target view provides supervision for learning part segmentation and articulation.
However, joint optimization of these intertwined tasks are challenging and very sensitive to their initialization.

To address the optimization challenge, 
we propose an initialization strategy using an auxiliary voxel grid, which provides an initial estimate for moving parts by computing the errors in the foreground masks when rendering target views for the source articulation.
Additionally, we introduce a decoupled optimization procedure that alternates between optimizing the part segmentation on the photometric error and the articulation parameters on the foreground prediction error.

The key advantage of our method, compared to other unsupervised articulation prediction methods \cite{jiang2022ditto,liu2023paris}, is its stable performance across different object and articulation types, its ability to learn from few target views and to model multiple moving parts.
Thanks to the stage-wise training, we achieve high-quality object geometry and appearance by using the well-optimized implicit models~\cite{barron2022mip}.
Since the part segmentation and articulation parameters form a small portion of the total weights, along with the initialization and decoupled optimization strategies, our method efficiently learns them from few target views, unlike the most relevant work \cite{liu2023paris} that jointly optimizes multiple part-specific implicit functions from scratch.

%% file: sections/02relwork.tex
\begin{figure}
\includegraphics[width=\linewidth]{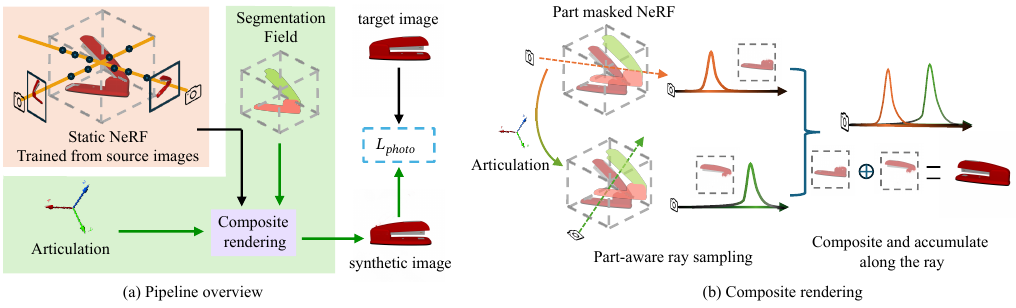}
\caption{\label{fig:overview} (a) Our method learns the geometry and appearance of an articulated object by first fitting a NeRF from (source) images of an object in a fixed articulation. Then, from another set of (target) images of the object in another articulation, we distill the relative articulation and part labels. \textcolor{green}{Green} lines show the gradient path during this distillation.
(b) Using the part geometry and appearance from NeRF, we render the target images by compositing the parts after applying the predicted articulations to the segmented parts. The photometric error provides the required supervision for learning the parts and their articulation without groundtruth labels.}
\end{figure}
\section{Related work}
\paragraph{Analysis of articulated objects}
The analysis of articulated objects typically involves segmentation of movable parts, and estimating their attributes such as position and direction of joint axes, rotation angles, and translation distances. 
Prior works study articulated objects using 3D input such as meshes~\cite{sharf2014mobility_mesh, mitra2010illustrating_mesh}, point clouds~\cite{yuan2016space, yi2018deep, wang2019shape2motion} or RGBD images \cite{liu2022toward,geng2023gapartnet} that are error-prone and labor-intensive to collect in real-world scenarios.
Recent works~\cite{jiang2022opd, qian2022understanding, sun2023opdmulti} that use RGB images to segment 3D planar object parts and estimate their articulation parameters simultaneously require ground-truth labels for segmentation and 3D articulation.

\paragraph{Articulated object modeling via novel view synthesis}
Neural implicit models~\cite{mescheder2019occupancy, park2019deepsdf,mildenhall2021nerf} that are originally designed to model static objects in a 3D consistent way have been extended to articulated objects by multiple recent works~\cite{mu2021asdf,tseng2022cla, jiang2022ditto,lewis2022narf22,wei2022nasam,liu2023paris}.
A-SDF~\cite{mu2021asdf} learns a disentangled latent space for shape and articulation to synthesize novel shapes and unseen articulated poses for a given object category.
CARTO~\cite{heppert2023carto} extends A-SDF to multiple object categories and uses a stereo image pairs as input. 
Similarly, CLA-NeRF~\cite{tseng2022cla} learns to perform joint view synthesis, part segmentation, and articulated pose estimation for an object category from multiple views.
NARF22~\cite{lewis2022narf22} learns a separate neural radiance fields (NeRF) for each part and composes to render the complete object.
Unlike our method, both A-SDF and CARTO require 3D shape and articulation pose supervision, CLA-NeRF requires part segmentation labels and is limited to modeling only 1D revolutions for each joint, NARF22 relies on groundtruth part labels and articulation pose.


Most related to our work, DITTO~\cite{jiang2022ditto} and PARIS~\cite{liu2023paris} aim to estimate part segmentation and articulation pose without labels using from an observation pair
of an object in two different articulations.
DITTO uses a pair of point cloud as input, can only model shape, whereas both PARIS and our method uses pairs of multi-view images, and to model both geometry and appearance. 
PARIS adopts the dynamic/static modelling of \cite{wu2022d,yuan2021star}, learns a separate NeRF for the dynamic and static parts. They are composited using the estimated relative articulation to render the observations with the different articulation.
While our method is also based on the same analysis-by-synthesis principle to supervise the training, it differs from PARIS in several key ways.
Our method involves only a single NeRF that is learned on multiple views of an object instance in a fixed articulation pose. Once the NeRF is learned, we freeze its parameters, and learn a segmentation head and articulation to selectively deform the rays while rendering views of different articulations.
This means our model's size remain nearly constant when the number of parts increases, which, combined with our two step optimization strategy, leads to more stable and data-efficient learning, yielding significantly better performance.
Furthermore, we show that our model goes beyond modeling a single moving object part as in PARIS, and successfully learn multiple moving parts.

\paragraph{Deformable NeRF}
There exists several techniques~\cite{park2021nerfies,pumarola2021d, tretschk2021non,park2021hypernerf, fang2022fast} that model differences between subsequent video frames of a dynamic scene through a deformation field in 3D.
While these techniques are general could be used in modeling articulated objects in principle, the deformation field does not provide explicit predictions for part segmentation and articulation.
Prior methods that focus on modeling specific articulated and deformable objects such as human body~\cite{weng2022humannerf, su2021anerf, huang2020arch, icsik2023humanrf, noguchi2021neural} and four-legged animals~\cite{wu2023magicpony} leverages specialized 3D priors~\cite{SMPL} that are not applicable to arbitrary object categories.

%% file: sections/03method.tex
\newcommand{\PT}{\ensuremath{\mathcal{P}}}


\section{Review: NeRF}
\label{sec:nerf}
Given a set of images $\mathcal{I}$ of an object, a NeRF function \cite{mildenhall2021nerf} maps a tuple $(\bm x, \bm d)$, where  $\bm{x} \in \mathbb{R}^3$ is a  3d position and $\bm{d} \in \mathbb{R}^2$ is a direction,  to an RGB color $\bm c$, and a positive scalar volume density $\sigma$. The model outputs volume density $\sigma$ at the location $\bm{x}$ along with a latent feature vector $z$ which is then concatenated with the viewing direction $\bm{d}$ and fed into an MLP to estimate the view-dependent RGB color $\bm c$. 
With a slight abuse of notation, we also use  $\sigma$ and $z$ as functions $\sigma(\bm x)$ and  $z(\bm x)$ respectively and color as a function with $c (\bm{x}, \bm{d})$. That is, we use boldface to denote vectors.
 
NeRFs are trained by expressing the color at each pixel of images in the training subset. 
The color at a pixel $C$ is estimated as $C_N$ via the volume rendering equation as a sum of contributions from points along the ray through the pixel, say $r\in \mathcal{R}$. Points $\bm x_i, i=1\cdots n$ are first sampled along the ray so that $\bm x_i = \bm o + t_i\bm d$ where $\bm o$  is the origin of the ray (usually the point of projection of a camera view) and $t_i$ is a scalar. 
Using these samples, the estimated color is given by
\begin{align}
     {C}_N(r) = \sum^n_{i = 1} \; T_i^r \; (1 - \exp(-\sigma_i^r \delta_i^r) \mathbf{c}^r_i) \quad \mathrm{where}  \quad T_i^r = \sum^{i-1}_{j=1} \exp(-\sigma_j^r \delta_j^r)
     \label{eqn:rendering}
\end{align}
where we use the shorthand notation $\bm c_i^r$ and $\sigma_i^r$ to denote $c(\bm x_i, \bm d)$ and $\sigma (\bm x_i)$ respectively, and $\delta_i = t_{i+1} - t_i$ is the distance between samples. And the opacity value is calculated as:
Similarly the opacity value for the pixel can be computed as ${O}_N(r) = \sum^n_{i = 1} \; T_i^r \; (1 - \exp(-\sigma_i^r \delta_i^r)$.
The parameters of $c$ and $\sigma$ are obtained by minimizing the photometric loss between predicted color ${C}_N(r)$ and the ground truth color $C(r)$ as
\begin{equation}
    \mathcal{L}_{\text{photo}} = \sum_{r\in \mathcal{R}} ||{C}_N(r) - C(r)||^2_2. 
    \label{eqn:photo_metric}
\end{equation}

\section{Method}
\label{sec:method}

Let $\mathcal{I}$ and $\mathcal{I}'$ be two observations of an object with $k$ (known a priori) rigid parts in articulation poses \PT\ and $\PT'$ respectively.
Each observation contains multiple views of the object along with the foreground masks.
We call $\mathcal{I}$ the source observation and $\mathcal{I}'$ the target observation.
\PT\ (resp. $\PT'$) is a pose tensor composed of $P_{\ell} \in SE(3), \; {\ell}=1\cdots k$ (resp. $P_{\ell}'$) transformations as $4\times 4$ matrices corresponding to the local pose of each of the $k$ parts. 
Our primary goal is to estimate a pose-change tensor $\mathcal{M}$ composed of $M_{\ell}\in SE(3), \; {\ell}=1\cdots k$ so that $P_{\ell}' = M_{\ell} P_{\ell}$.
We solve this by starting with the construction of a NeRF from $\mathcal{I}$, as described in \cref{sec:nerf} along with the efficient coarse-to-fine volume sampling strategy in \cite{barron2022mip}, followed by a novel modification to build awareness at the part-level. 
We use this modified parameterization to optimize an auxiliary voxel grid corresponding to the parts via pose estimation and part segmentation pipelines. 
Once we have identified the parts and their relative transformations, we are able to re-render novel views corresponding to articulation pose $\PT'$ by transforming view rays suitably by $M_{\ell}^{-1}$. 




%
%
%
%
%

\subsection{Part-aware rendering}
\label{sec:part_seg}

Once a NeRF function, which we call `static NeRF', is learned over $\mathcal{I}$, we freeze its parameters and append a segmentation head $s$ to it towards obtaining part-level information.
$s$ is instantiated as 2-layer MLP, and maps the latent feature vector $z(\bm{x})$ to a probability distribution over the $k$ object parts.
We denote the probability of a 3D point $\bm{x}$ to belong object part $\ell =1\cdots k$ as $s_{\ell}(z(\bm{x}))$.

If the segmentation class-probabilities $s_{\ell}$ and pose-change transformations $M_{\ell}$ are known, then the object in an unseen articulation pose can be synthesised by suitably transforming the statically trained NeRF without modifying its parameters and then compositing the individual part contributions. 
To model pose change in each part, we use a different ray, \emph{virtual ray} associated with each part 
\begin{equation}
    r_{\ell} = M_{\ell}^{-1} r
    \label{eqn:ray_deformation}
\end{equation} and the final rendered color is:
\begin{equation}
    {C}_P(r) = \sum_{i=1}^{n}\; \hat{T}_i^r  \;\; 
    \sum_{\ell=1}^{k} 
    \left(1- \exp{\left(
        -
            \left(
            s_{\ell}^{r_{\ell}} (\bm x_i) \; \sigma^{r_{\ell}} (\bm x_i)  \; \delta_{j}^{r_{\ell}}
            \right)
            \right)} \right)\mathbf{c}_i^{r_{\ell}}
    \label{eqn:color_composite}
\end{equation} where $\hat{T}_i^r = \sum_{j=1}^{i-1} 
        \exp{\left(
        -\sum_{\ell=1}^{k} 
            \left(
            s_{\ell}^{r_{\ell}} (\bm x_j) \; \sigma^{r_{\ell}} (\bm x_j)  \; \delta_{j}^{r_{\ell}}
            \right)
            \right)}$ is the transmittance sum.
Compared to the original formulation in \cref{eqn:rendering}, it can seen as the density replaced by a part-aware density which is the product of $s$ and $\sigma$. 
Note that we made a similar modification to the efficient volume sampling strategy of \cite{barron2022mip} and refer to the \cref{sec:part_proposal} for more details.

Since the groundtruth part segmentation and articulation are unknown, one could optimize the parameters of $s$ and $\mathcal{M}$ on the target observation such that the photometric error in between the part-aware rendered views and $\mathcal{I}'$.
Note that ${C}_P(r)$ is function of $\mathcal{M}$ (through $r_{\ell}$ in \cref{eqn:ray_deformation}) and $s$, they can be jointly optimized through backpropagation.
However, jointly solving these two intertwined problems through rendering is challenging and highly sensitive to their initial values, as their solutions depend on each other.
There, we introduce an alternating optimization strategy next.




\subsection{Decoupled optimization of $s$ and $\mathcal{M}$}
\label{sec:optimization}

To learn the part segmentation and pose-change, we first introduce an auxiliary voxel grid that assigns each 3D coordinate either to a background or a part.
Once the voxel is initialized, our training iterates multiple times over a three-step  procedure that includes optimization of $\mathcal{M}$, $s$, and refinement of the voxel grid entries.

\begin{figure}[t!]
    \centering
    \includegraphics[width=\linewidth]{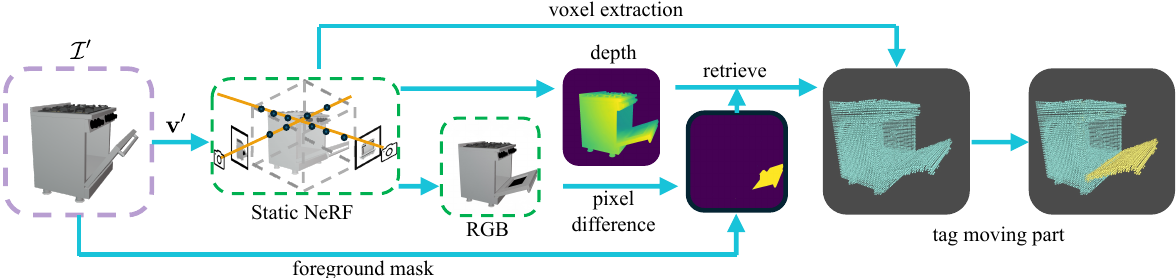}
    \caption{Voxel initialization: identify the voxels belonging to moved parts based on pixel opacity difference. }
    \label{fig:voxel_init}
\end{figure}

\paragraph{Initialization of voxel grid} Here our key idea is to find the pixel differences between the source and target observations by rendering a target view $v'$ by using the static NeRF, and label the 3D locations of the different pixels as the movable parts to provide a good initialization for estimating the articulation (see \cref{fig:voxel_init}). 
To this end, we first build a 3D voxel of the static NeRF by discretizing its 3D space into 128 bins along each dimension.
To compute the voxel entries, we query the density value from the static NeRF at each voxel coordinate and binarize it into occupancy values with criteria $\text{exp}(-\sigma(\bm{x}) \delta ) > 0.1$.
Next, we use the static NeRF to render images for the viewpoints $v'$ that are used to capture $\mathcal{I}'$.
We compute difference in opacity value using $O_N$ between these rendered images and the foreground masks of $\mathcal{I}'$, and binarize each value with $0.5$ as threshold, and collect the ones that are not contained in the groundtruth foreground mask of the corresponding view in $\mathcal{I}'$.
For those pixels, we further compute their depth by accumulating their density values along the corresponding ray, and use their estimated depth values to tag the occupied voxel entries either with static or dynamic parts. 
In the case of multiple moving parts, which is assumed to be known, we perform a clustering step~\cite{ester1996dbscan} to identify $k-1$ clusters corresponding to moving parts, as we assume that 1 part of the object is static without loss of generality.
Finally, we gather the voxel coordinates that is assigned to part $\ell$ to form a matrix of 3D coordinates $X_{\ell}$. 


\begin{wrapfigure}[12]{r}{0.5\textwidth}
    \includegraphics[width=\linewidth]{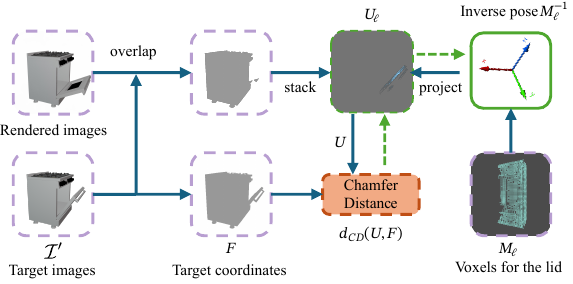}
    \caption{Illustration for optimization of $M_\ell$. The \textcolor{green}{green dotted line} shows the gradient flow.}
    \label{fig:opt_M}
\end{wrapfigure}

\paragraph{Step 1: Optimization of $M_{\ell}$}
As depicted in \cref{fig:opt_M}, we project the 3D coordinates $X_{\ell}$ for part $\ell$ onto the image plane for each camera viewpoint $v'$ included in $\mathcal{I}'$ by using $3 \times 4$ dimensional intrinsic camera matrix $K$ as $U_{\ell} = K M_{\ell}^{-1} \bm{v'} X_{\ell}$, and denote the projected point matrix as $U_{\ell}$.
We concatenate the part-specific matrices to obtain $U$.
As the groundtruth pixelwise part segmentation for part $\ell$ is unknown, we instead use the foreground masks for each view $\mathcal{I}'$ that consists of all the part pixels, and stack the 2D coordinates to form the matrix $F$.
Note that both $U$ and $F$ are matrices where each row corresponds to a 2D image coordinate where $F$ has significantly more rows, as $U$ is obtained from a coarse voxel grid.
Then we minimize the 2D Chamfer distance between $U$ and $F$:
\begin{align}
    \mathcal{M}^* =  \arg\min_{\mathcal{M}} \,\, d_\text{CD} (U,F).
    \label{eqn:chamfer_distance}
\end{align}
In practice, we only project $X_\ell$ for the moved parts, and stack them with 2D image coordinates of the foreground pixels in both the rendered image and the groundtruth image, as illustrated in \cref{fig:opt_M}.

\paragraph{Step 2: Optimization of $s_{\ell}$}
Once we obtain the solution of the pose-change $\mathcal{M}^*$ from Step 1, we plug in it to the ray deformation in \cref{eqn:ray_deformation} and render each view in $\mathcal{I}'$ in \cref{eqn:color_composite}.
Then we compute photometric loss between the rendered views and $\mathcal{I}'$, and minimize it with respect to the parameters of $s$ only.  
In the case of multiple moving parts, we initialize the parameters in $s$ using $X_\ell$ for supervision. 
Please refer to \cref{sec:training_settings} for more details. 

\paragraph{Step 3: Voxel grid refinement} 
The initial part segmentation estimates in the voxel grid are often sparse and noisy due to inaccurate density estimation in NeRF and misassignment of pixels around the foreground and part boundaries.
To refine them, we build a 3D voxel again following steps in initialization stage, but we assign those 3D coordinates for different part $\ell$ based on the predicted label from segmentation head $s$. 
We denote the new 3D coordinates as $X_\ell^*$. 
To maintain the consistency, we maintain and update voxels in $X_\ell^*$ only when the predicted part labels agree with the voxel labels within a 3D local neighborhood of $X_\ell$. This neighborhood is defined as being within the distance of one grid cell at a resolution of 128.
Then we update $X_{\ell}$ accordingly.
Note that we only perform the refinement step, after approximately 2k iterations to ensure that the segmentation head produces confident results. 
Then we also increase the voxel resolution to 256 for each dimension. 



%% file: sections/04experiments.tex
\begin{table}[t!]
    \centering
    
    \SetTblrInner{rowsep=1pt,colsep=1pt}
    \normalsize
    \begin{tblr}{|c|c|c|c|c|c|c|c|}
        \hline
        \SetCell[r=2]{c}Metric & \SetCell[r=2]{c}{Method} &\SetCell[c=4]{c}{Revolut} &  & &  &\SetCell[c=2]{c}{Prismatic}&  \\
        \cline{3-12}
            &   &laptop &oven &stapler &fridge &blade &storage \\
        \hline
        \SetCell[r=2]{c}{$e_d$ $\downarrow$} 
        &PARIS  &0.68 $\pm$ 0.40 &1.04 $\pm$ 0.68 &2.42 $\pm$ 0.91 &0.81 $\pm$ 0.60 &48.58 $\pm$ 25.43 &\textbf{0.34 $\pm$ 0.09} \\
        &Ours  &\textbf{0.33 $\pm$ 0.04} &\textbf{0.34 $\pm$ 0.03} &\textbf{0.33 $\pm$ 0.04} &\textbf{0.54 $\pm$ 0.08} &\textbf{1.54 $\pm$ 0.07} &1.11 $\pm$ 0.14\\
        \hline
        \SetCell[r=2]{c}{$e_p$$\downarrow$\\ ($10^{-2}$)} 
        &PARIS & \textbf{0.18 $\pm$ 0.15} & \textbf{0.49 $\pm$ 0.53} &55.54 $\pm$ 39.88 &\textbf{0.33 $\pm$ 0.14} & - & - \\
        &Ours & 0.48 $\pm$ 0.06 & 1.29 $\pm$ 0.04 & \textbf{0.17 $\pm$ 0.05} & 0.44 $\pm$ 0.03 & - & - \\
        \hline
        \SetCell[r=2]{c}{$e_g$$\downarrow$}
        &PARIS          &0.60 $\pm$ 0.32  &0.68 $\pm$ 0.39  &44.62 $\pm$ 6.17     &0.87 $\pm$ 0.55  & - & - \\
        &Ours         &\textbf{0.25 $\pm$ 0.03}  &\textbf{0.35$\pm$ 0.06}  &\textbf{0.290$\pm$0.03}     &\textbf{0.60 $\pm$ 0.05}  & - & - \\
        \hline
        \SetCell[r=2]{c}{$e_t$ $\downarrow$} 
        &PARIS & - & - & - & -  &1.13 $\pm$ 0.52 &0.30$\pm$ 0.01\\
        &Ours & - & - & - & -  &\textbf{0.01 $\pm$ 0.01} &\textbf{0.02$\pm$ 0.03}\\
    \hline
    \end{tblr}
    \caption{\textbf{Part-level pose estimation results.} Our method outperforms PARIS in majority of object categories while having lower variation over multiple runs in the performance.}
    \label{tab:comparison_with_paris}
\end{table}
\section{Experiment}
\label{sec:exp}
\subsection{Dataset} 
We evaluate our method on the synthetic 3D PartNet-Mobility dataset~\cite{Xiang_2020_SAPIEN, Mo_2019_CVPR,chang2015shapenet}. 
While the dataset contains more than 2000 articulated objects from 46 different categories, we use a subset of the dataset with 6 shapes that was used in \cite{liu2023paris}.
For a fair comparison, we downloaded the processed dataset from \cite{liu2023paris} which contains 2 sets of 100 views along with their foreground masks, each with a different articulation, and also groundtruth part segmentation labels, for each shape.
In addition, we select 4 additional shapes, each with two moving parts, and apply the same data generation process to them.
Additionally we collected images of a toy car with a handheld device, with camera viewpoint estimated from kiri engine application\cite{kiri_engine}.

We train the static NeRF on 100 views from the first observation, train the part segmentation and articulation on 100 views from the second observation.
We report the performance of our method for varying number of views from the second observation in \cref{tab:ablation_frames}.


Following \cite{liu2023paris}, we report performance in different metrics for pose estimation, novel-view/articulation synthesis, and part segmentation.
\noindent\textbf{Pose estimation:} To report articulation pose performance, we report results in i) direction error $e_d$ that measures the angular discrepancy in degrees between the predicted and actual axis directions across all object categories, ii) position error $e_p$ and geodesic distance $e_d$ for only revolute objects to evaluate the error between the predicted and true axis positions and the error in the predicted rotation angles of parts in degrees, respectively, iii) translation error $e_t$ to measure the disparity between the predicted and actual translation movements for prismatic objects.
\noindent\textbf{Novel-view and -articulation synthesis:} We evaluate the quality of novel view synthesis generated by the models using the Peak Signal-to-Noise Ratio~(PSNR) where higher values indicate better reconstruction fidelity.
\noindent\textbf{Part segmentation:} We use mean Intersection over Union (mIoU) on the rendered semantic images to evaluate the accuracy of part segmentation, which is tightly related to the rendered image quality of objects in different articulation states. The ground truth segmentation is generated within the Sapien framework in \cite{Xiang_2020_SAPIEN}.
Finally, due to the lack of groundtruth segmentation label and articulation in the real data, we only report PSNR and provide qualitative evaluation.

\subsection{Results}

\paragraph{Baseline}
We compare our approach with the state-of-the-art unsupervised technique, PARIS \cite{liu2023paris} which constructs separate NeRF models for each part of an articulated object and optimizes motion parameters in an end-to-end manner. 
However, it is limited to two-part objects with only one movable part. 
As the authors of PARIS do not report their results over multiple runs in their paper, we use their official public code with the default hyperparameters, train both their and our model over 5 random initializations and report the average performance and the standard deviation.
We use 2 sets of 100 views for each object to train the models.
More implementation details can be found in the supplementary material.
We would like to note that the performance of PARIS in our experiments significantly differ from the original results despite all the care taken to reproduce them in the original way\footnote{Similar problems have been pointed out and the reproducibility of these results has been acknowledged as challenging by the authors in \href{https://github.com/3dlg-hcvc/paris/issues/10}{issue 1} and \href{https://github.com/3dlg-hcvc/paris/issues/8}{issue 2}}.




\begin{figure}
    \centering
    \SetTblrInner{rowsep=1pt,colsep=1pt}
    \scriptsize
        \begin{tblr}{
        colspec={cccccccc}
        }
        PARIS&
        \includegraphics[valign=c, width=0.12\columnwidth]{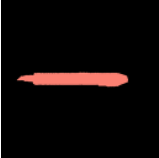}&
        \includegraphics[valign=c, width=0.12\columnwidth]{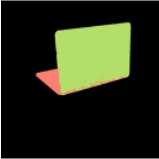}&
        \includegraphics[valign=c, width=0.12\columnwidth]{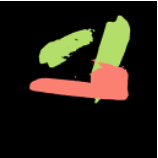}&
        \includegraphics[valign=c, width=0.12\columnwidth]{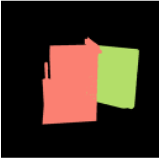}&
        \includegraphics[valign=c, width=0.12\columnwidth]{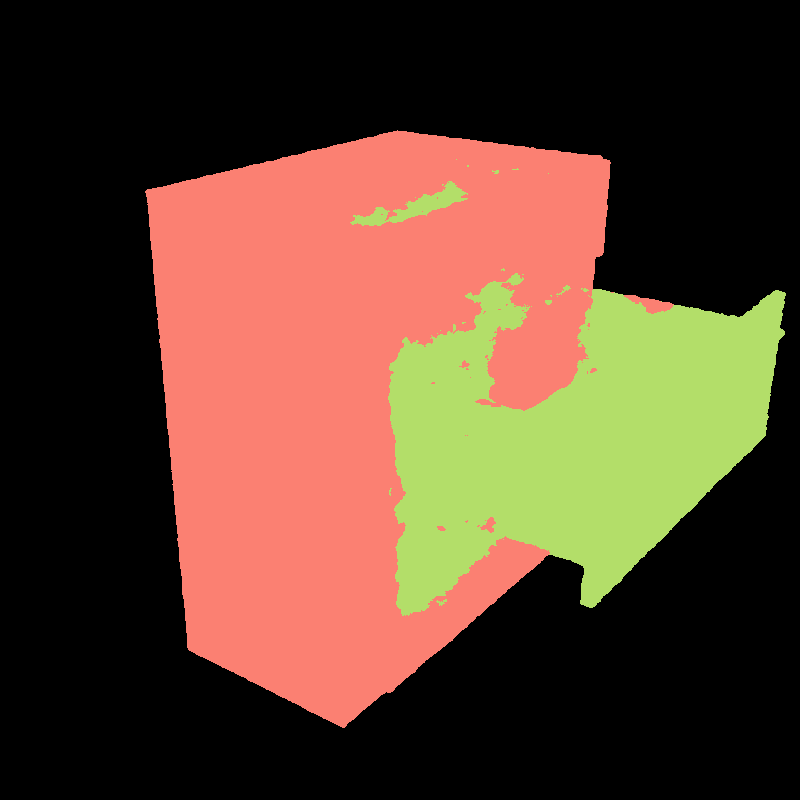}&
        \includegraphics[valign=c, width=0.12\columnwidth]{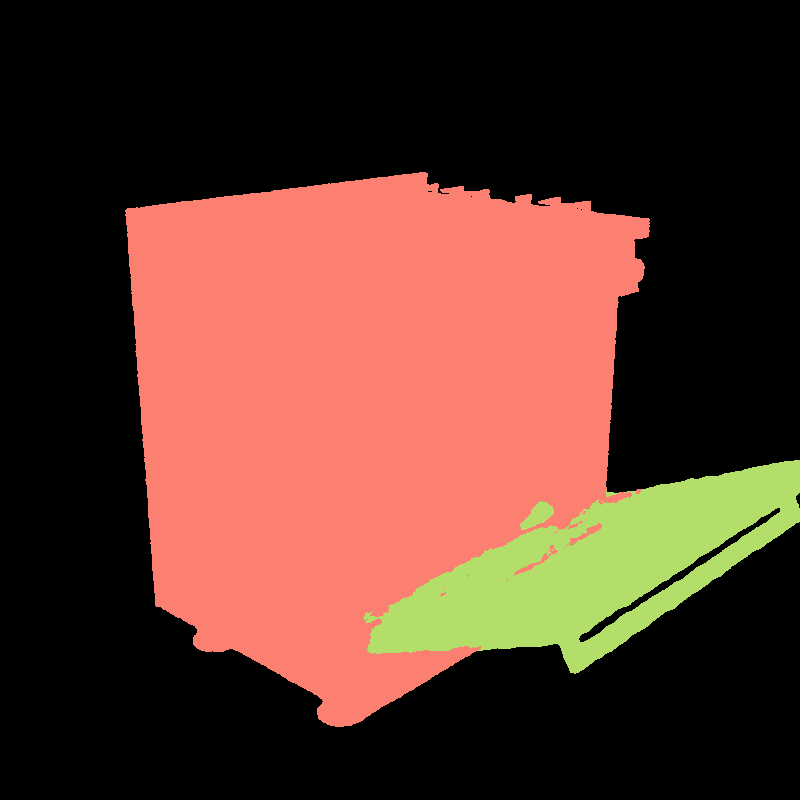}&
        \includegraphics[valign=c, width=0.12\columnwidth]{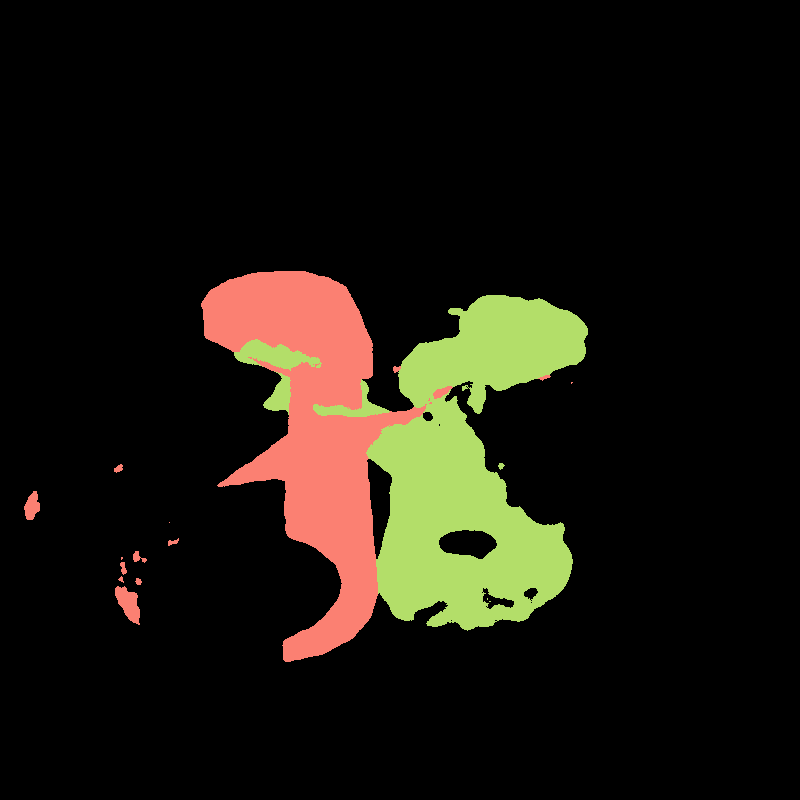}&\\
        Ours&
        \includegraphics[valign=c, width=0.12\columnwidth]{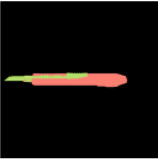}&
        \includegraphics[valign=c, width=0.12\columnwidth]{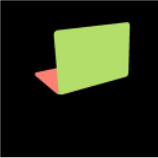}&
        \includegraphics[valign=c, width=0.12\columnwidth]{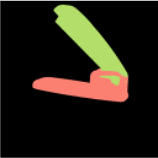}&
        \includegraphics[valign=c, width=0.12\columnwidth]{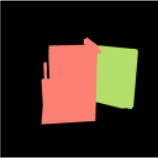}&
        \includegraphics[valign=c, width=0.12\columnwidth]{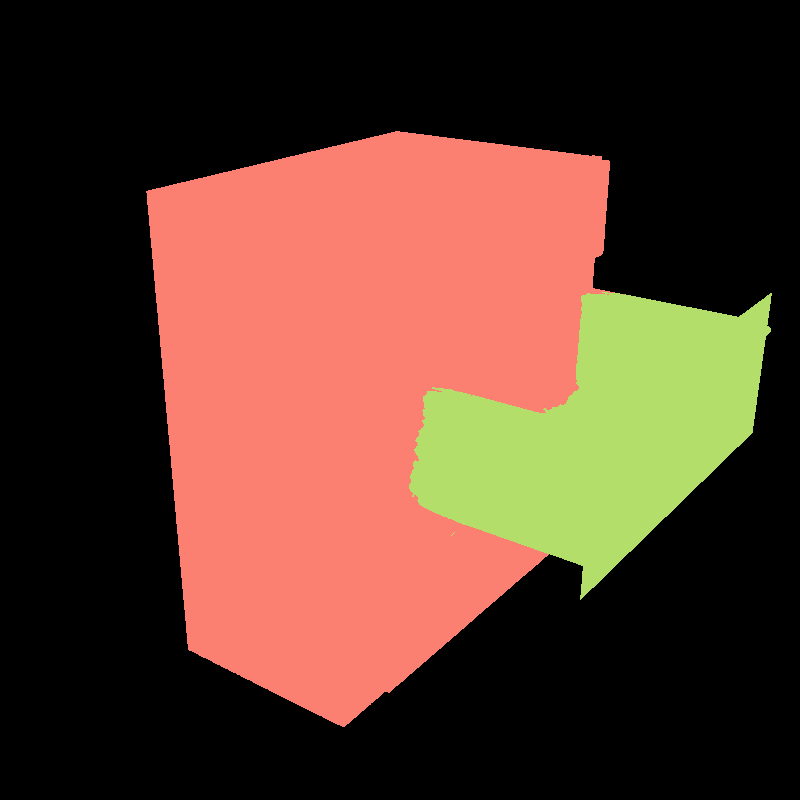}&
        \includegraphics[valign=c, width=0.12\columnwidth]{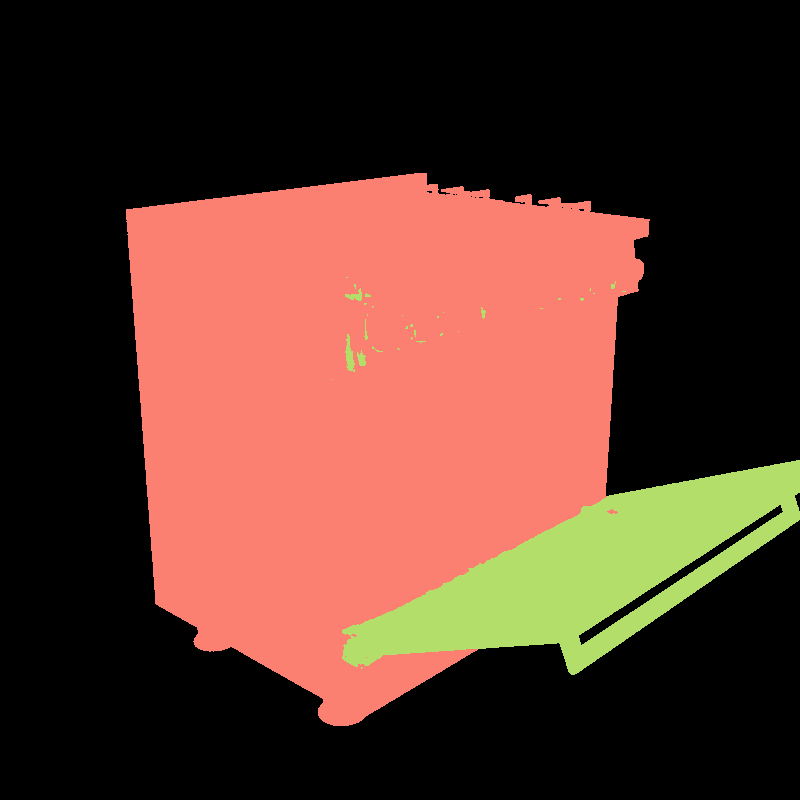}&
        \includegraphics[valign=c, width=0.12\columnwidth]{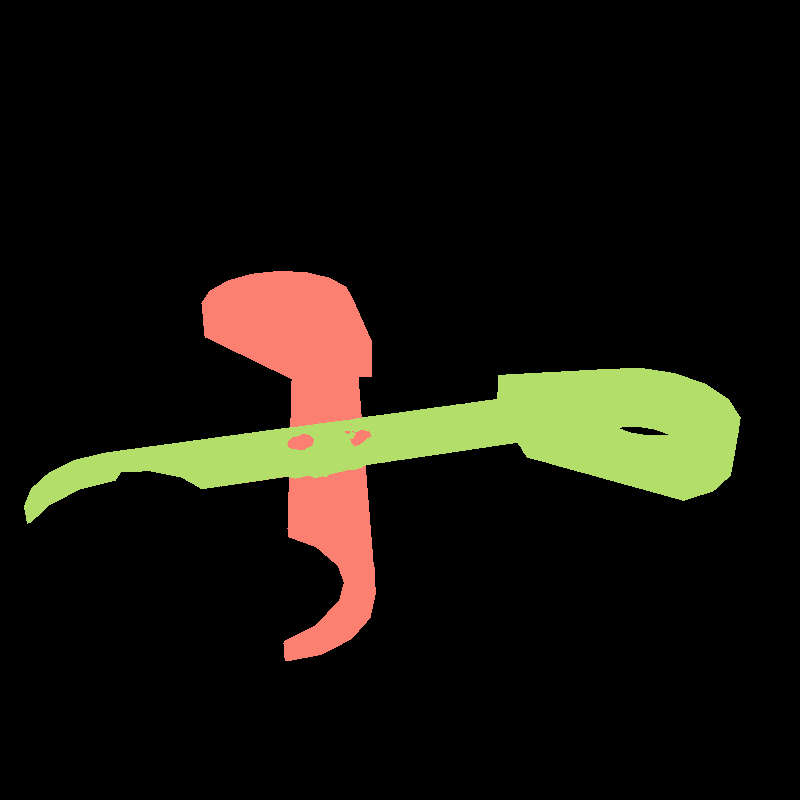}&\\
        GT &
        \includegraphics[valign=c, width=0.12\columnwidth]{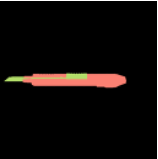}&
        \includegraphics[valign=c, width=0.12\columnwidth]{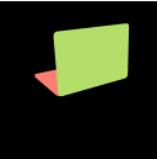}&
        \includegraphics[valign=c, width=0.12\columnwidth]{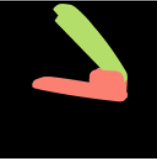}&
        \includegraphics[valign=c, width=0.12\columnwidth]{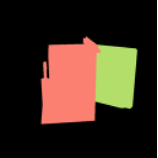}&
        \includegraphics[valign=c, width=0.12\columnwidth]{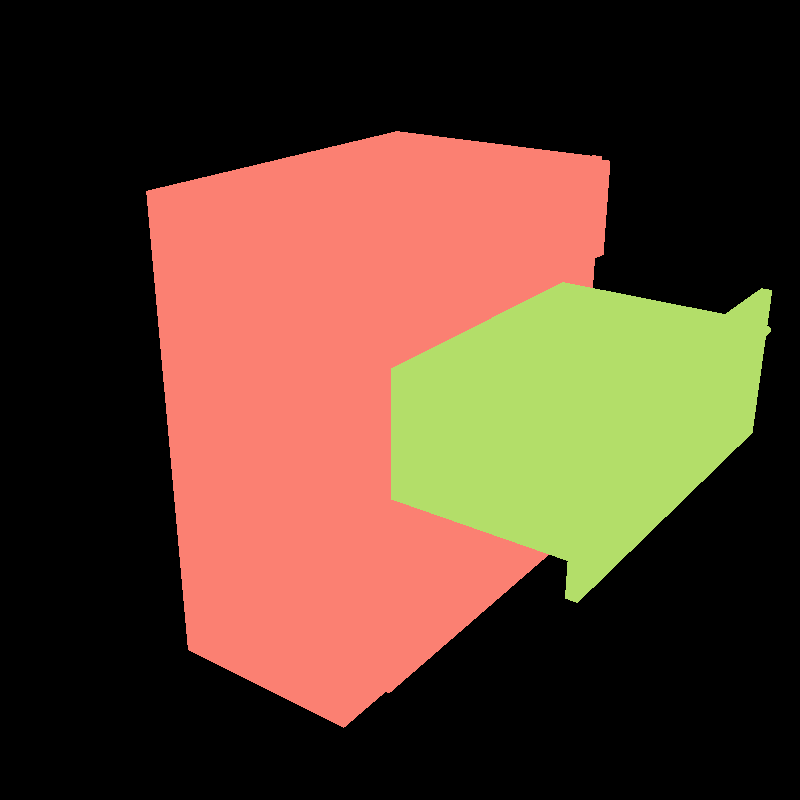}&
        \includegraphics[valign=c, width=0.12\columnwidth]{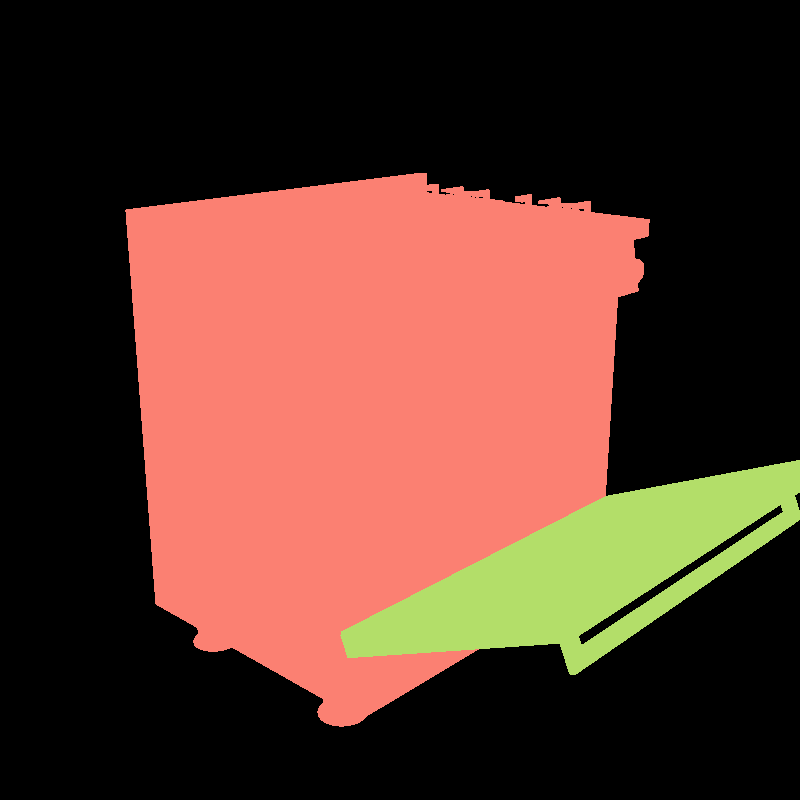}&
        \includegraphics[valign=c, width=0.12\columnwidth]{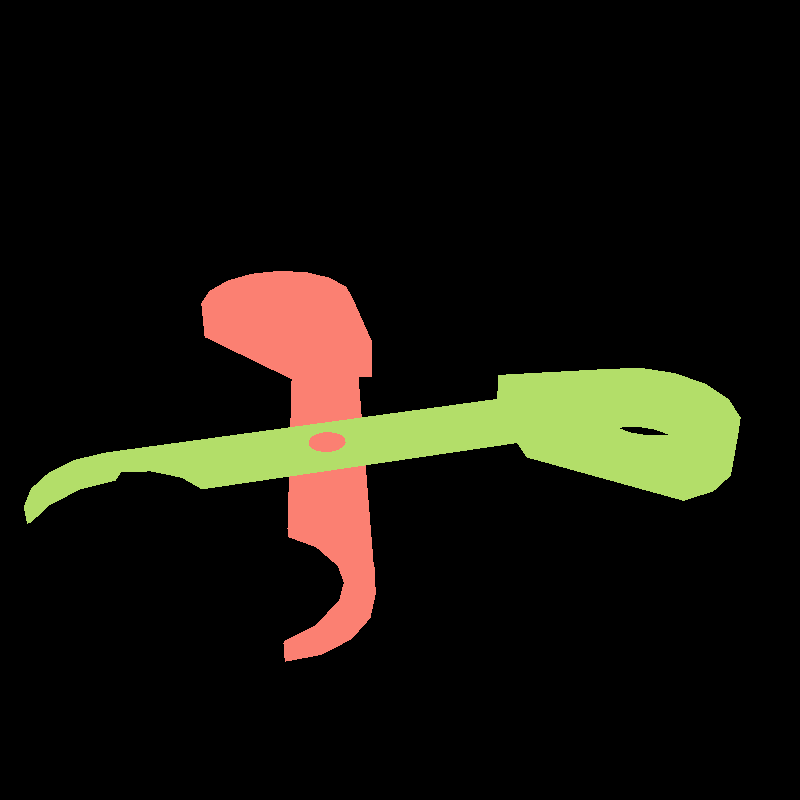}&\\
       & 
       Blade & Laptop & Stapler & Fridge & Storage & Oven & Scissor \\
    \end{tblr}
    \caption{\textbf{Qualitative 2D part segmentation results.} Pixels in \textcolor{green}{green} denotes the movable parts. Our method demonstrates consistent performance across all tested objects while PARIS failed for Blade, Laptop and Scissor. }
    \label{fig:seg_comparison}
\end{figure}
\normalsize

\begin{figure}
    \centering
    \resizebox{\columnwidth}{!}{%
    \SetTblrInner{rowsep=0pt,colsep=1pt}
    \scriptsize
    \begin{tblr}{cc|ccccc|c}
    {Method} &{ GT in $\mathcal{P}$} & \SetCell[c=5]{c}{ Novel articulation synthesis}& &&&& { GT in $\mathcal{P}'$}
        \\
        PARIS &
        \includegraphics[valign=c, width=0.1\columnwidth]{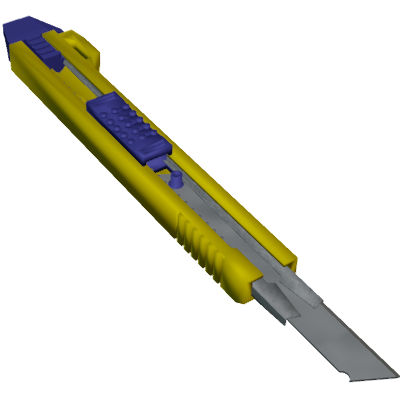}&
        \includegraphics[valign=c, width=0.1\columnwidth]{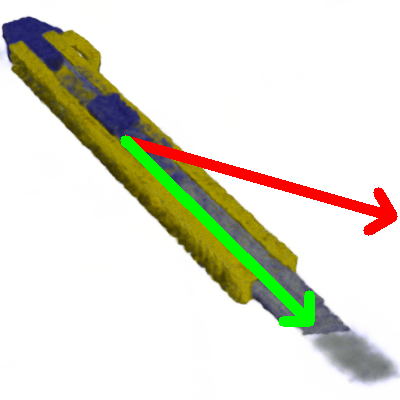}&
        \includegraphics[valign=c, width=0.1\columnwidth]{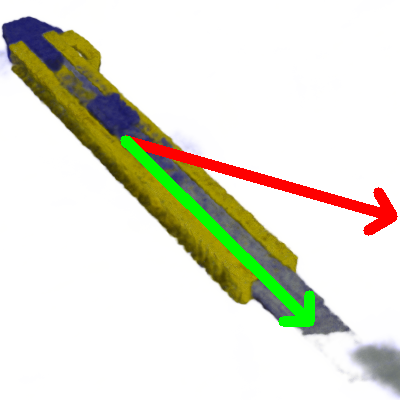}&
        \includegraphics[valign=c, width=0.1\columnwidth]{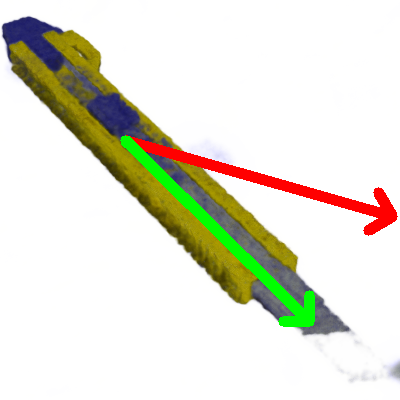}&
        \includegraphics[valign=c, width=0.1\columnwidth]{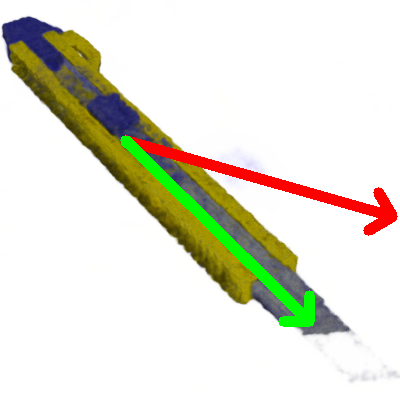}&
        \includegraphics[valign=c, width=0.1\columnwidth]{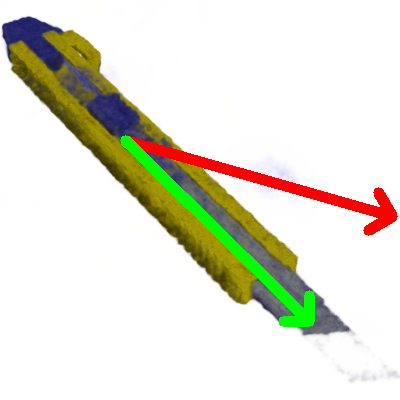}&
        \includegraphics[valign=c, width=0.1\columnwidth]{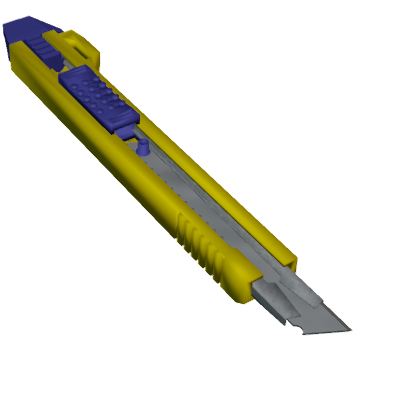}\\
        Ours &
        \includegraphics[valign=c, width=0.1\columnwidth]{figures/input_state/blade_crop/0033_start.png}&
        \includegraphics[valign=c, width=0.1\columnwidth]{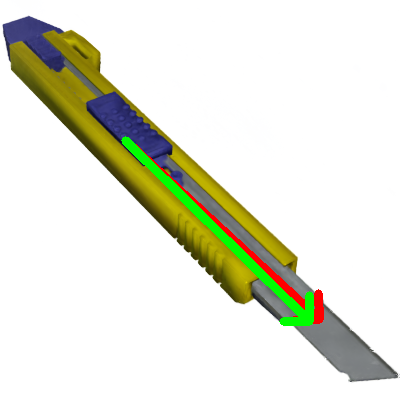}&
        \includegraphics[valign=c, width=0.1\columnwidth]{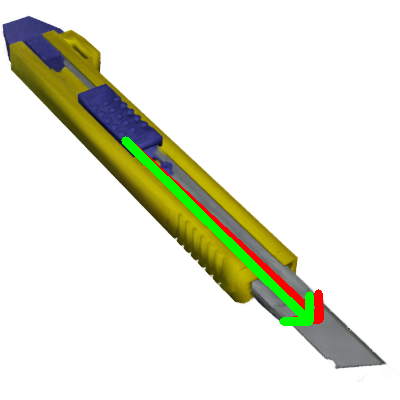}&
        \includegraphics[valign=c, width=0.1\columnwidth]{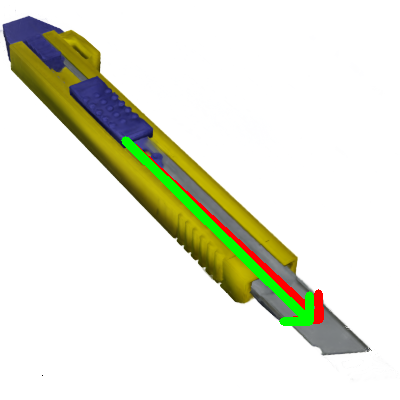}&
        \includegraphics[valign=c, width=0.1\columnwidth]{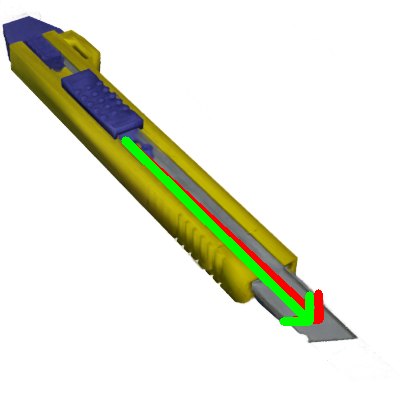}&
        \includegraphics[valign=c, width=0.1\columnwidth]{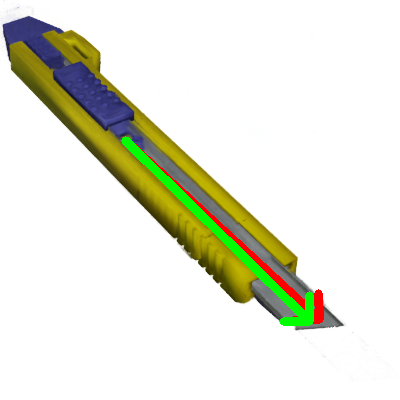}&
        \includegraphics[valign=c, width=0.1\columnwidth]{figures/input_state/blade_crop/0033_end.png}\\
        \hline
        PARIS &
        \includegraphics[valign=c, width=0.1\columnwidth]{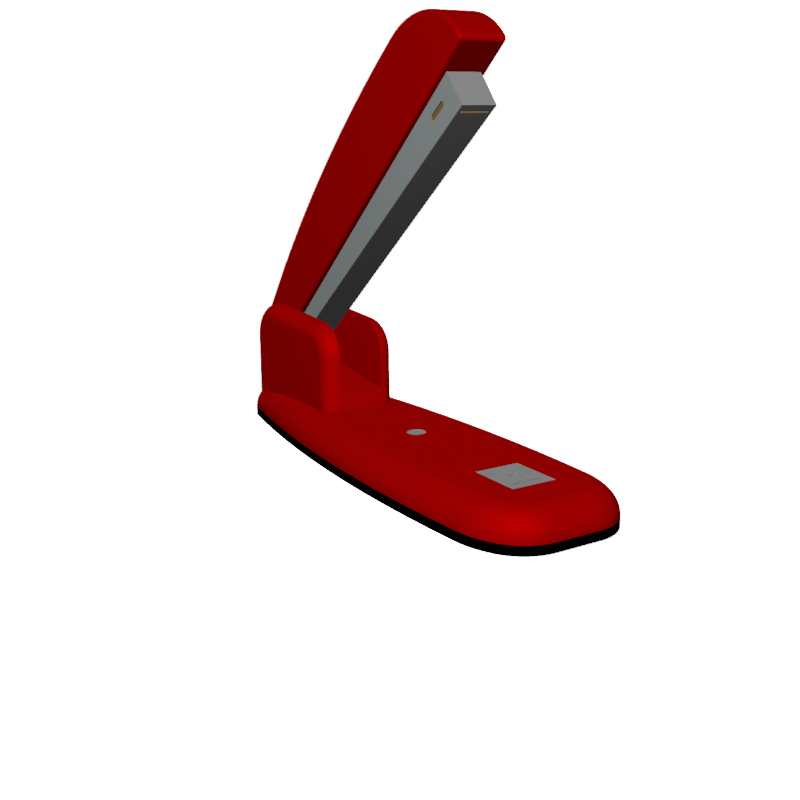}&
        \includegraphics[valign=c, width=0.1\columnwidth]{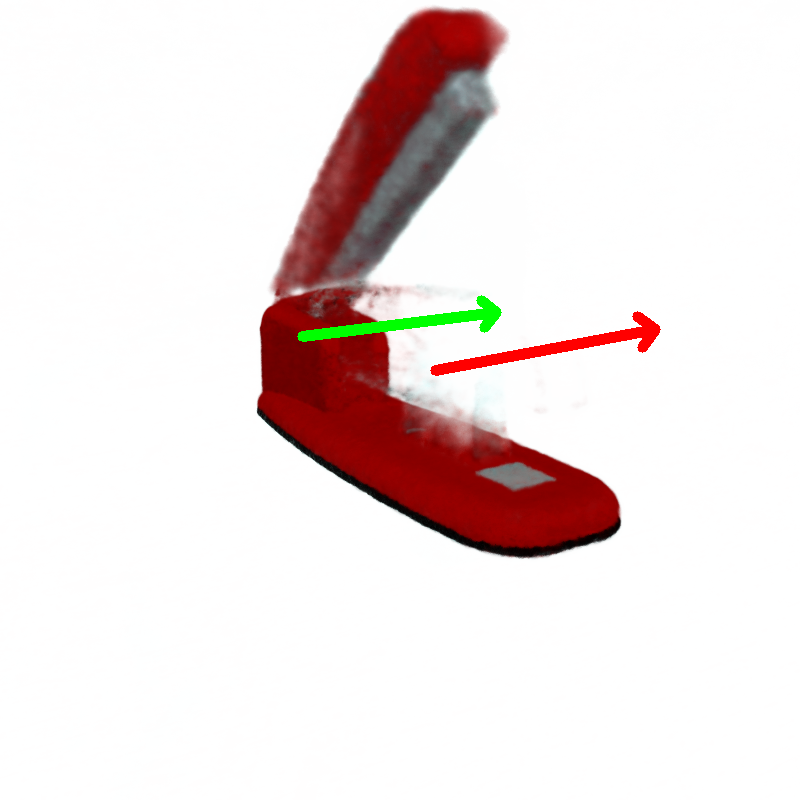}&
        \includegraphics[valign=c, width=0.1\columnwidth]{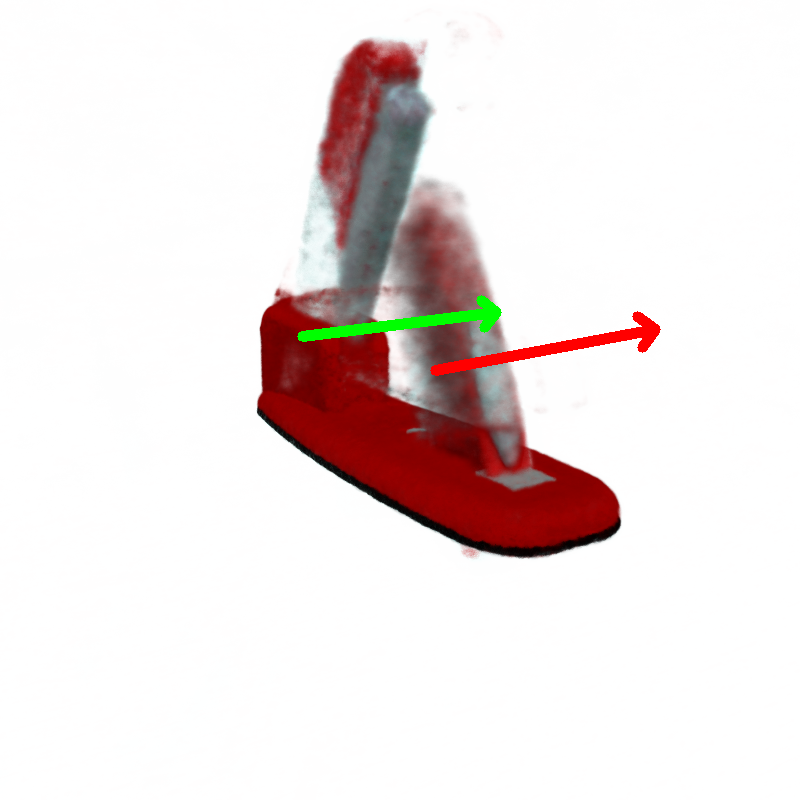}&
        \includegraphics[valign=c, width=0.1\columnwidth]{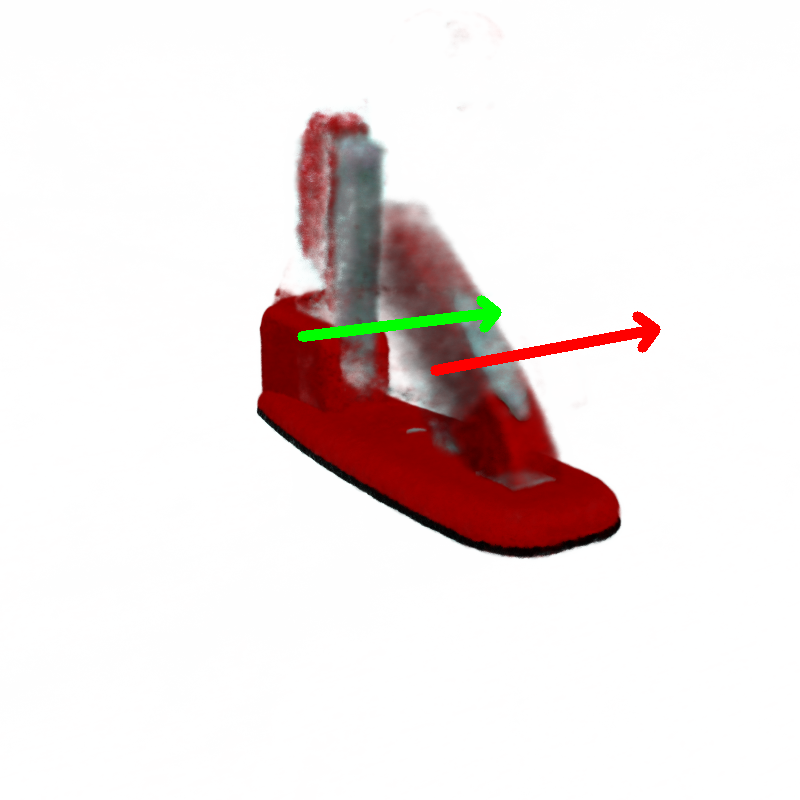}&
        \includegraphics[valign=c, width=0.1\columnwidth]{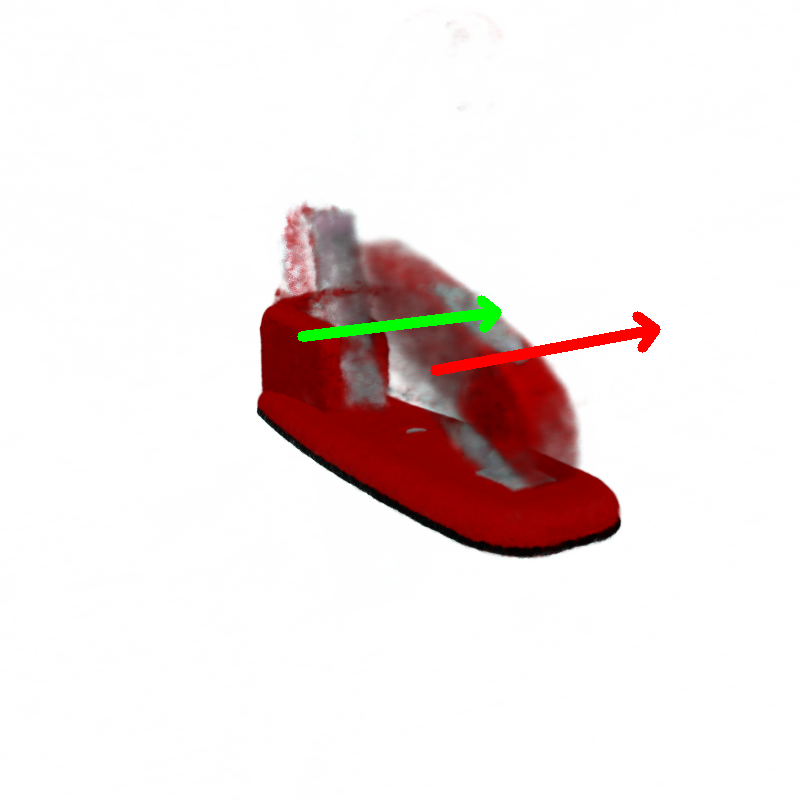}&
        \includegraphics[valign=c, width=0.1\columnwidth]{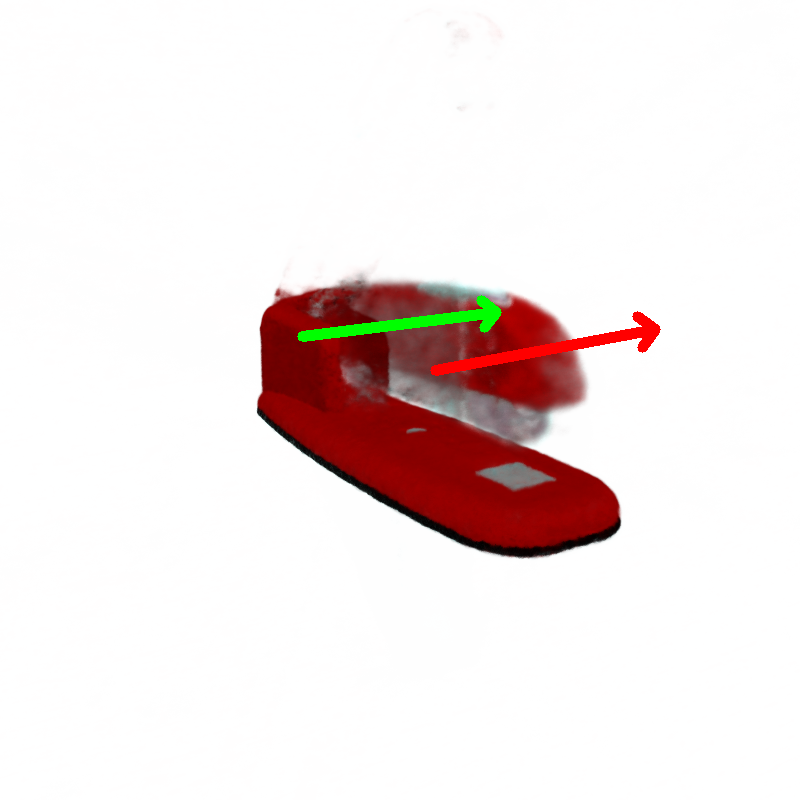}&
        \includegraphics[valign=c, width=0.1\columnwidth]{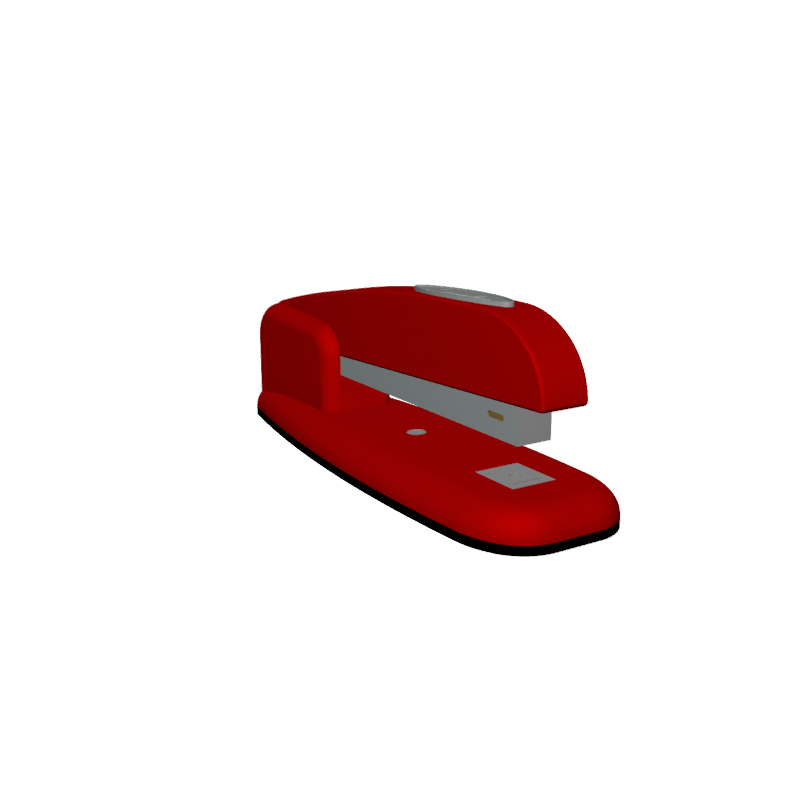}&\\
        Ours &
        \includegraphics[valign=c, width=0.1\columnwidth]{figures/input_state/stapler/0033_end.png}&
        \includegraphics[valign=c, width=0.1\columnwidth]{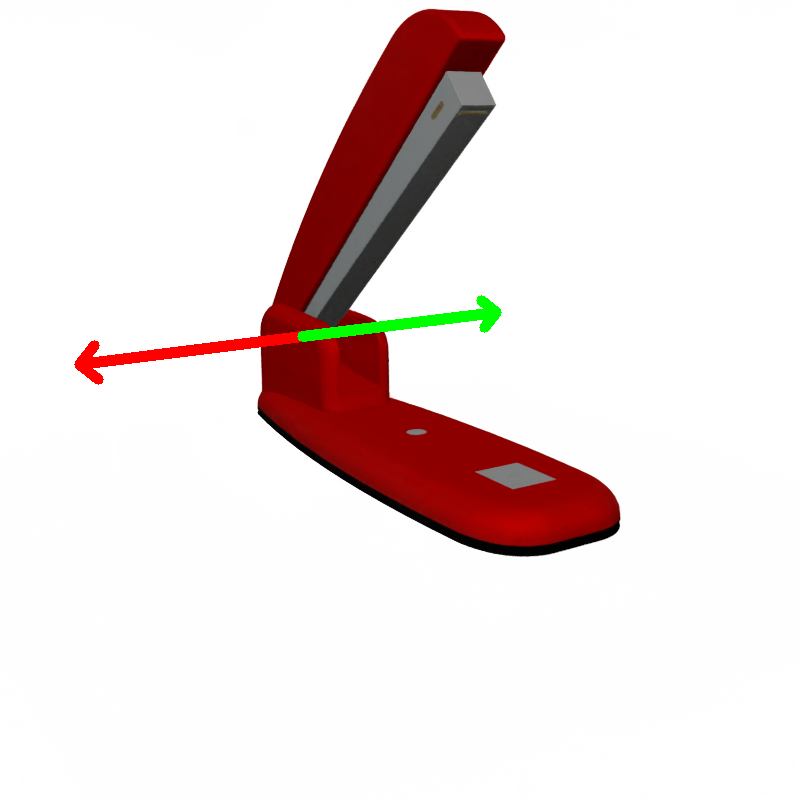}&
        \includegraphics[valign=c, width=0.1\columnwidth]{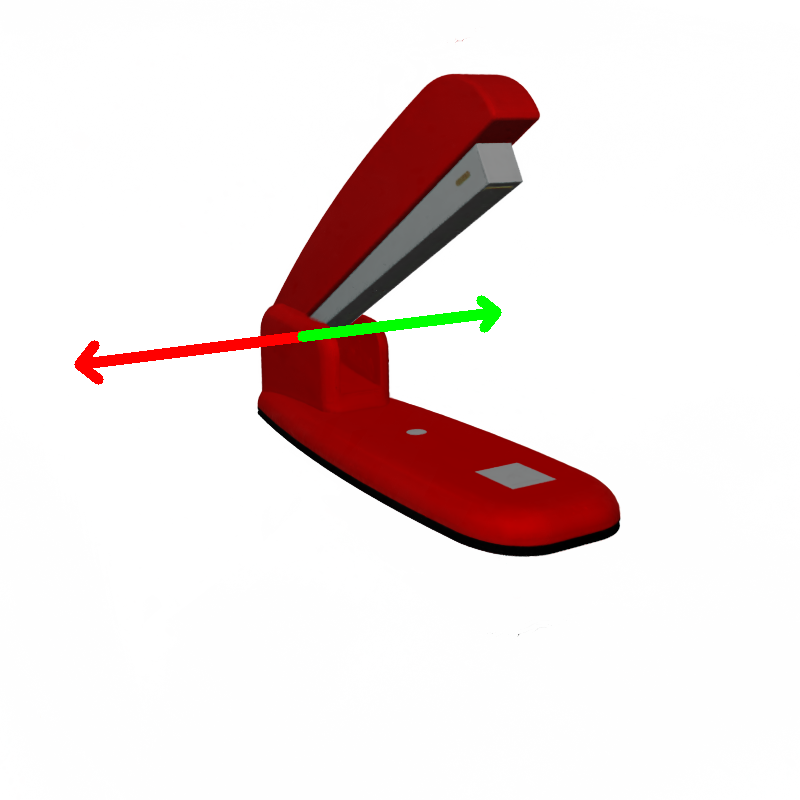}&
        \includegraphics[valign=c, width=0.1\columnwidth]{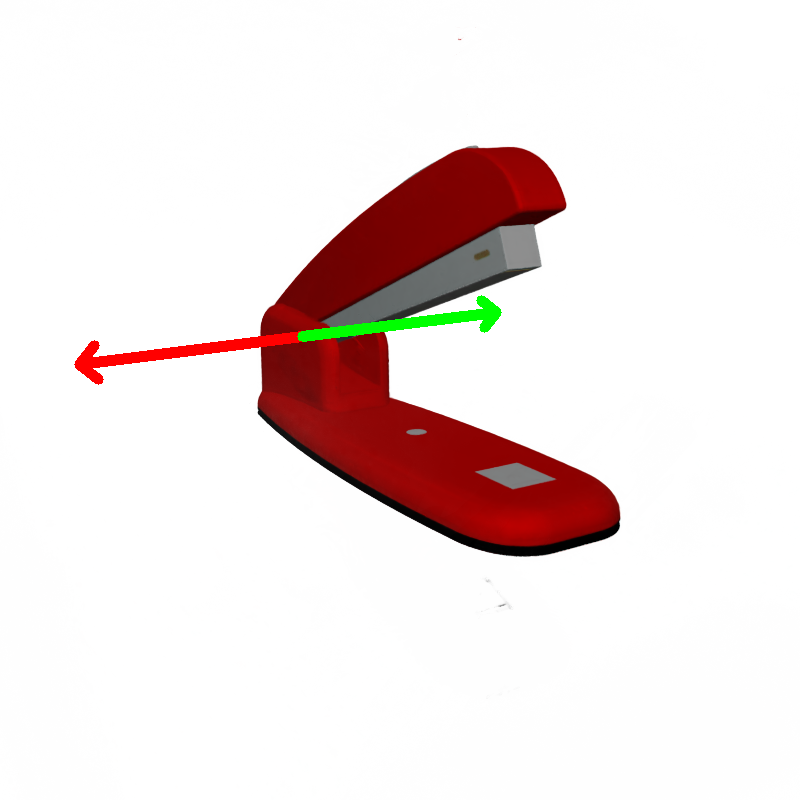}&
        \includegraphics[valign=c, width=0.1\columnwidth]{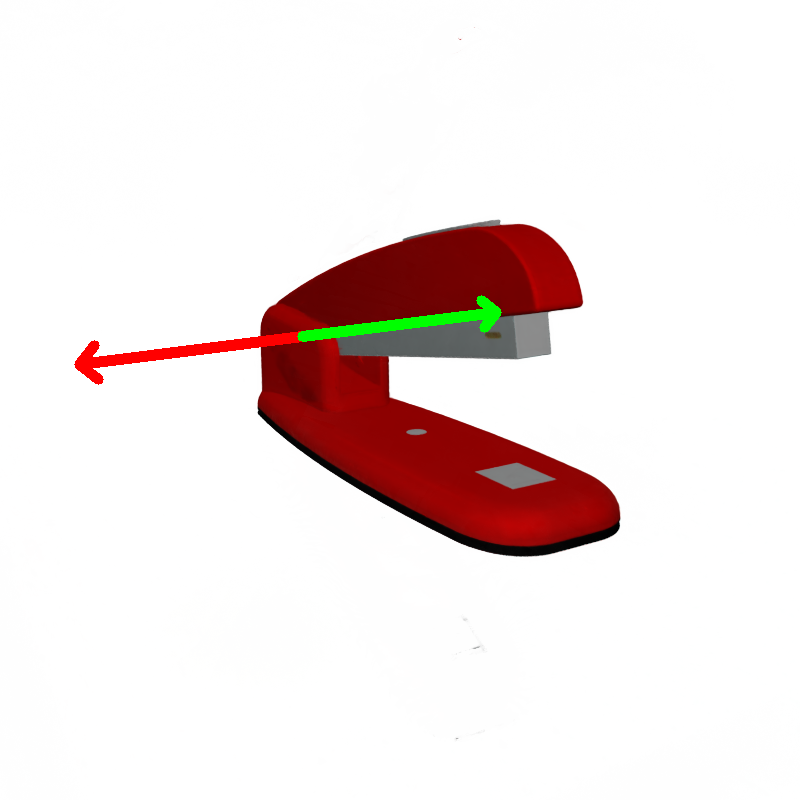}&
        \includegraphics[valign=c, width=0.1\columnwidth]{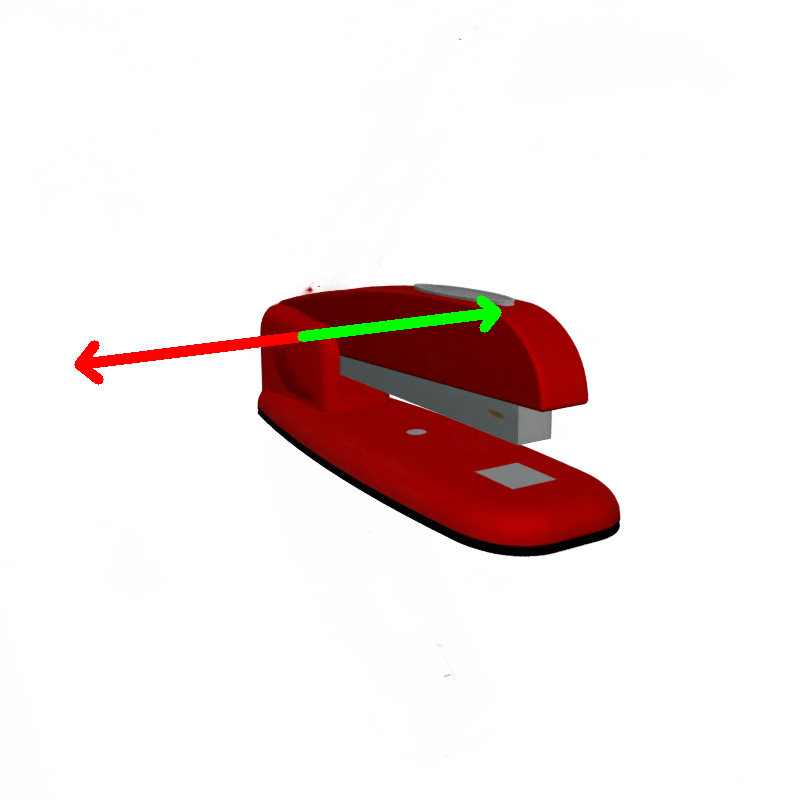}&
        \includegraphics[valign=c, width=0.1\columnwidth]{figures/input_state/stapler/0033_start.png}&\\
        \hline
        Ours &
        \includegraphics[valign=c, width=0.1\columnwidth]{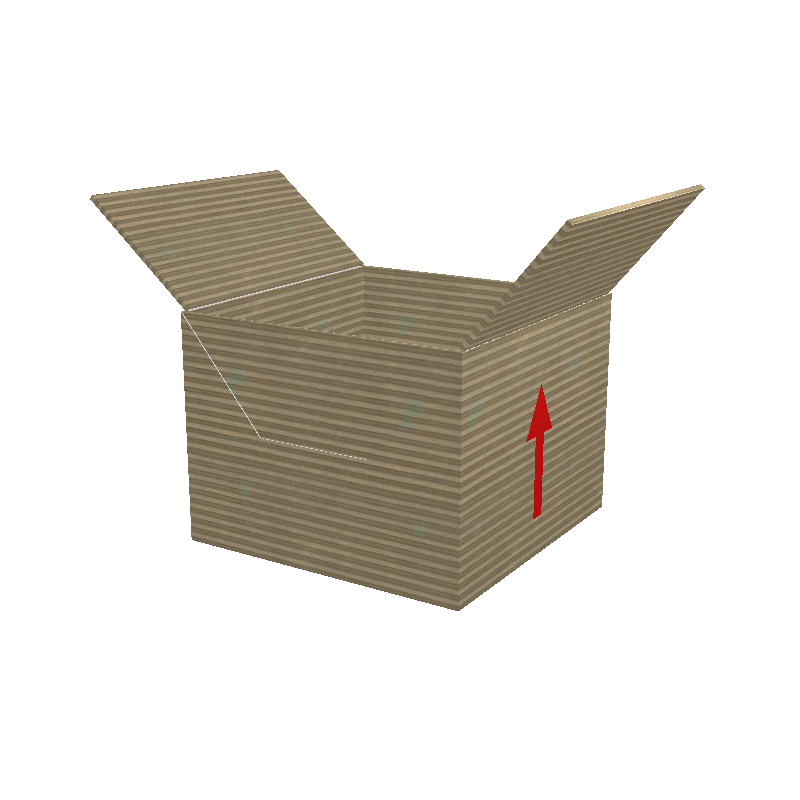}&
        \includegraphics[valign=c, width=0.1\columnwidth]{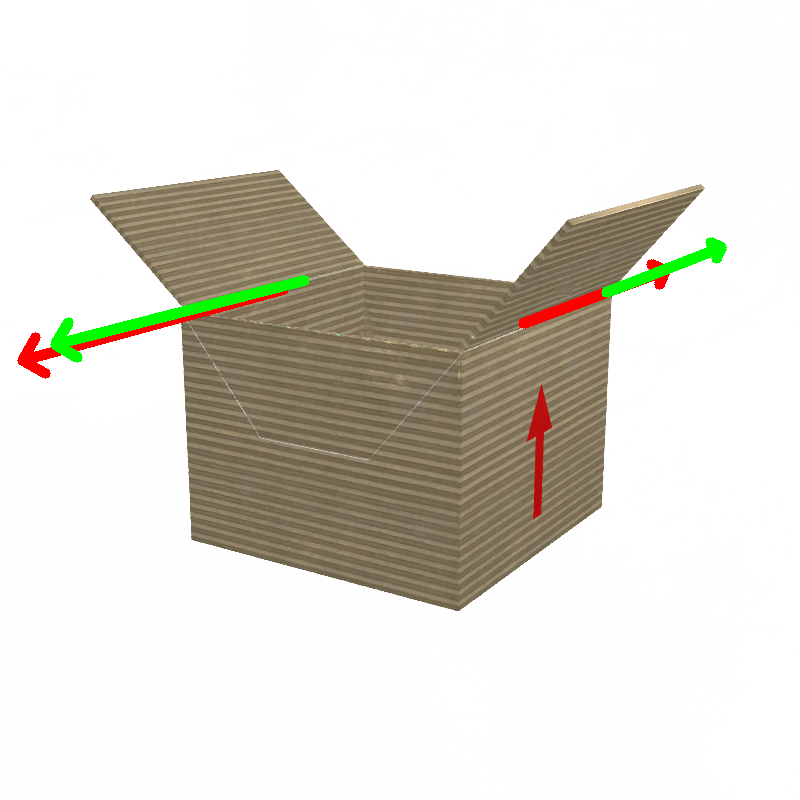}&
        \includegraphics[valign=c, width=0.1\columnwidth]{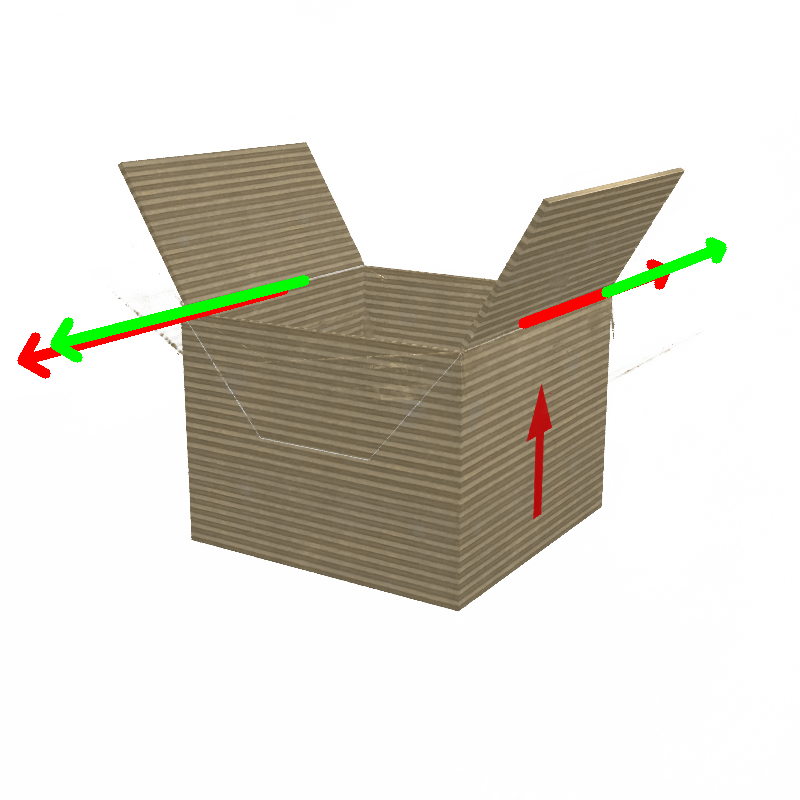}&
        \includegraphics[valign=c, width=0.1\columnwidth]{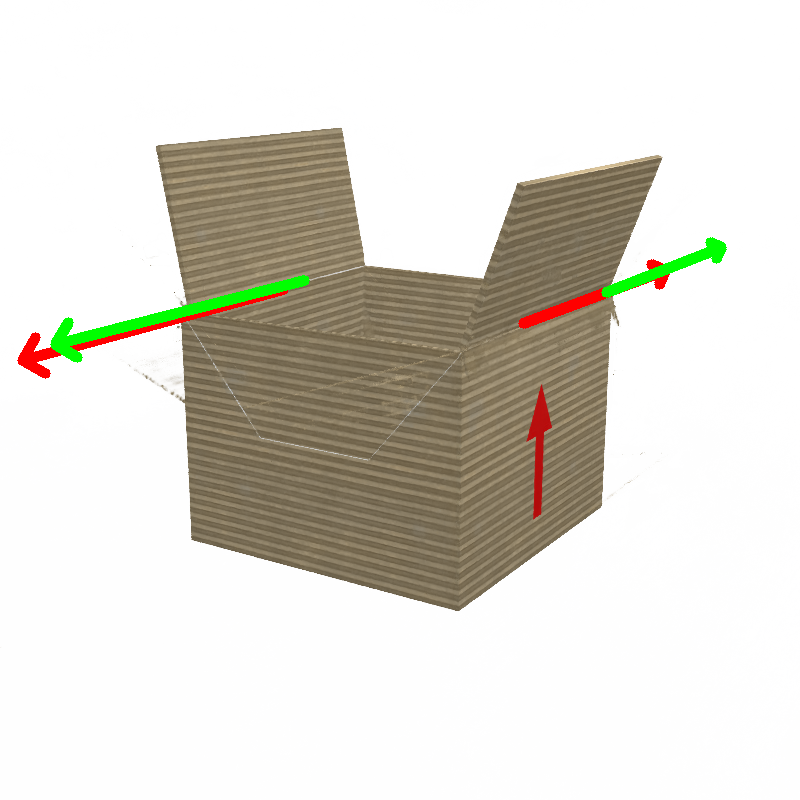}&
        \includegraphics[valign=c, width=0.1\columnwidth]{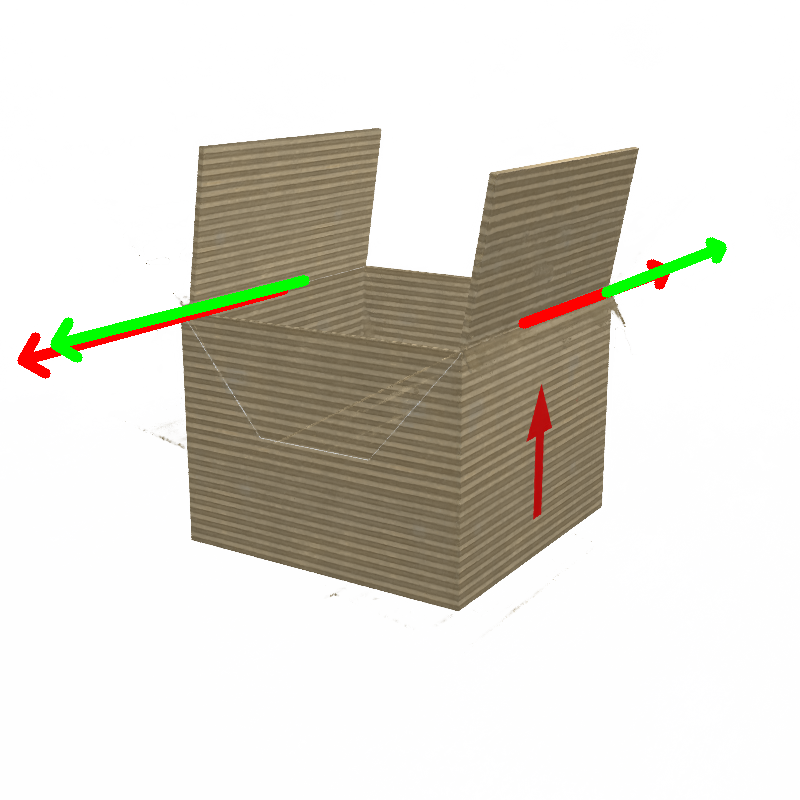}&
        \includegraphics[valign=c, width=0.1\columnwidth]{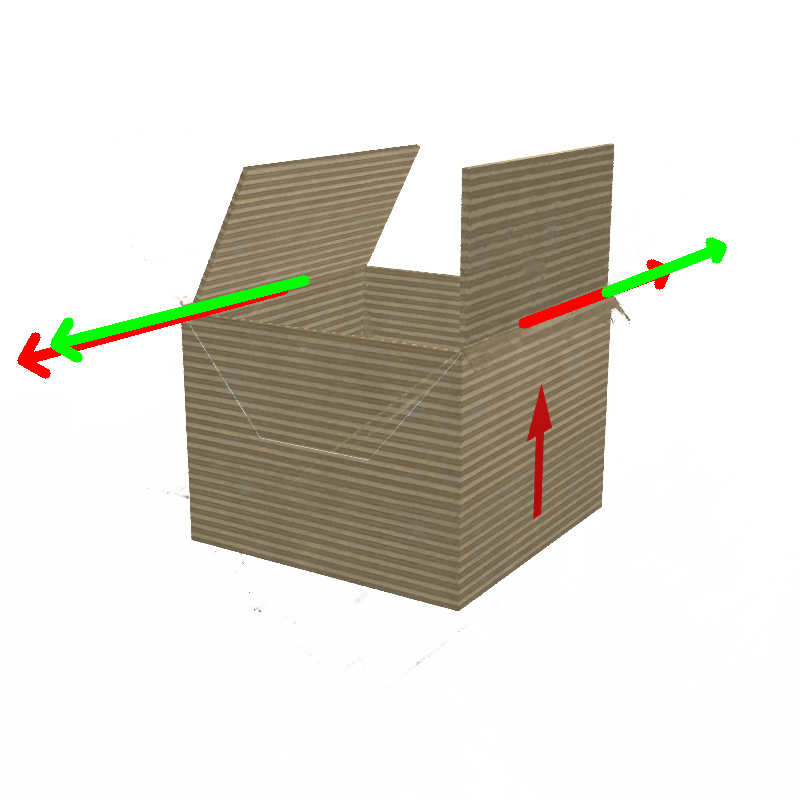}&
        \includegraphics[valign=c, width=0.1\columnwidth]{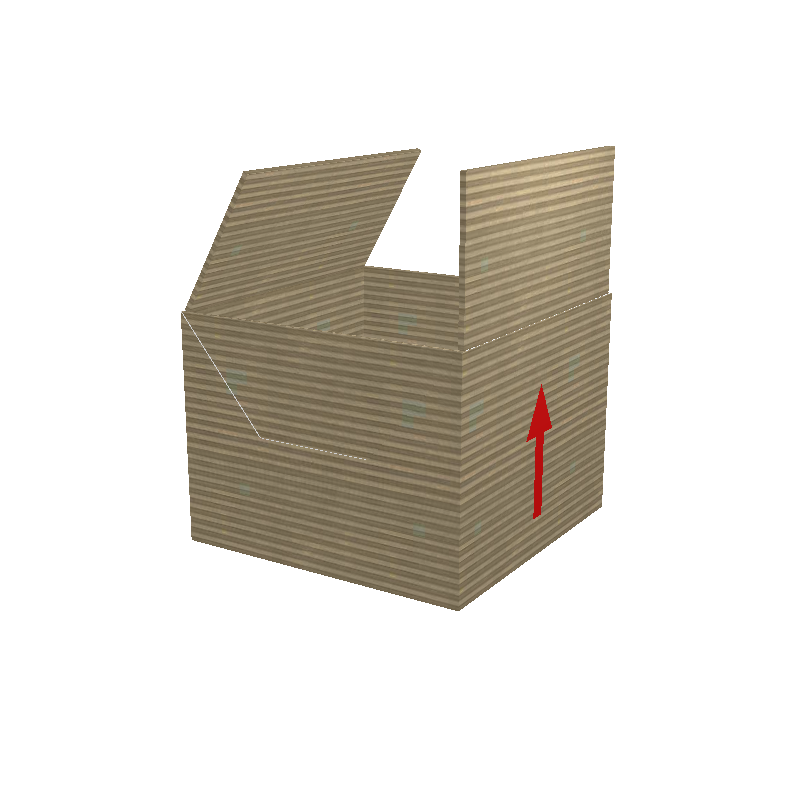}&\\
    \end{tblr}
    }
    \caption{\textbf{Qualitative evaluation for novel articulation synthesis.} The ground truth axis is denoted in \textcolor{green}{green} and the predicted axis is denoted in \textcolor{red}{red}. Please refer to the supplementary for more visualizations.
    }
    \label{fig:qualitative}
\end{figure}

\paragraph{Part-level pose estimation}
For part-level pose estimation, we provide quantitative results in average performance over 5 runs and their standard deviations in \cref{tab:comparison_with_paris}, and a qualitative analysis in \cref{fig:qualitative}. 
The results indicate that our method consistently achieves lower errors across most evaluation metrics compared to PARIS, except the joint position error $e_p$ where the differences are negligible, within the $10^{-3}$ range. 
Notably, the performance of PARIS on the stapler and blade exhibits significantly higher errors. 
We observe that PARIS fails to converge to a good solution in all 5 runs for these objects.
As shown in \cref{fig:qualitative}, PARIS fails to accurately segment the parts in the stapler and blade, which can also be easily identified in the novel articulation synthesis experiments in \cref{fig:qualitative}. 
Poor part reconstruction in PARIS results in inaccurate part-level pose estimations.
Additionally, the lower standard deviation across all reconstructed objects for our method indicates its stable learning and ability to work for different object instances.
We attribute the stability of our method to the decoupled optimization along with the stage-wise training, which contrasts with the joint optimization approach used by PARIS.

\scriptsize
\begin{wrapfigure}[16]{r}{0.5\textwidth}
\centering
    \begin{tblr}{
        rowsep=1pt,
        colsep=1pt,
        colspec={ccccc}
        }
        Ours&
        \includegraphics[valign=c, width=0.11\columnwidth]{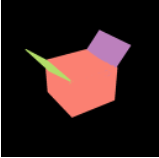}&
        \includegraphics[valign=c, width=0.11\columnwidth]{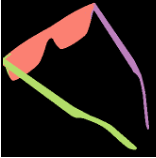}&
        \includegraphics[valign=c, width=0.11\columnwidth]{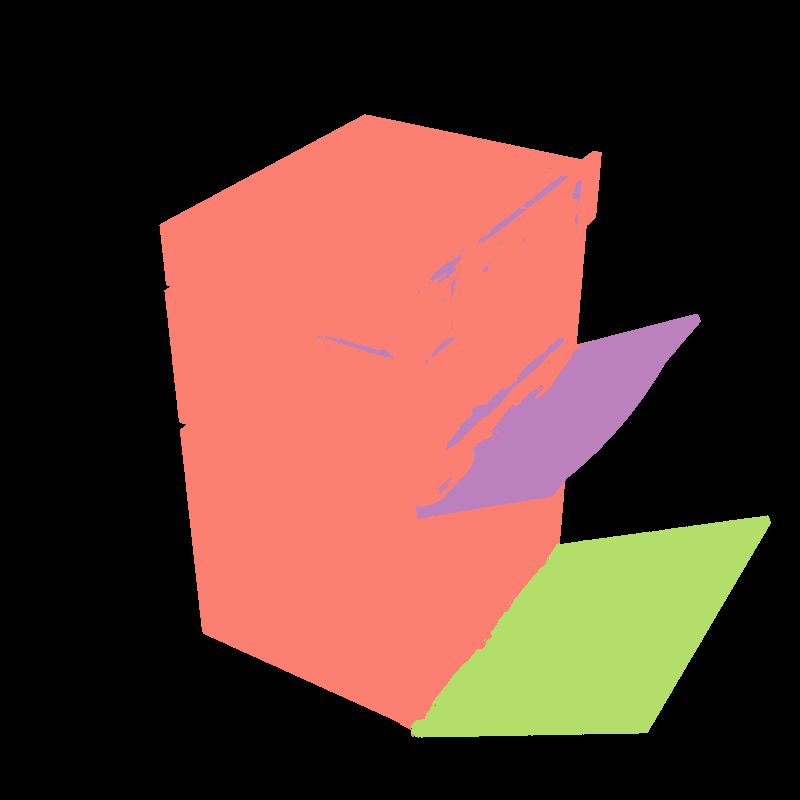}&
        \includegraphics[valign=c, width=0.11\columnwidth]{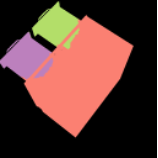}\\
        GT &
        \includegraphics[valign=c, width=0.11\columnwidth]{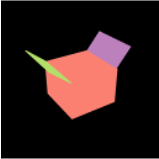}&
        \includegraphics[valign=c, width=0.11\columnwidth]{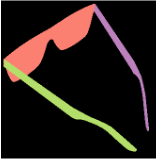}&
        \includegraphics[valign=c, width=0.11\columnwidth]{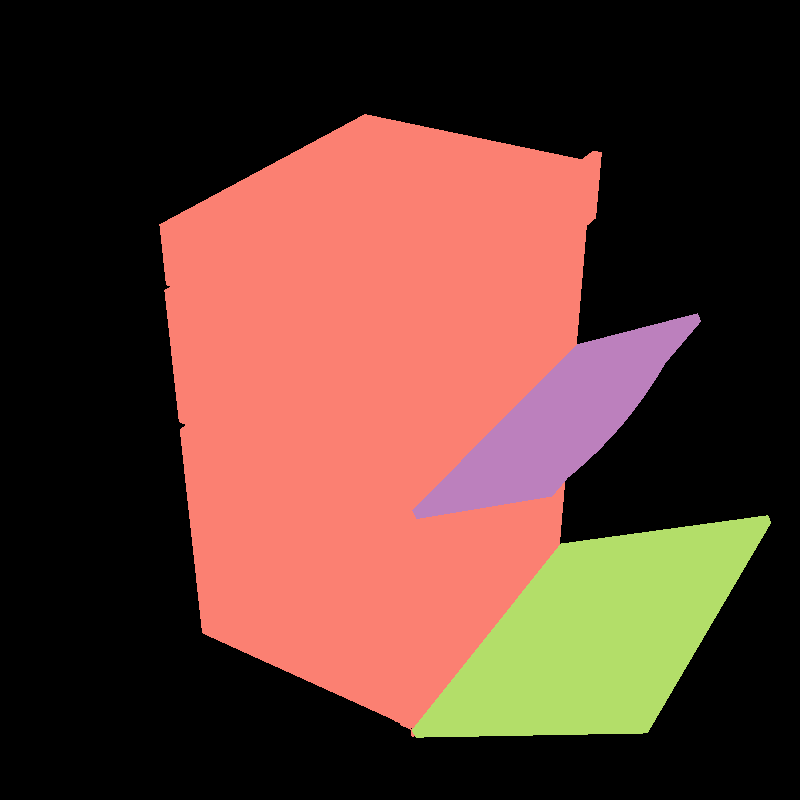}&
        \includegraphics[valign=c, width=0.11\columnwidth]{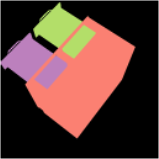}\\
       & 
       Box & Glasses & Oven & Storage\\
    \end{tblr}
    \caption{\textbf{Qualitative results for 2D multi-part segmentation.} The \textcolor{pink}{pink} color denotes the static part, while other colors denote the moving parts.\hb{add a bit explanation} }
    \label{fig:seg_multipart}

\end{wrapfigure}
\normalsize

\paragraph{Segmentation and composite rendering}

Part segmentation and novel view synthesis in pose $\mathcal{P}'$ are reported in \cref{tab:seg_psnr}, complemented by a qualitative analysis of part segmentation in \cref{fig:seg_comparison}. 
Our method outperforms PARIS across most evaluated objects in part segmentation and image synthesis quality, with the only exception being a minor difference in the PSNR for the laptop.

As our model builds on a static NeRF that achieves high quality novel view synthesis for one observation, the rendering quality is largely preserved after the second stage.
We provide more detailed analysis in \cref{sec:perfromance_comp_with_static}.
Additionally, benefiting from accurate pose estimation via a decoupled approach, our method achieves more robust and precise part segmentation, as depicted in \cref{fig:seg_comparison}. 
Here, our method consistently delivers accurate segmentation results for challenging objects such as the blade, stapler, and scissors, where PARIS struggles with accurate part reconstruction. 
In the other instances including the laptop, storage, and oven, our method achieves visibly better results.

\paragraph{Evaluation on objects with multiple movable parts}
A key advantage of our model is its ability to model objects with multiple moving parts. 
For such objects, we report pose estimation results in \cref{tab:multi_part}, qualitative for part segmentation in \cref{fig:seg_multipart} and novel articulation synthesis  in \cref{fig:qualitative}. 
As in the single part moving object experiments, our method performs consistently performs over the multi-part objects. 
Notably, we observed a marginally higher joint direction error for glasses, which we attribute to the thin structures such as temples and failure to segment them accurately which can be possibly improved by using higher resolution images.

\begin{table}[b!]
    \centering
    \begin{minipage}{0.60\linewidth}
    \centering
        \SetTblrInner{rowsep=2pt,colsep=2pt}
        \footnotesize
        \begin{tblr}{|c|c|c|c|c|c|c|c|}
        \hline
        \SetCell[r=2]{c}Metric & \SetCell[r=2]{c}{Method} &\SetCell[c=4]{c}{Revolut} &  & &  &\SetCell[c=2]{c}{Prismatic}&  \\
        \hline
            &   &laptop &oven &stapler &fridge &blade &storage \\
        
        \hline
        \SetCell[r=2]{c}{mIoU$\uparrow$} 
        
        &PARIS & 0.98 & 0.99 & 0.16 & 0.98& 0.76 & 0.94 \\
        
        &Ours & \textbf{0.99} & \textbf{0.99} & \textbf{0.98} & \textbf{0.99}& \textbf{0.94} & \textbf{0.96} \\
        
        \hline
        \SetCell[r=2]{c}{PSNR $\uparrow$} 
        
        &PARIS  & \textbf{30.31}   &31.48             &24.36         &32.74 &31.87 &30.63 \\
        
        &Ours  & 29.27             & \textbf{32.08}   &\textbf{34.31}     &\textbf{35.10} &\textbf{36.47} &\textbf{34.51} \\
        
        \hline
    \end{tblr}
    \caption{\textbf{Articulation synthesis and part segmentation results.} Average performance over 5 runs ( best results in \textbf{boldface}).}
    \label{tab:seg_psnr}
    \end{minipage}
    \hfill
    \begin{minipage}{0.38\linewidth}
    \centering
    \SetTblrInner{rowsep=1pt,colsep=1pt}
    \footnotesize
    \begin{tblr}{|c|c|c|c|c|}
    \hline
       \SetCell[r=2]{c}{Metric}  & \SetCell[c=3]{c}{Revolut} & & &Prismatic  \\
       \hline
         & oven & glasses & box & storage \\ 
     \hline
     $e_d$ $\downarrow$ &1.02 &2.35&0.56&1.82 \\
     \hline
    $e_p$ $\downarrow$ &0.16  & 0.47&0.27  & -\\
     \hline
     $e_g$ $\downarrow$ &1.03 & 1.01 &0.65& -\\
     \hline
     $e_t$ $\downarrow$ ($10$)& -& -&-&0.11 \\
     \hline
     PSNR $\uparrow$ & 32.98&29.22&28.61&28.25 \\
     \hline
     mIoU $\uparrow$ & 0.97 & 0.98 & 0.99 & 0.94 \\
     \hline
    \end{tblr}
    
    \caption{\textbf{Objects with multiple parts}. Errors using multiple metrics for pose estimation (averaged over all joints). }
    \label{tab:multi_part}
    \end{minipage}
\end{table}

\subsection{Ablation studies}


We assessed the effectiveness of the proposed decoupled pose estimation of $M_\ell$ (DP) and iterative refinement of $X_\ell$ (IR) on the `fridge' object. 
When DP is disabled, the segmentation and articulation are simultaneously learned. 
In contrast, disabling IR maintains the initial $X_\ell$ for pose estimation. 
The results in \cref{tab:ablation_component}, particularly the first row, demonstrate that joint optimization without DP inaccurately predicts the articulated pose, treating the entire object as static. While enabling DP improves the performance, as shown in the second row, the performance is still poor due to the noisy initial values compared to our full model.

Additionally, we evaluated the impact of the number of views in the target observation for training. We train our model on the randomly subsampled views for multiple runs and averaged their performance (see \cref{tab:ablation_frames}). 
Results show a significant drop in pose estimation and rendering quality with 4 images or less. 
Notably, even with only 8 images, our approach surpasses the performance of PARIS trained with 100 images (indicated with subscript $P$ in \cref{tab:ablation_frames}). This result clearly show that our method are more robust against fewer viewpoints from the target viewpoints and allows efficiently learning `articulate' a pretrained NeRF from few target views only.


\begin{table}[h!]
    \centering
    \begin{minipage}{0.58\columnwidth}
        \centering
        \SetTblrInner{rowsep=1pt,colsep=2pt}
    \footnotesize
            \begin{tblr}{|c|c|c|c|c|c|c||c|}
                \hline
                 \SetCell[r=2]{c}Metric&\SetCell[c=7]{c}Num. of images & & & & & &\\
                   \hline
                   &2 & 4 & 8 & 16 & 32 & 100 & 100$_{P}$\\ 
                   \hline
                 $e_d$ $\downarrow$ & 46.05 &8.89 &0.59	&0.58	&0.50	&0.54 &0.81\\
                 \hline
                 $e_g$ $\downarrow$ & 44.74 &20.79 &0.70	&0.60	&0.56 &0.49&0.87\\ 
                 \hline
                 PSNR $\uparrow$ & 22.65&29.95 &34.00 &34.28	&34.88 &35.10 &32.74\\
                 \hline
                \end{tblr}
    \caption{\textbf{Ablation studies with different number of target images.} }
    \label{tab:ablation_frames}
    \end{minipage}
    \hfill
    \begin{minipage}{0.38\columnwidth}
        \centering
        \SetTblrInner{rowsep=2pt,colsep=2pt}
        \footnotesize
        \begin{tblr}{|c|c|c|c|c|}
            \hline
            \SetCell[c=2]{c} Init.& &\SetCell[c=3]{c} Metric& & \\
            \hline
            DP & IR & $e_d$ & $e_g$ & PSNR \\
            \hline
            - & -  & 2.57 & 26.95 & 18.09 \\
            \hline
            \checkmark & - & 2.38 & 14.96 & 25.16\\
            \hline
            \checkmark & \checkmark & 0.54 & 0.49 & 35.10 \\
            \hline
        \end{tblr}
        \caption{\textbf{Ablation study over different initialization strategies.} }
        \label{tab:ablation_component}
    \end{minipage}
\end{table}

\subsection{Real-world example}

\begin{figure}
    \centering
    \resizebox{\columnwidth}{!}{
    \footnotesize
    \begin{tblr}{ccccc|ccccc}
        \includegraphics[width=0.13\columnwidth]{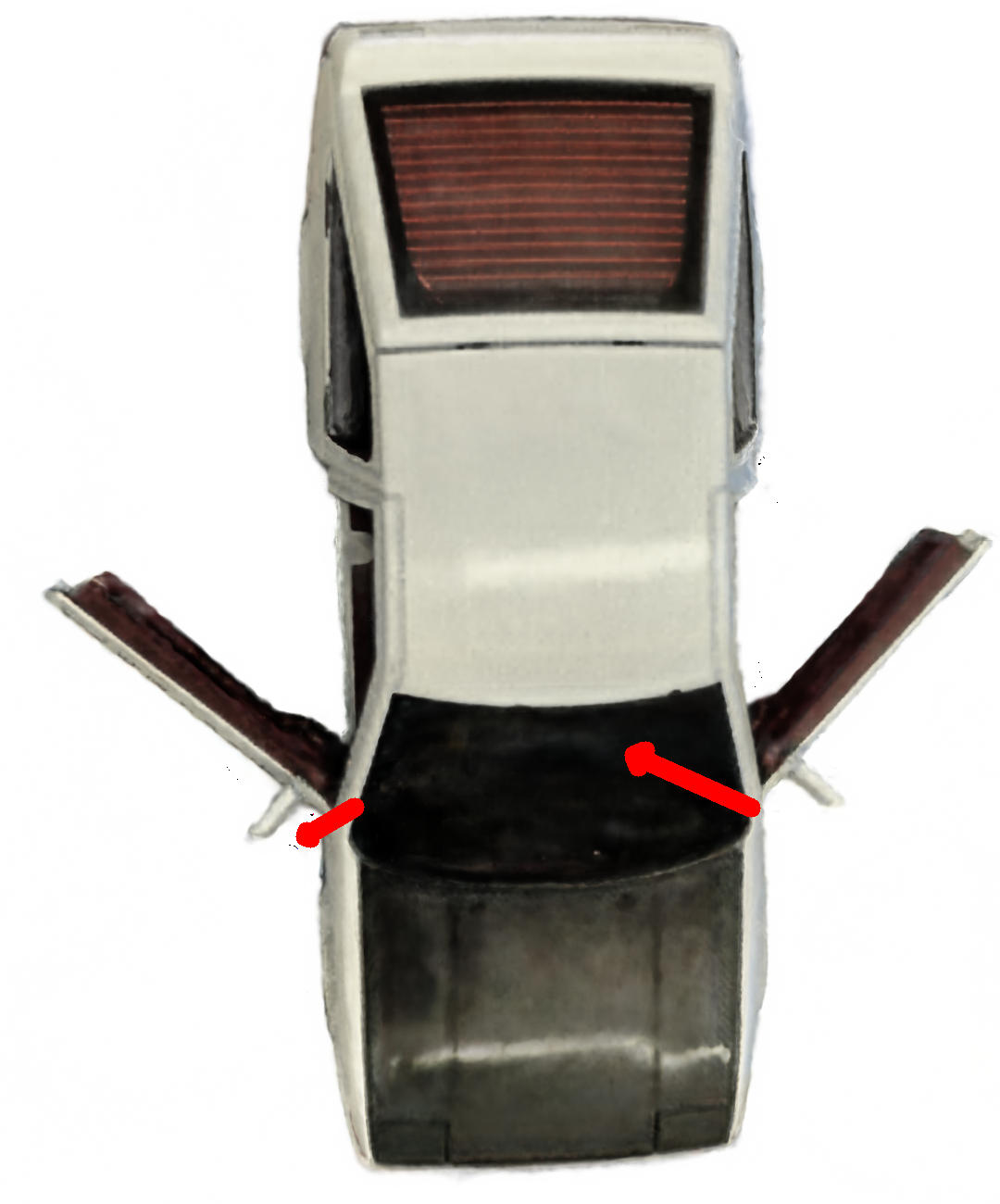}&
        \includegraphics[width=0.13\columnwidth]{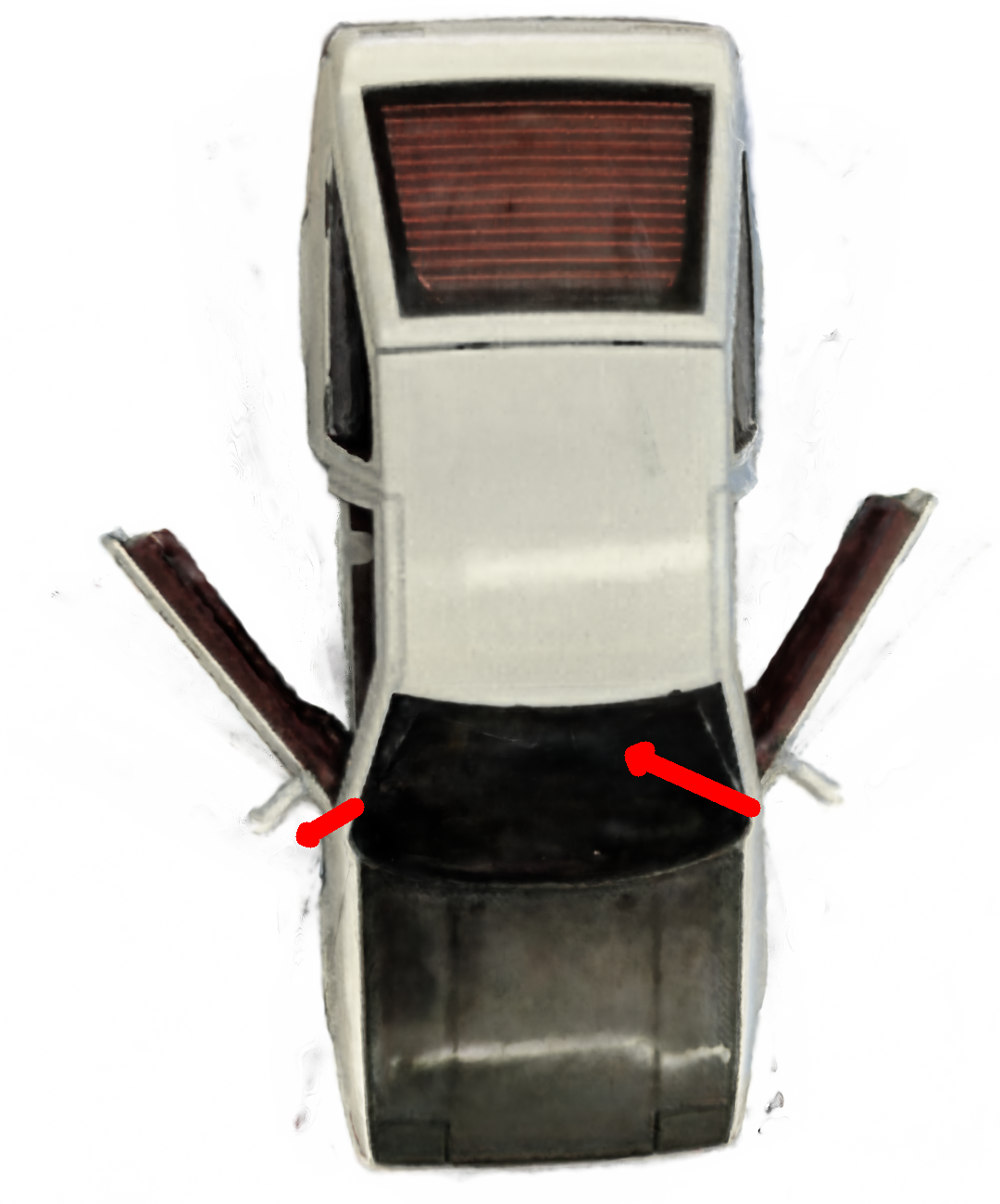}&
        \includegraphics[width=0.13\columnwidth]{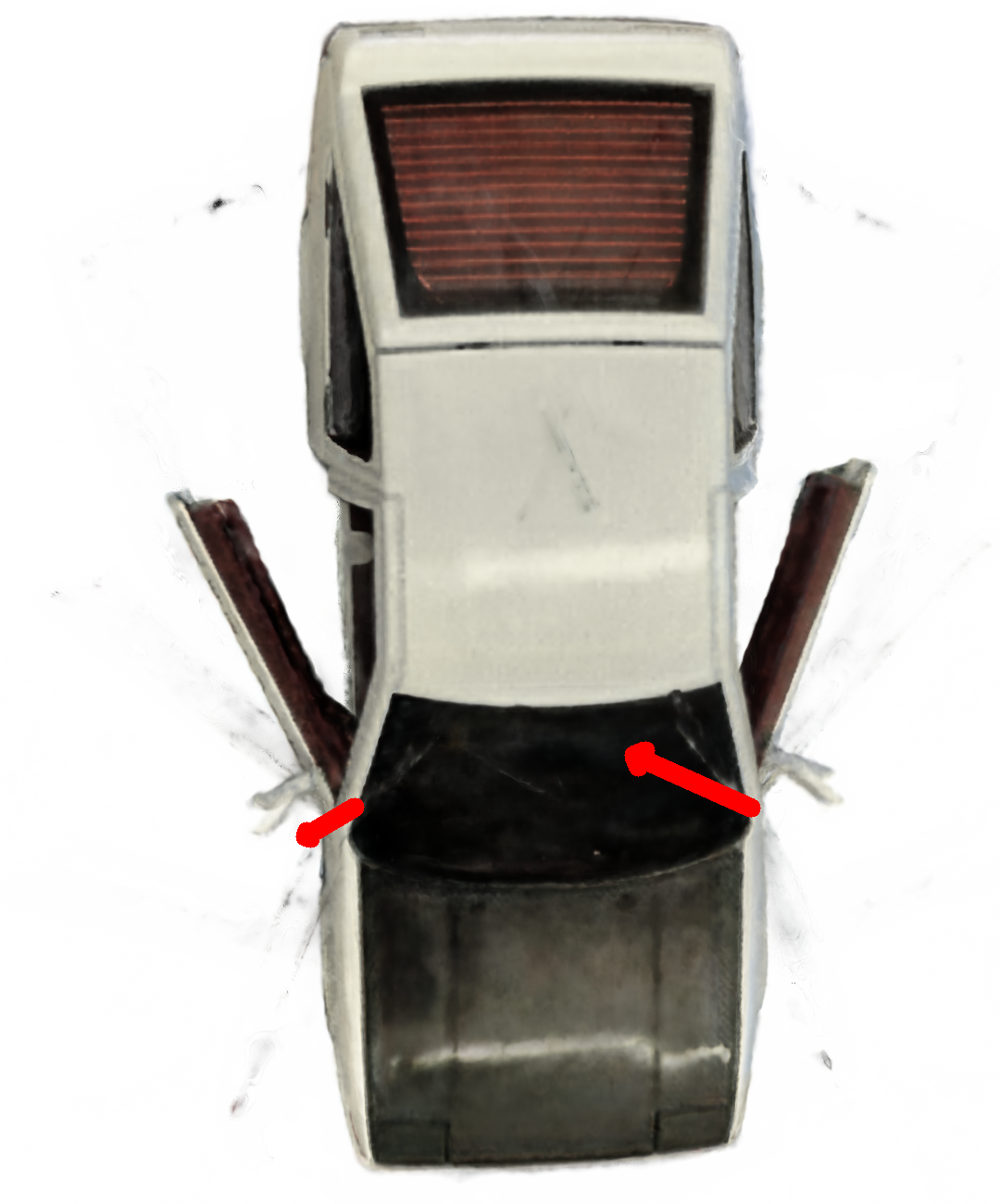}&
        \includegraphics[width=0.13\columnwidth]{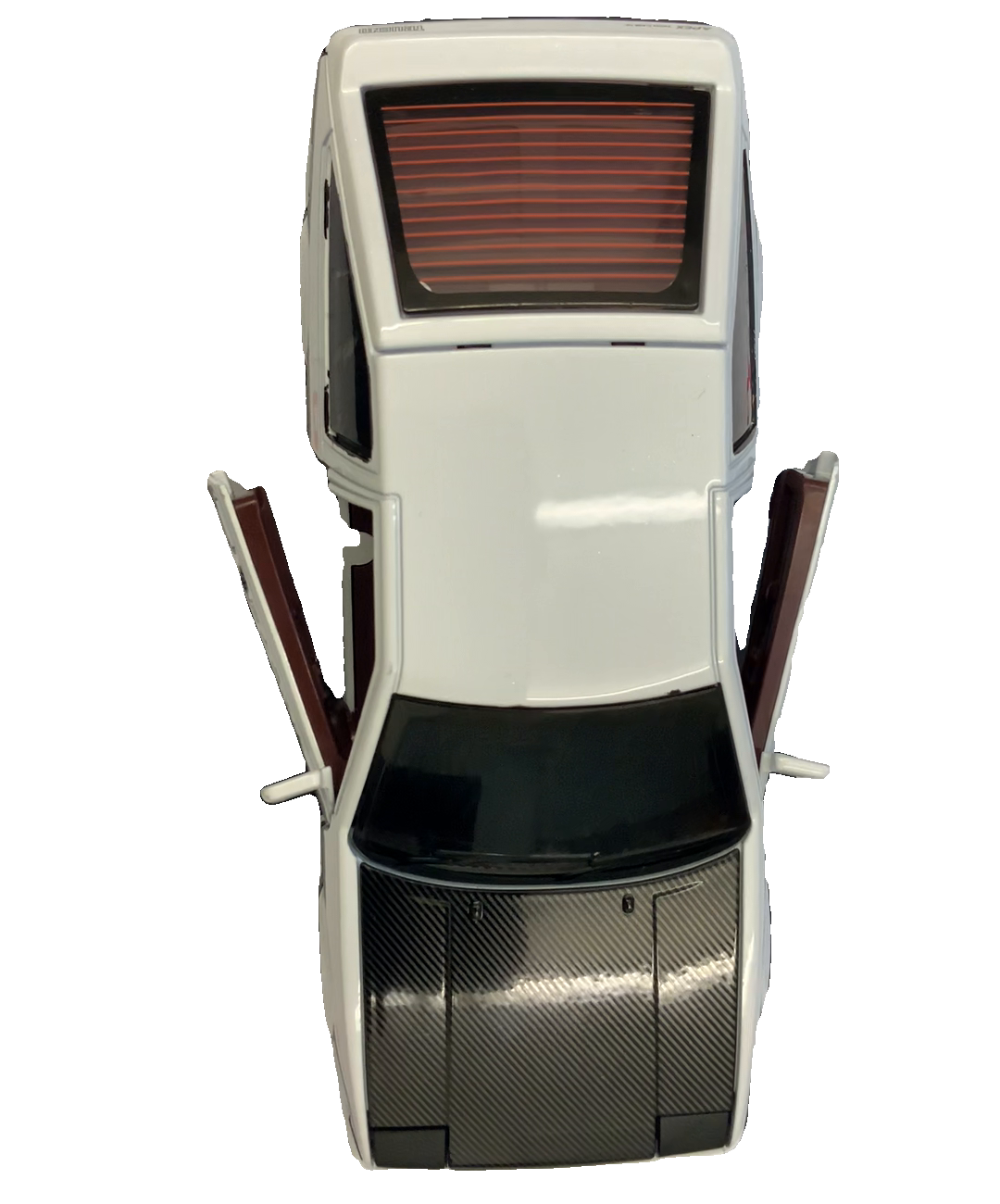}&
        \includegraphics[width=0.13\columnwidth]{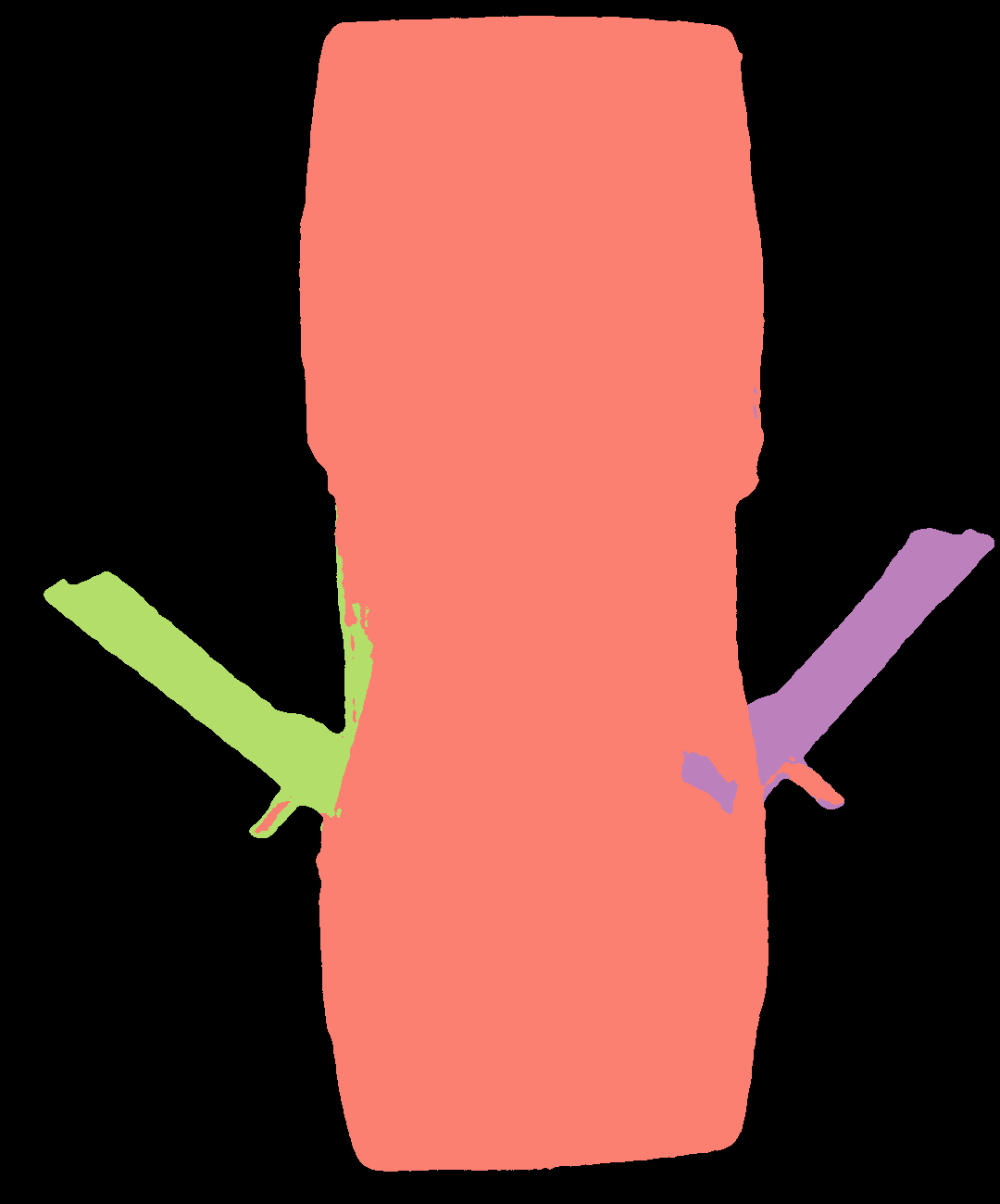}
        &
        \includegraphics[width=0.13\columnwidth]{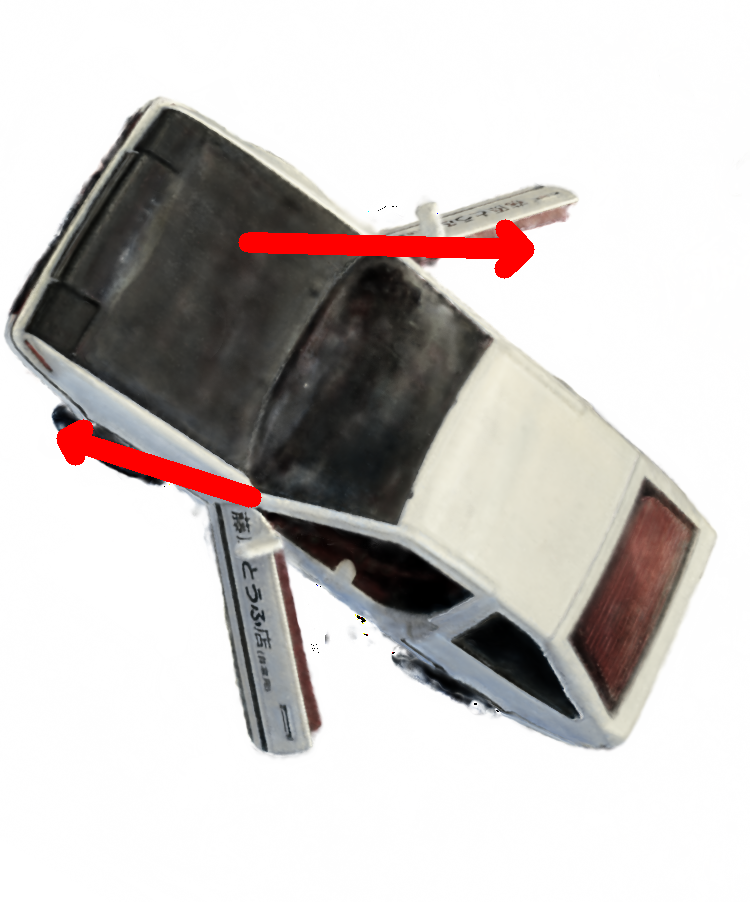}&
        \includegraphics[width=0.13\columnwidth]{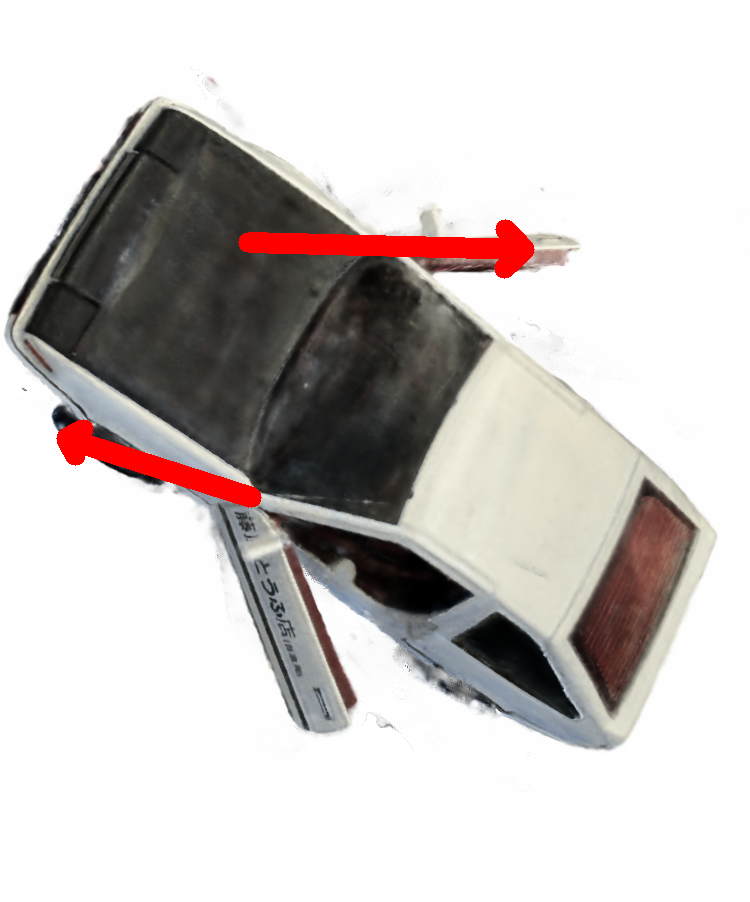}&
        \includegraphics[width=0.13\columnwidth]{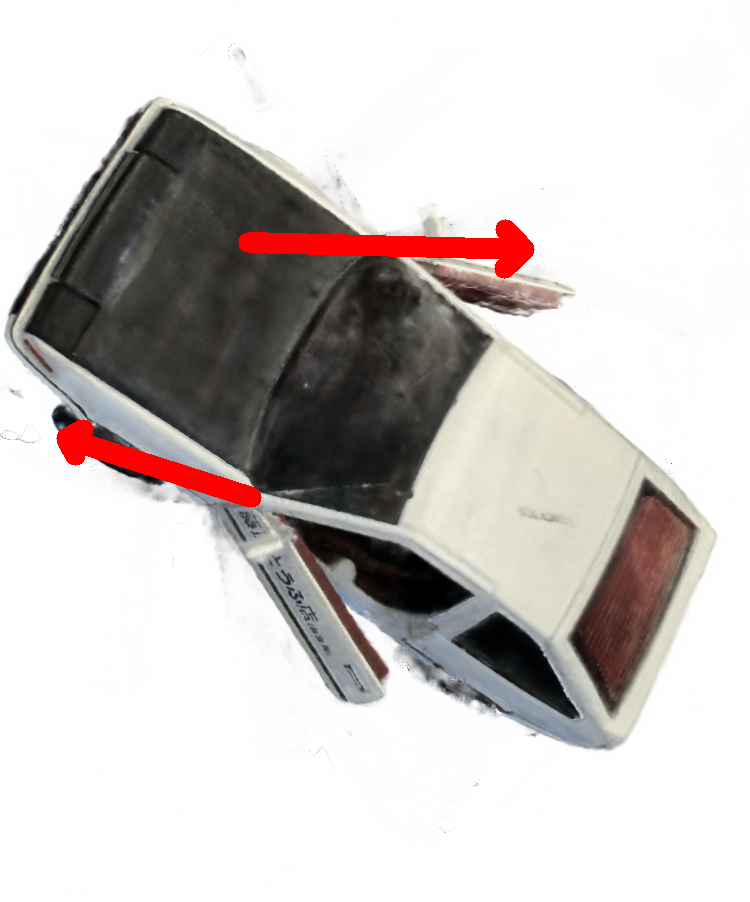}&
        \includegraphics[width=0.13\columnwidth]{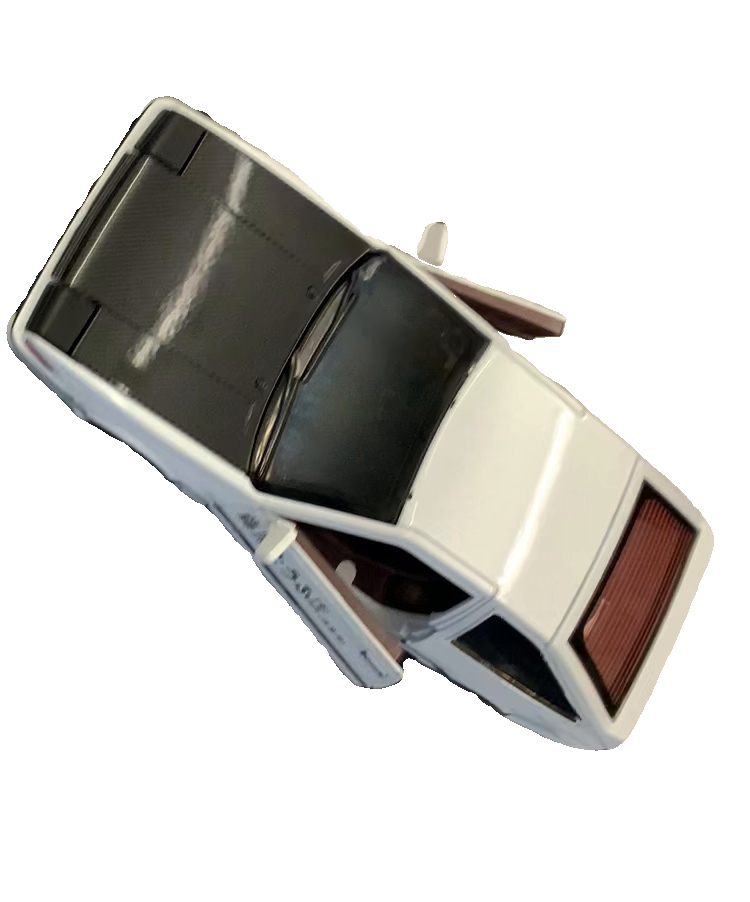}&
        \includegraphics[width=0.13\columnwidth]{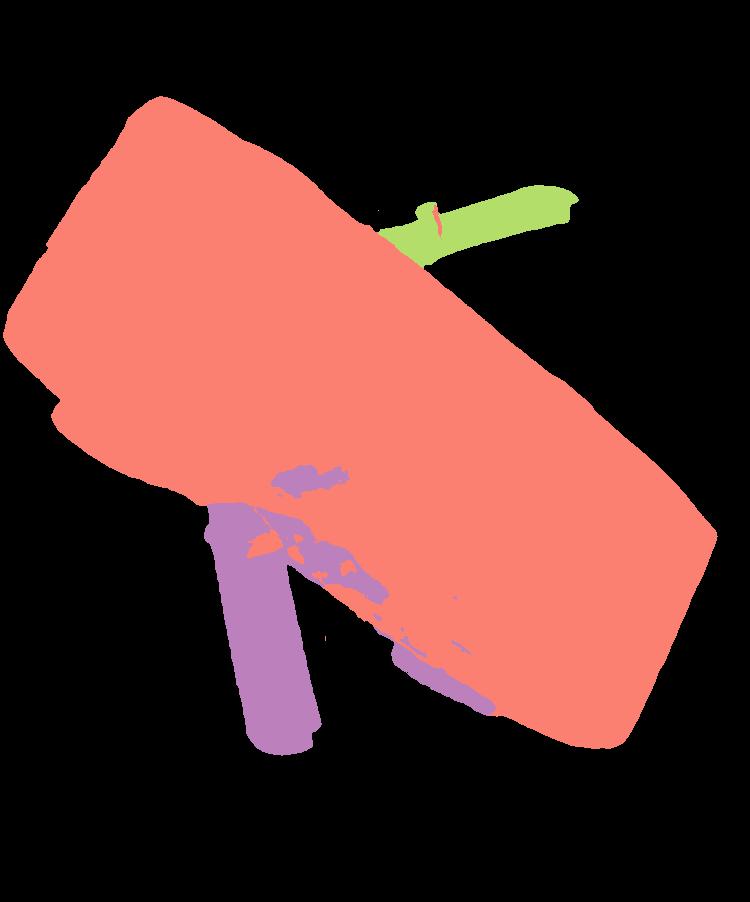}\\
        \SetCell[c=3]{c}{Articulation interpolation}& & & GT & Seg. & \SetCell[c=3]{c}{Articulation interpolation} & && GT & Seg.
        
    \end{tblr}
    }
    \caption{Results on real world examples, the \textcolor{red}{red line} indicates the estimated joint axis direction. \textcolor{green}{Green} and \textcolor{purple}{purple} color denotes the moving car door, while \textcolor{pink}{pink} denotes the body of the toy car. Please refer to \cref{fig:appendix_real_vis} for more qualitative evaluation.}
    \label{fig:realworld_examples}
\end{figure}

The qualitative evaluation can be found in \cref{fig:realworld_examples}. The pose $\mathcal{P}$ for training the static NeRF is with two doors wide open, and the reconstruction results from NeRF is shown in the first row in \cref{fig:realworld_examples}. Qualitative comparison for ground truth images in pose $\mathcal{P}'$ is presented in second and third row of \cref{fig:realworld_examples}.  The average PSNR we obtain from the static NeRF is , and the PSNR for 28.19, while the PSNR for pose $\mathcal{P}'$ is 24.32. More detail results for segmentaion and articulation interpolation are also presented in \cref{fig:realworld_examples}.

\subsection{Limitations}
Our method has few limitations too.
As we use rendering to supervise our segmentation and pose, our method may fail to segment very thin parts when the rendering error is small.
When the object parts are nearly symmetry, our method may fail to find the correct articulation and choose another articulation that produces similar rendering.
We provide examples of such failure cases in \cref{sec:limitations}.
Additionally, it also inherits the limitations of the implicit models that are built on such as failure to model transparent objects, and inaccurate 3D modeling when the viewpoint information is noisy.
Finally, our method is limited to articulated objects with rigid parts and cannot be used to model soft tissue deformations.

%% file: sections/05conclusion.tex
\section{Conclusion}

In this study, we tackled the challenges of self-supervised part segmentation, articulated pose estimation, and novel articulation rendering for objects with multiple movable parts. Our approach is the first known attempt to model multipart articulated objects using only multi-view images acquired from two arbitrary articulation states of the same object. We evaluated our method with both synthetic and real-world datasets. The results suggest that our method competently estimates part segmentation and articulation pose parameters, and effectively renders images of unseen articulations, showing promising improvements over existing state-of-the-art techniques. Further, our real-world data experiments underscore the method's robust generalization capabilities. At last, the code and the data used in this project will be released upon acceptance.
However, the reliance on geometric information from moving parts for articulation pose estimation poses challenges in modeling highly symmetrical objects. Future work could improve our method by incorporating both appearance and geometric data into the pose estimation process, potentially enhancing accuracy and applicability.

\paragraph{Broader Impacts} The proposed technique could be potentially used to understand articulated objects for their robotic manipulation in future. The authors are not aware of any potential harm that may arise when the technology is used

%% file: sections/08supp.tex
\appendix

\section{Appendix / supplemental material}
Optionally include supplemental material (complete proofs, additional experiments and plots) in appendix.
All such materials \textbf{SHOULD be included in the main submission.}

\subsection{Part-aware proposal network}
\label{sec:part_proposal}
To achieve part-aware composite rendering for articulated objects, we also need to modify the proposal network for correct sampling along the ray $r \in \mathcal{R}'$. In the composite rendering, the proposal network is now required to produce the similar distribution of the weights for samples along the ray. Thus, following the similar design in the part-aware NeRF, we append an extra segmentation field to the proposal network. Now the valid density for each samples becomes segmentation-mask density as $s_\ell^{r_\ell}(\bm x_j)\sigma^{r_\ell}$. The transmittance are then calculated using \cref{eqn:color_composite} without the color term, which will be later used for online distillation. Please refer to the original paper \cite{barron2022mip} for more details about the online distillation. During the training of part-aware NeRF, the original parameters in the proposal network will be frozen and only updates the segmentation field. 

\subsection{Comparison with PARIS}

Here we also provide the 5-time best comparison between ours and PARIS. Besides, the reported performance for PARIS is also provided in this subsection. From \cref{tab:comparison_5_time_best}, we can see that the original reported performance from PARIS are exceptionally better compared to the reproduced ones, even though with the same data and the same configurations. If we only focus on the 5-time best comparison between reproduced PARIS and ours, we can see that we achieve comparable performance. Another thing we can notice is that for prismatic objects, while PARIS has much better estimations on joint axis direction. However, our method later show much better estimation for estimating the moved distance for the dynamic parts. While one thing to notice is that PARIS failed to reconstruct the stapler for all 5 runs in our experiment. 

\begin{table}[t!]
    \centering
    \SetTblrInner{rowsep=2pt,colsep=2pt}
    \normalsize
    \begin{tblr}{|c|c|c|c|c|c|c|c|}
        \hline
        \SetCell[r=2]{c}Metric & \SetCell[r=2]{c}{Method} &\SetCell[c=4]{c}{Revolut} &  & &  &\SetCell[c=2]{c}{Prismatic}&  \\
        \cline{3-12}
            &   &laptop &oven &stapler &fridge &blade &storage \\
        \hline
        \SetCell[r=3]{c}{$e_d$ $\downarrow$} 
        &PARIS$^\dagger$  & $\bm{0.03}$ & $\bm{0.03}$ & $\bm{0.07}$ & $\bm {0.00}$ & $\bm {0.00}$ & \underline{0.37} \\
        &PARIS  & \underline{0.18} & 0.48& 1.30 & \underline{0.22}& \underline{0.27} & $\bm {0.24}$\\
        &Ours  &0.26 &\underline{0.28} &\underline{0.27} &0.42 &1.41 &0.86\\
        \hline
        \SetCell[r=3]{c}{$e_p$$\downarrow$\\ ($10^{-2}$)} 
        &PARIS$^\dagger$ & $\bm{0.01}$ & $\bm{0.03}$ & $\bm{0.06}$ & $\bm{0.02}$& - & -\\
        &PARIS & \underline{0.08} & \underline{0.04} & 2.12 & \underline{0.17} & - & -\\
        &Ours & 0.39 & 1.24 & \underline{0.13} & 0.41 & - & - \\
        \hline
        \SetCell[r=3]{c}{$e_g$$\downarrow$}
        &PARIS$^\dagger$   &$\bm{0.03}$ & $\bm{0.00}$ &$ \bm{0.00}$ & $\bm{0.00}$ & - & - \\
        &PARIS &0.24 & \underline{0.25} &37.86 & \underline{0.37} & - & - \\
        &Ours         &\underline{0.22}  &0.31  &\underline{0.24}     &0.55  & - & - \\
        \hline
        \SetCell[r=3]{c}{$e_t$ $\downarrow$ ($10^{-1}$)} 
        &PARIS$^\dagger$ & - & - & - & -  &$\bm{0.06}$ & $\bm{0.00}$\\
        &PARIS & - & - & - & -  &5.84 &2.93\\
        &Ours & - & - & - & -  &\underline{0.12} &\underline{0.05}\\
        
    \hline
    \end{tblr}
    \caption{Comparison with PARIS in 5-time best setting. PARIS$^\dagger$ denotes performance reported in the original paper. Best results are shown in \textbf{bold}, second best are shown with \underline{underline}.}
    \label{tab:comparison_5_time_best}
\end{table}

\subsection{Limitations}
\label{sec:limitations}

Details about the limitations and failure cases of our methods.

\begin{figure}
    \centering
    \begin{subfigure}{0.9\linewidth}
        \includegraphics[width=0.31\linewidth]{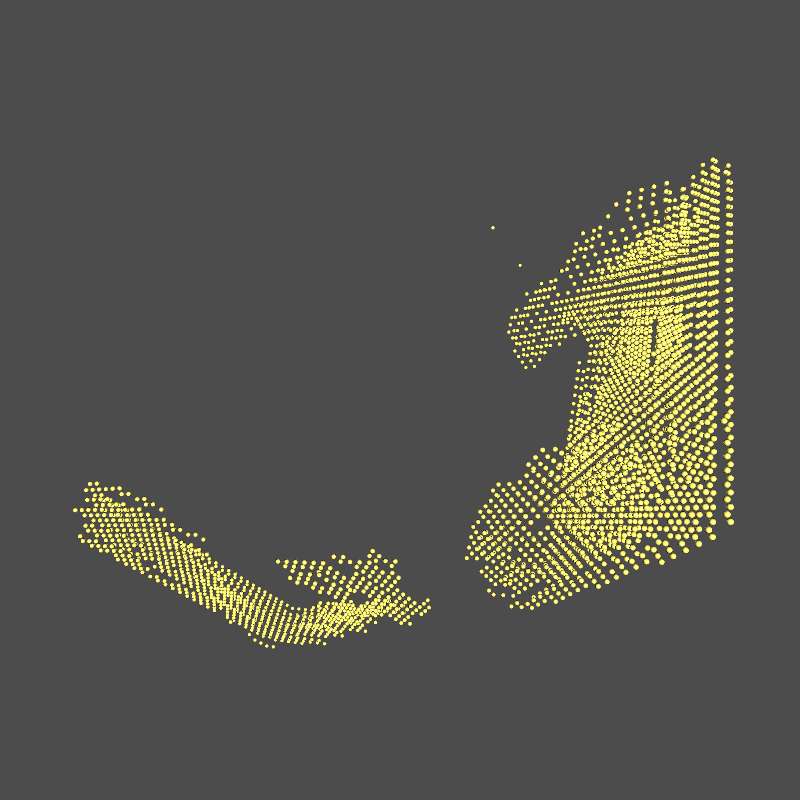}
        \includegraphics[width=0.31\linewidth]{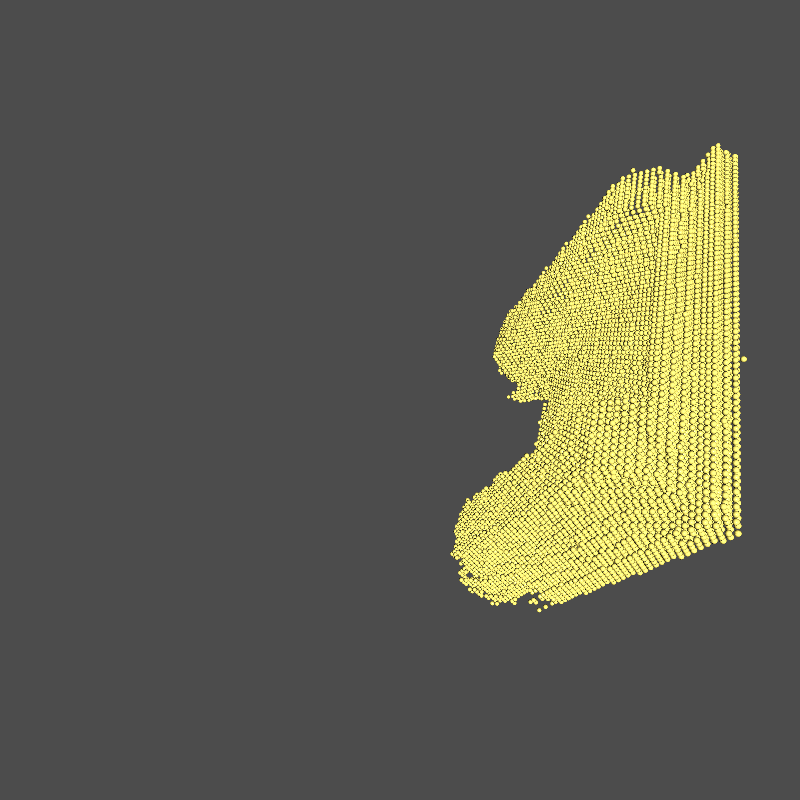}
        \includegraphics[width=0.31\linewidth]{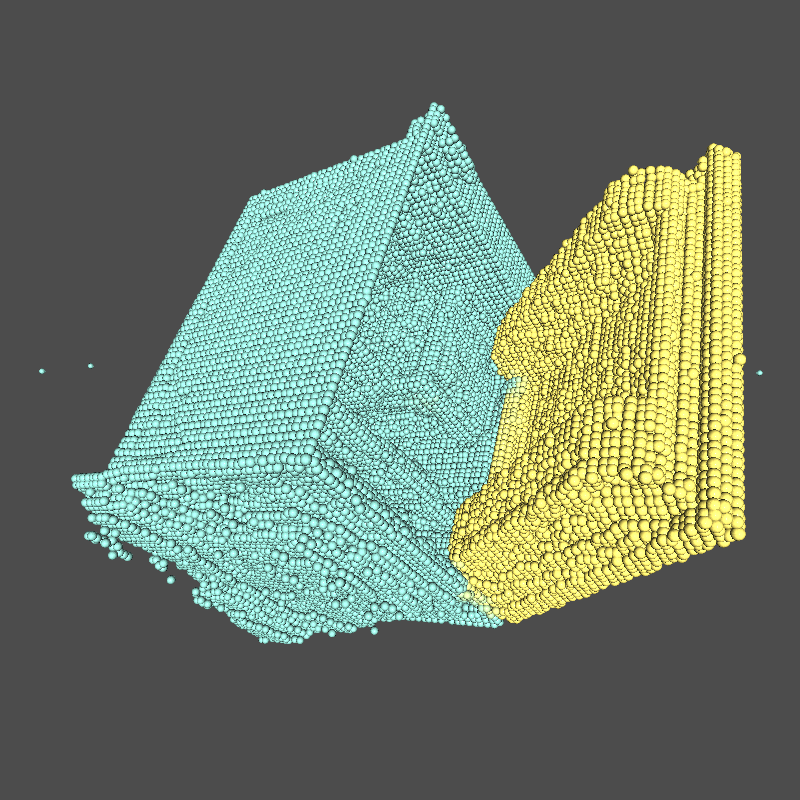}
        \label{fig:DP}
        \caption{Visualization of dynamic voxel update. From left to right: initialized $X_\ell$, updated $X_\ell$, final $X_\ell$}
    \end{subfigure}
    \hfill
    \begin{subfigure}{0.9\linewidth}
        \includegraphics[width=0.31\linewidth]{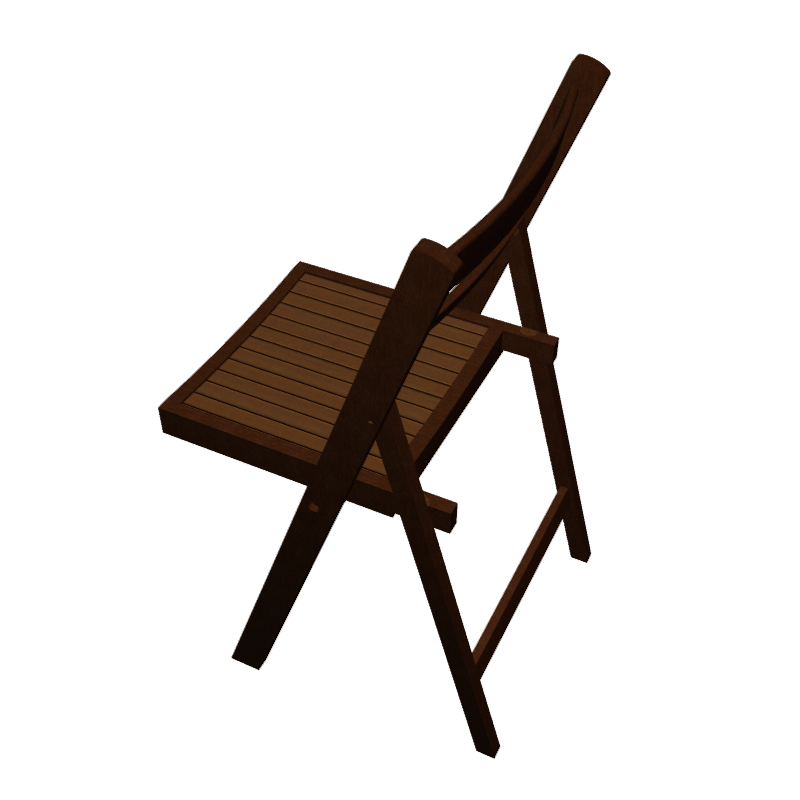}
        \includegraphics[width=0.31\linewidth]{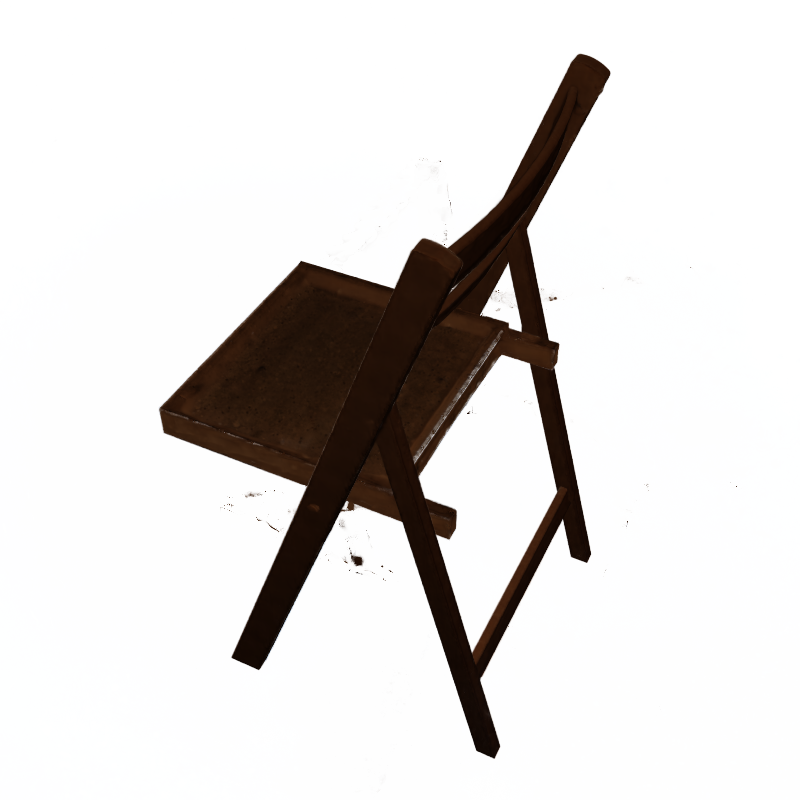}
        \includegraphics[width=0.31\linewidth]{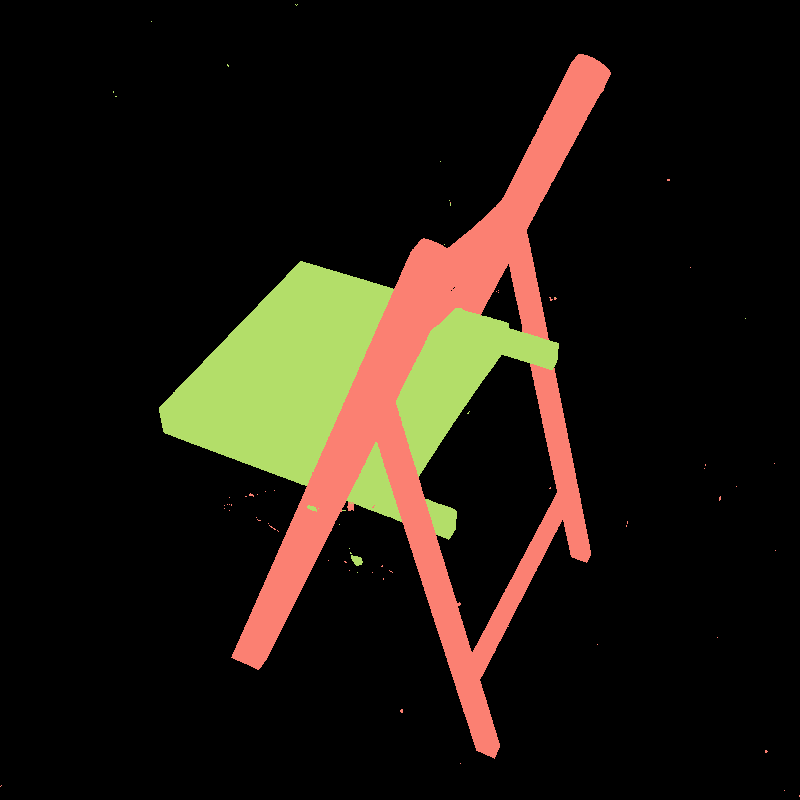}
        \caption{Failure cases for foldchair, from left to right: groundtruth RGB, rendered RGB, part segmentation.}
        \label{fig:limitation_foldchair}
    \end{subfigure}
    \hfill
    \begin{subfigure}{0.9\linewidth}
        \includegraphics[width=0.45\linewidth]{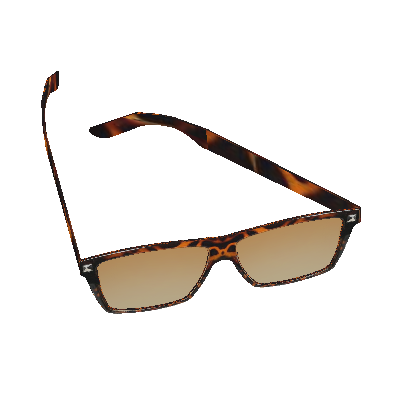}
        \includegraphics[width=0.45\linewidth]{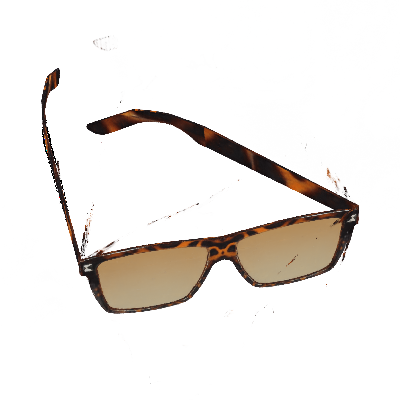}
        \caption{Artifacts for thin parts, the left one is the groundtruth, the right rendering result.}
        \label{fig:limitation_glasses}
    \end{subfigure}
    \label{fig:limitation}
\end{figure}

In our decoupled optimization strategy, the articulated pose estimation relies exclusively on the geometric information from the moving parts. As a result, our method faces challenges with highly symmetrical components. For example, in the case of a folding chair \cref{fig:limitation}, although the segmentation accurately identifies the chair's seat, the pose estimation mistakenly flips the seat by 180 degrees, resulting in the seat being oriented with the bottom upwards as shown in \cref{fig:limitation_foldchair}. Additionally, our method may encounter difficulties when the articulation motion is minimal, which can lead to insufficient pixel differentiation in images across different articulation poses for initial model estimation. In such cases, both the pose estimator and the iterative updates of $X_\ell$ may underperform or even fail. We also notice that the proposal network have difficulties handling very thin parts in the novel articulation synthesis. As we can see in \cref{fig:limitation}, some artifacts appeared at the edge of the frame and the temples for the glasses as in \cref{fig:limitation_glasses}, which should be ideally masked out without any opacity and color for those pixels.

Furthermore, the reliance on a pre-trained static NeRF limits the upper boundary of our method's performance in rendering novel articulations. Addressing these limitations could significantly enhance the robustness and applicability of our strategy in handling complex articulated objects. For furhter analysis of this, please refer to \label{sec:perfromance_comp_with_static}.

\subsection{Training setting}
\label{sec:training_settings}
As detailed in \cref{sec:optimization}, we perform optimization on $M_\ell$ for 4000 iterations and on $s_\ell$ for 2000 iterations for all evaluated objects. Checkpoints are saved following Step 3. We conduct this process through 5 cycles for objects with a single moving part and 6 cycles for objects with multiple moving parts. Checkpoints with best PSNR during validation will be used for test.

During the first step of $M_\ell$ optimization, we begin with a learning rate of 0.01, which linearly decays by 0.5 every 500 iterations. We accumulate gradients from 8 viewpoints to simulate a batch size of 8, initializing $M_\ell$ identically in the first cycle and using the previously estimated $M_\ell$ for subsequent cycles.

In the second step of $s_\ell$ optimization using the Adam optimizer, the initial learning rate is set at 0.01 and linearly decays by a factor of 0.01 every 100 iterations. For multiple moving parts, initialization involves training the segmentation head $s$ using pre-assigned labels on $X_\ell$, and querying predictions for $\bm x \in X_\ell$. Cross-entropy loss optimized over 1000 iterations with a learning rate of $1e^{-3}$ shapes the learnable parameters in $s$.

Our experiments, requiring around 16 GB of VRAM, complete in approximately 30 minutes on a single RTX 4090 GPU for a single object. As for the training of static NeRF, it takes about 10 minutes with less than 10 GB of VRAM. The estimated total GPU time for this project would be about 2 GPU months.

\subsection{Ablations}

\subsubsection{Performance cost for novel articulation synthesis}
\begin{figure}
    \centering
    \begin{subfigure}{0.3\linewidth}
        \includegraphics[width=\linewidth]{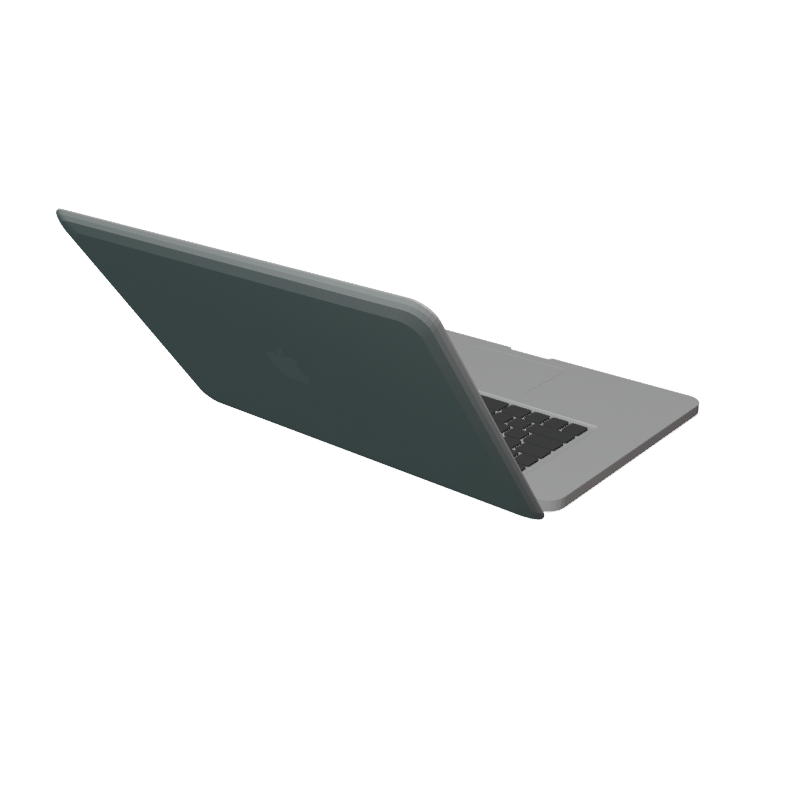}
        \caption{Ground truth}
    \end{subfigure}
    \begin{subfigure}{0.3\linewidth}
        \includegraphics[width=\linewidth]{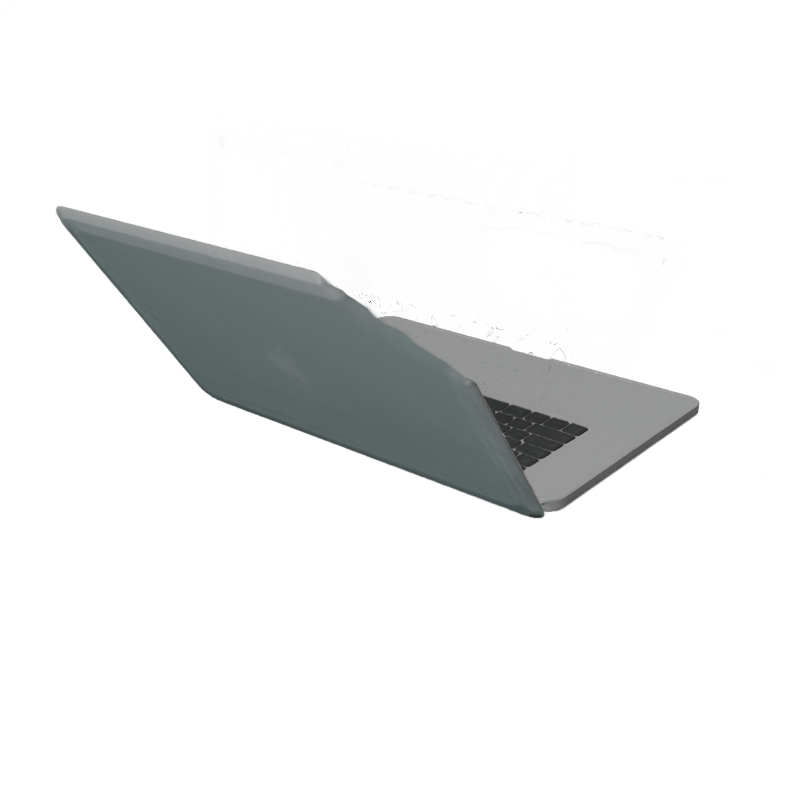}
        \caption{Rendered}
        \label{fig:laptop_fail}
    \end{subfigure}
    \begin{subfigure}{0.3\linewidth}
        \includegraphics[width=\linewidth]{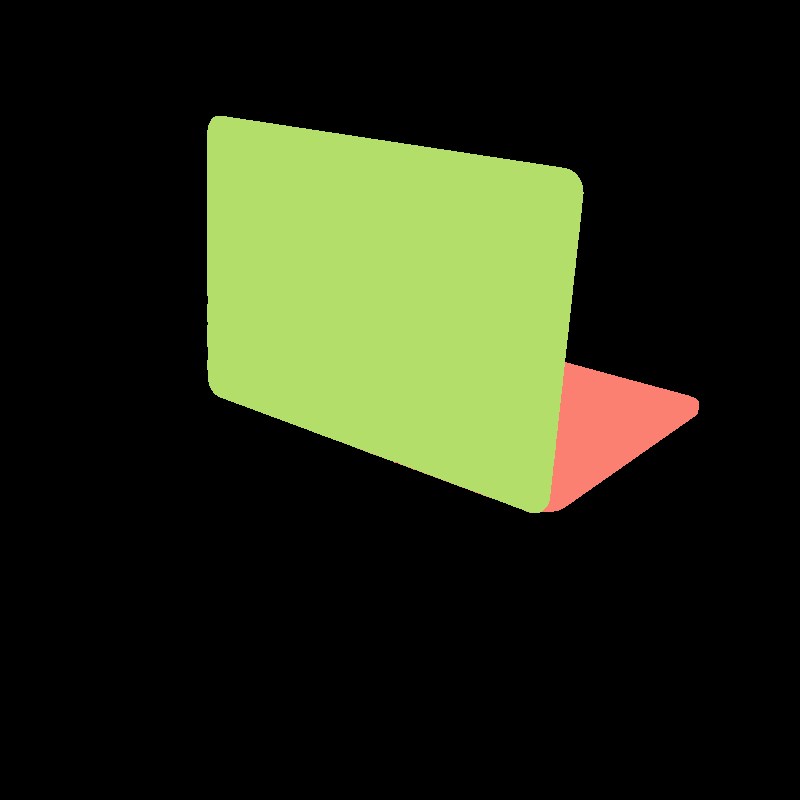}
        \caption{Segmention in original pose}
    \end{subfigure}
    \caption{We can see in the \cref{fig:laptop_fail} that the corner of the laptop screen is missing in the novel articulation rendering. While it looks perfect when we check the segmentation in the original pose. Thus, we suspect it is the proposal network than failed to estimate the density distribution for the screen from certain viewpoints. }
    \label{fig:lap_top_explain}
\end{figure}

We also provide an ablation study to investigate the performance drops for conditional novel articulation synthesis compared to the pre-trained static model. As shown in \cref{tab:ablation_performance}, the performance drop is subject to different category of objects ranging from smallest 13.3\% for the fridge and blade to 31.4\% laptop. We suspect the significant drops for laptop is caused by the deteriorate performance of the proposal network on the thin laptop screen. Visualizations can be found in \cref{fig:lap_top_explain}. The results indicate that our method can benefit from the high quality of appearance reconstruction from the pre-trained static NeRF. 

\begin{table}[h!]
    \centering
    \resizebox{\linewidth}{!}{
    \begin{tblr}{|c|c|c|c|c|c|c|c|c|c|c|}
    \hline
        \SetCell[r=2]{c} Method &\SetCell[c=10]{c}{Object} & &&&&&&&&  \\
        \hline
         & laptop & oven & stapler & fridge & blade & storage & oven$^*$ & glasses$^*$ & box$^*$ & storage$^*$  \\ 
         \hline
         Static & 42.70 & 39.69 & 42.39& 40.50 & 42.7& 40.33& 39.03& 37.93&36.56& 35.23\\
         \hline
         Art. & 29.27 & 32.08 & 34.31& 35.1& 36.47& 34.51& 32.98& 29.22&28.61& 28.25\\
         \hline
         $\Delta$& -31.4\% & -19.2\% & -19.1\% & -13.30\%& -13.3\%& -14.4\% & -15.5\% & -22.8\% & -21.8\%& -19.8\% \\
         \hline
    \end{tblr}
    }
    \caption{Comparison of rendering quality between objects in their original pose $\mathcal{P}$ and articulated pose $\mathcal{P}'$. 'Static' refers to the rendering performance of the object in its original pose $\mathcal{P}$, whereas 'Art.' indicates the rendering quality of the object in articulated pose $\mathcal{P}'$ using our method with the static NeRF. Objects marked with $^*$ represent those with multiple movable parts.}
    \label{tab:ablation_performance}
\end{table}

\subsection{Visualization}

Here we show more visualization for qualitative evaluations. Besides, additional animated images for the real-world toy car is also provided in the supplementary materials.

\begin{figure}
    \centering
    \resizebox{\columnwidth}{!}{%
    \SetTblrInner{rowsep=0pt,colsep=1pt}
    \scriptsize
    \begin{tblr}{cc|ccccc|c}
    {Method} &{ GT in $\mathcal{P}$} & \SetCell[c=5]{c}{ Novel articulation synthesis}& &&&& { GT in $\mathcal{P}'$}
        \\
        PARIS &
        \includegraphics[valign=c, width=0.1\columnwidth]{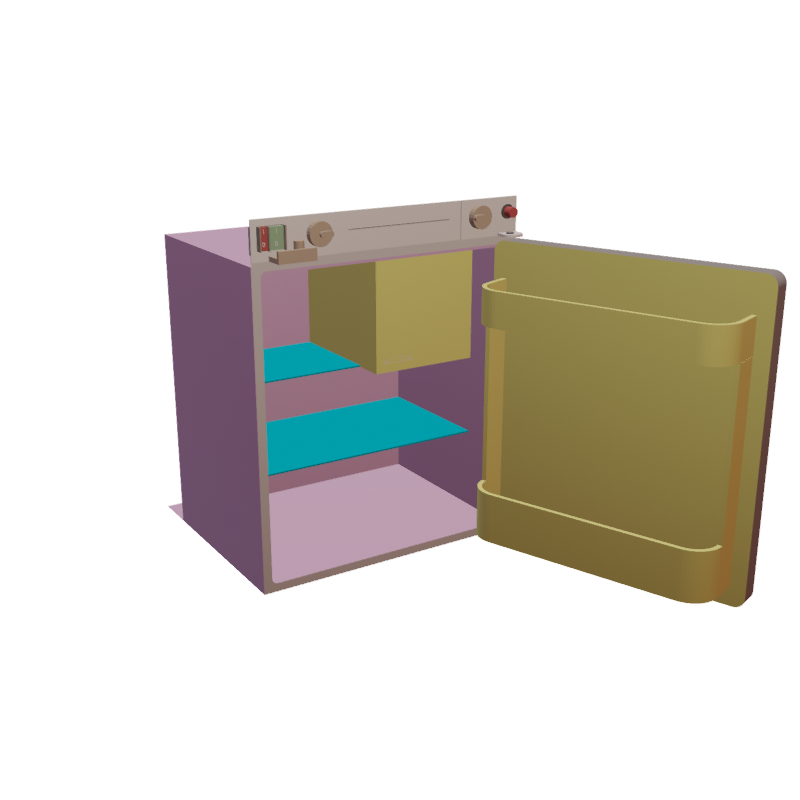}&
        \includegraphics[valign=c, width=0.1\columnwidth]{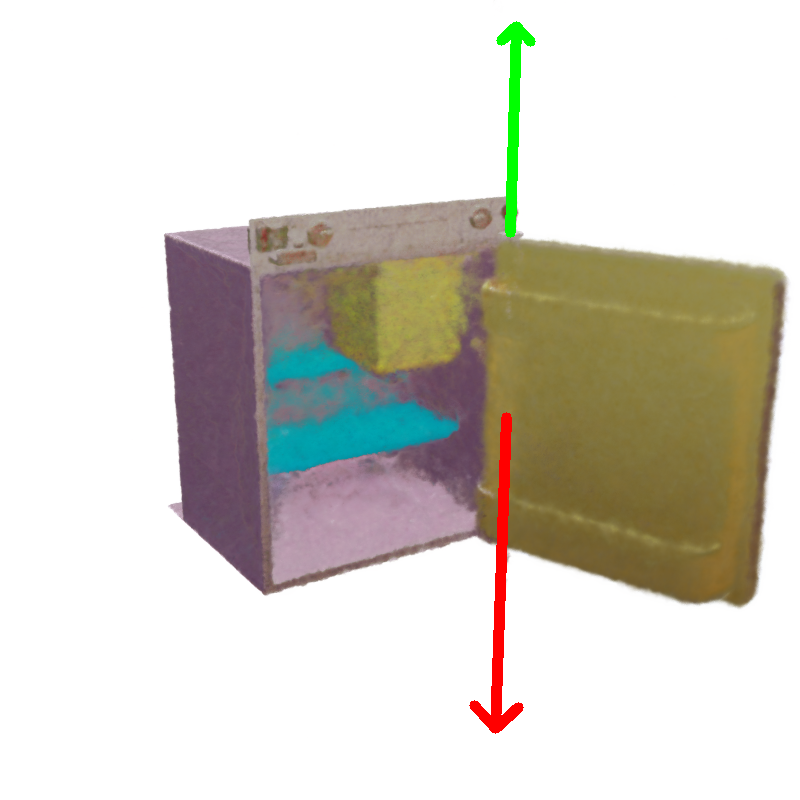}&
        \includegraphics[valign=c, width=0.1\columnwidth]{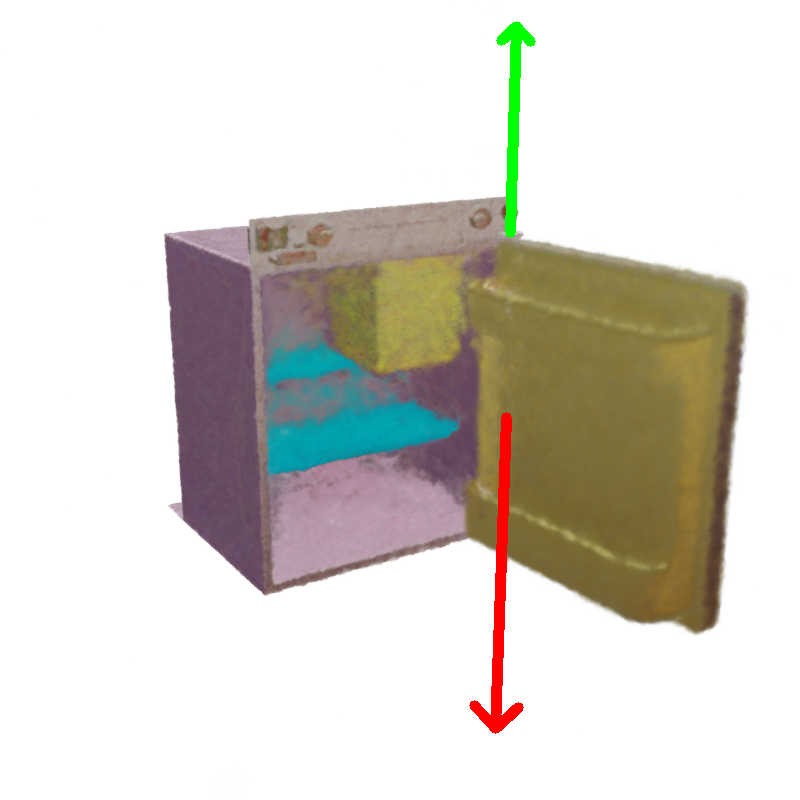}&
        \includegraphics[valign=c, width=0.1\columnwidth]{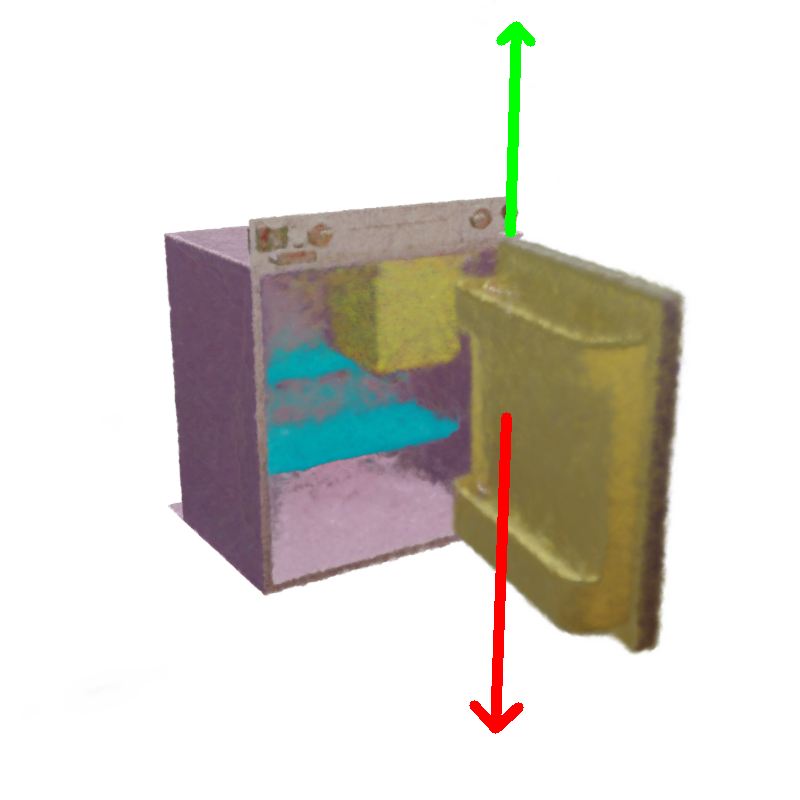}&
        \includegraphics[valign=c, width=0.1\columnwidth]{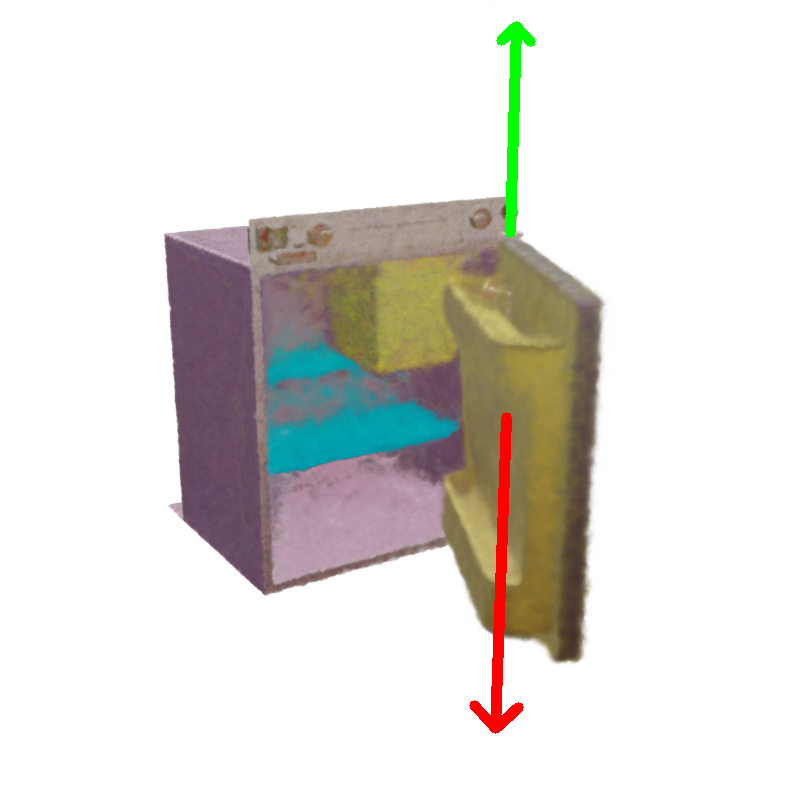}&
        \includegraphics[valign=c, width=0.1\columnwidth]{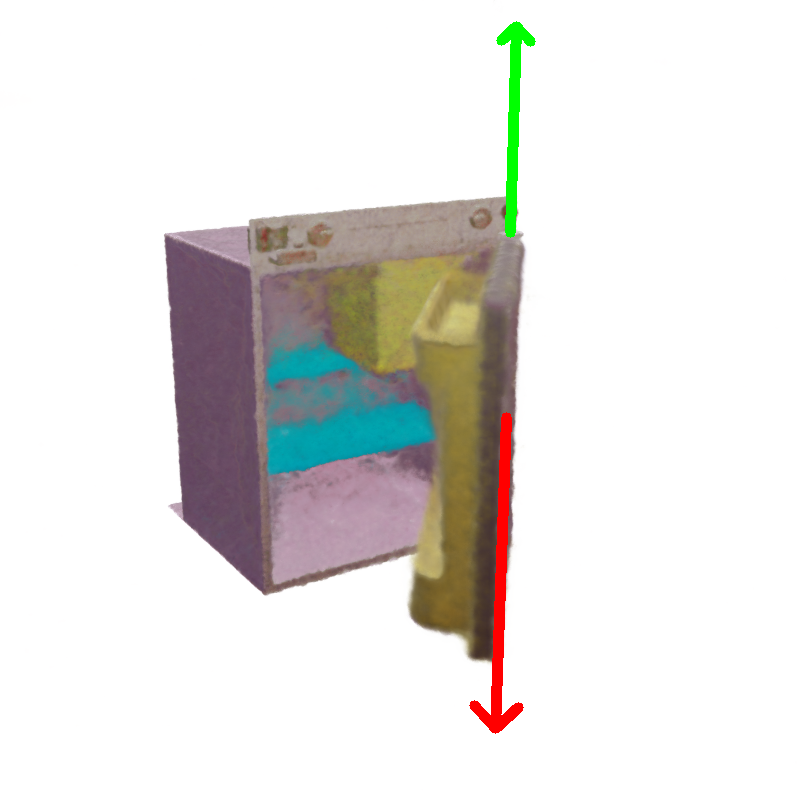}&
        \includegraphics[valign=c, width=0.1\columnwidth]{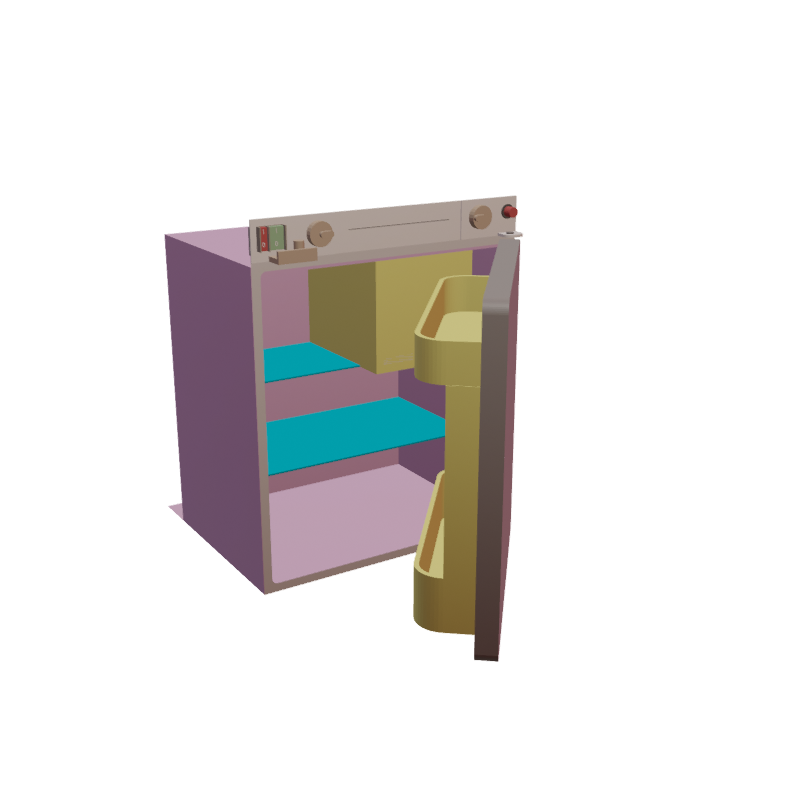}\\
        Ours &
        \includegraphics[valign=c, width=0.1\columnwidth]{figures/input_state/fridge/0033_start.png}&
        \includegraphics[valign=c, width=0.1\columnwidth]{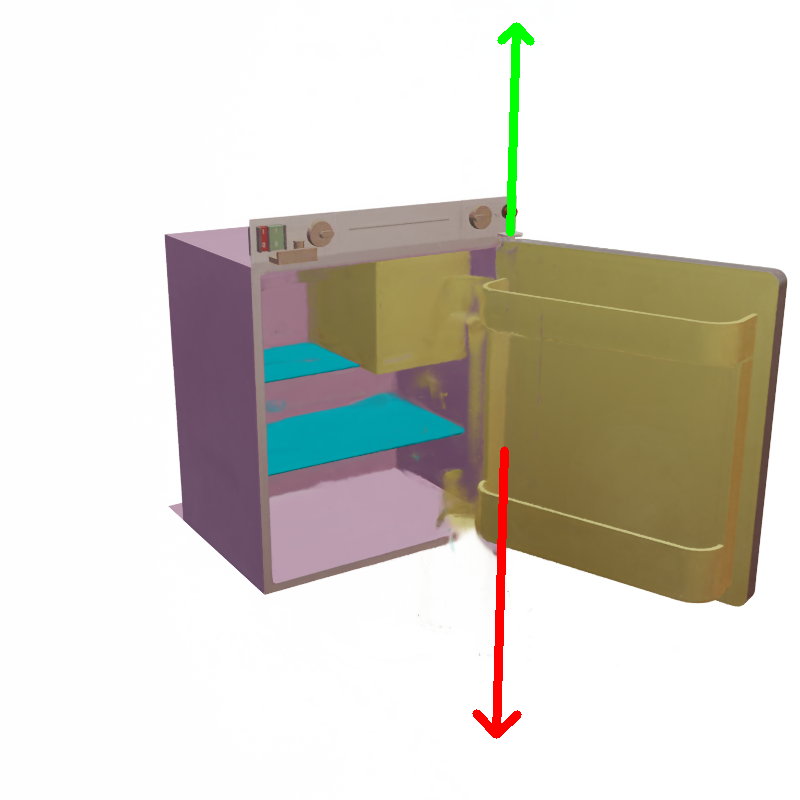}&
        \includegraphics[valign=c, width=0.1\columnwidth]{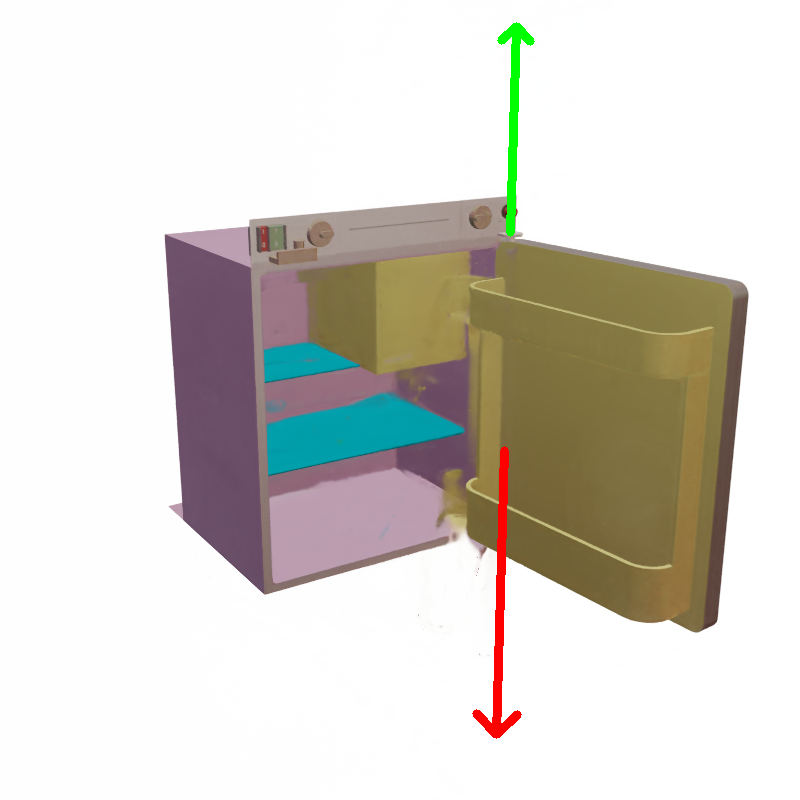}&
        \includegraphics[valign=c, width=0.1\columnwidth]{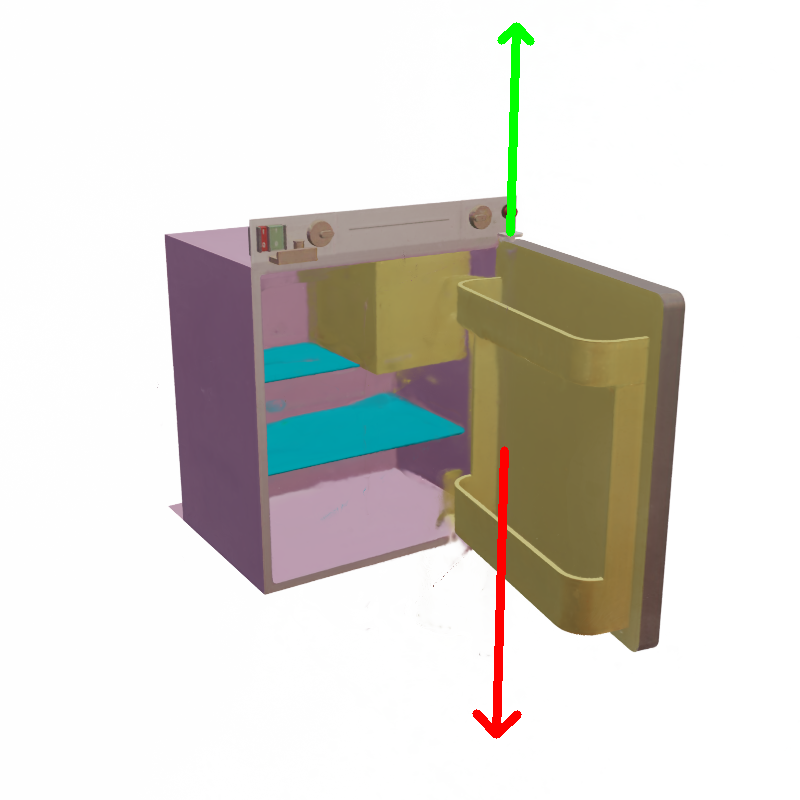}&
        \includegraphics[valign=c, width=0.1\columnwidth]{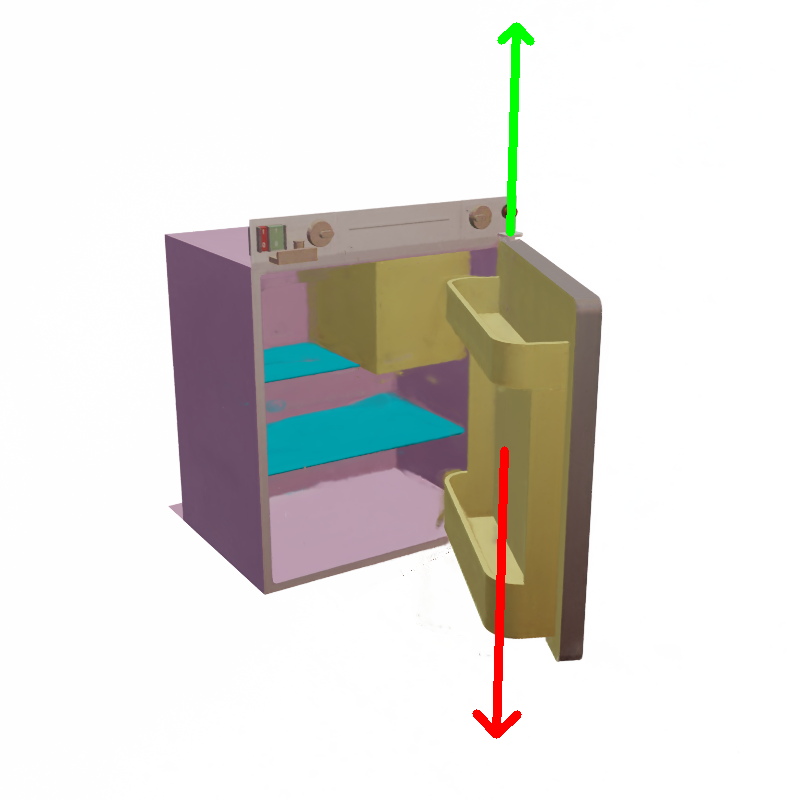}&
        \includegraphics[valign=c, width=0.1\columnwidth]{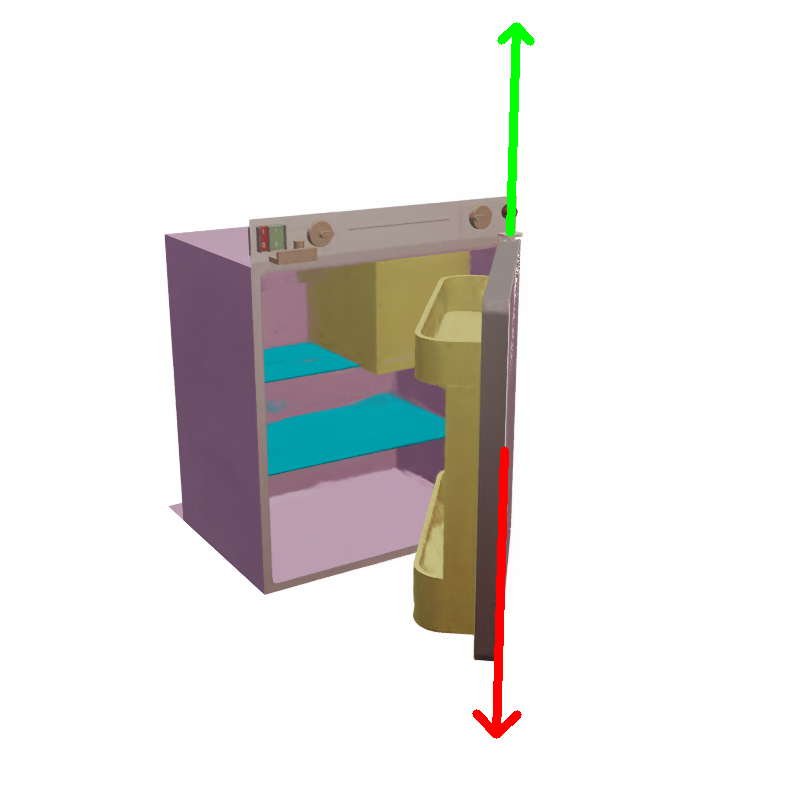}&
        \includegraphics[valign=c, width=0.1\columnwidth]{figures/input_state/fridge/0033_end.png}\\
        \hline
        PARIS &
        \includegraphics[valign=c, width=0.1\columnwidth]{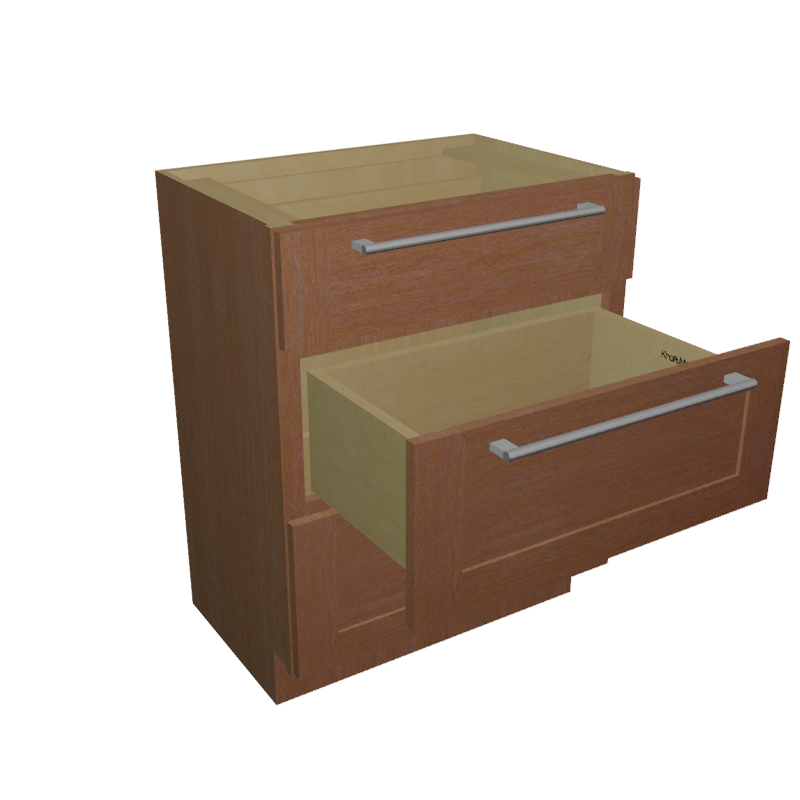}&
        \includegraphics[valign=c, width=0.1\columnwidth]{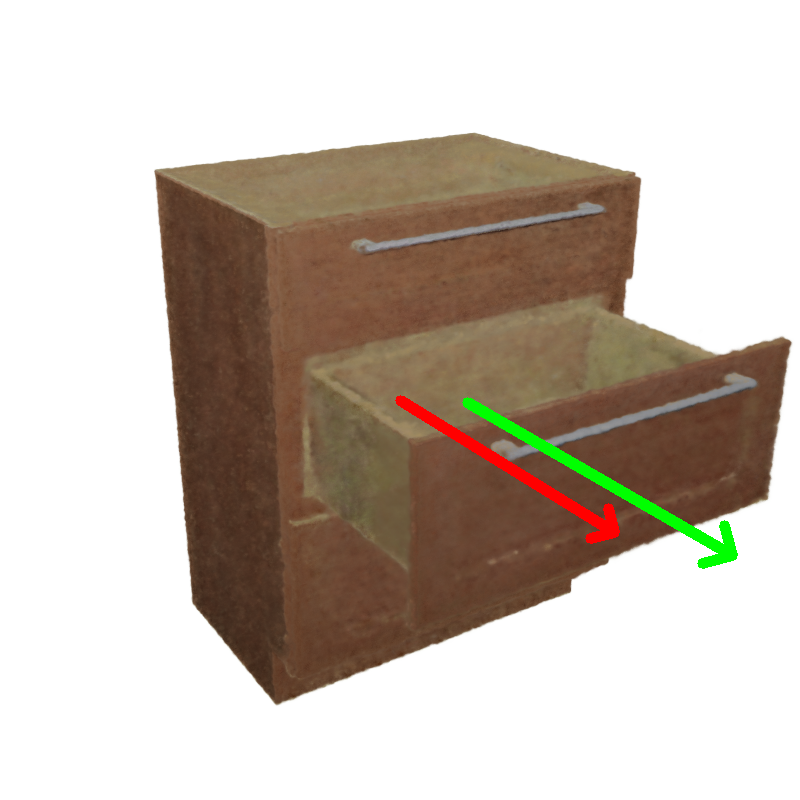}&
        \includegraphics[valign=c, width=0.1\columnwidth]{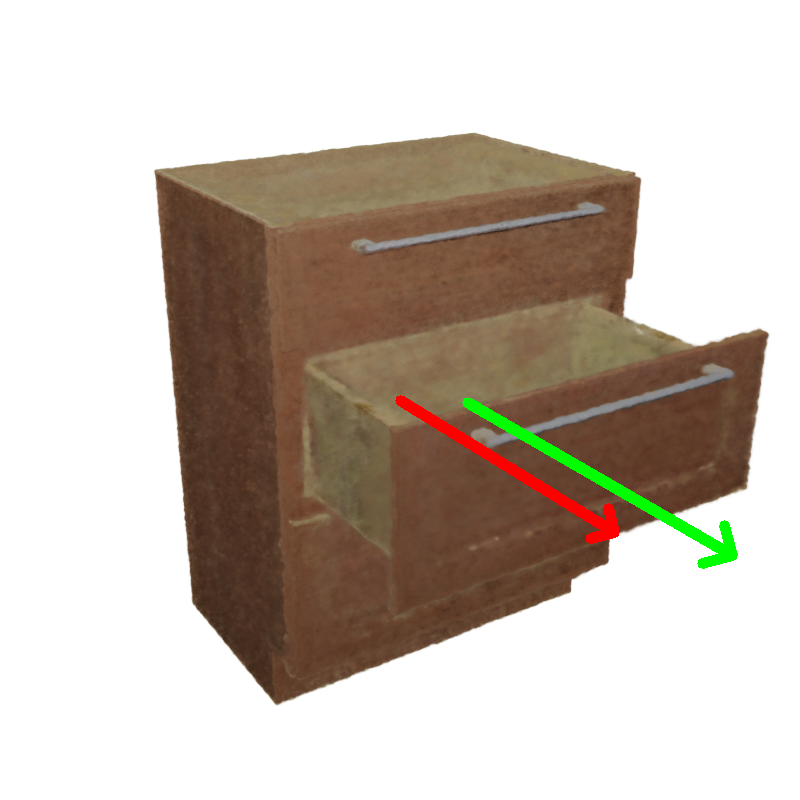}&
        \includegraphics[valign=c, width=0.1\columnwidth]{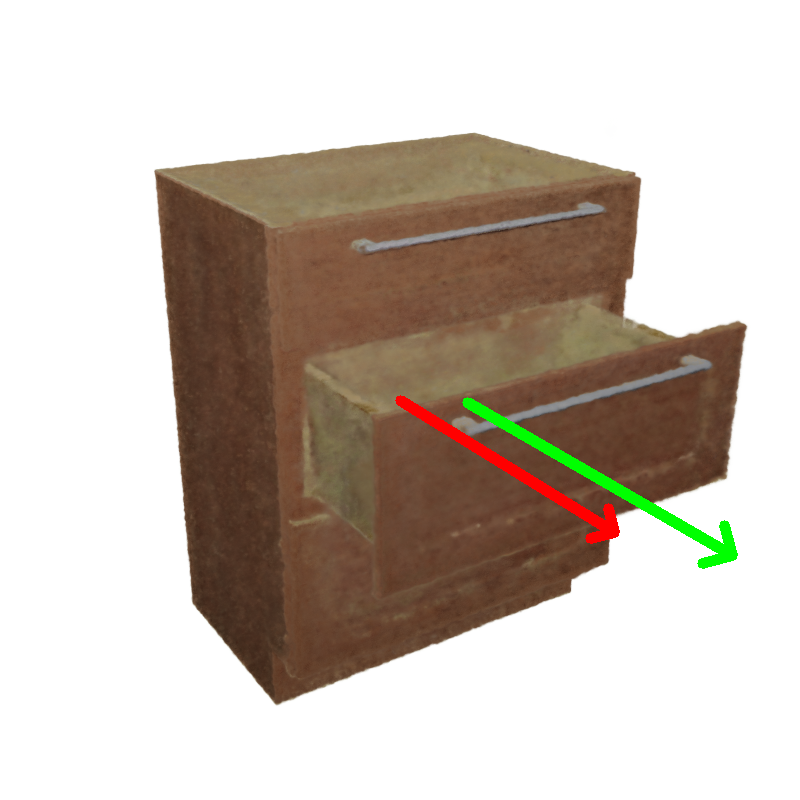}&
        \includegraphics[valign=c, width=0.1\columnwidth]{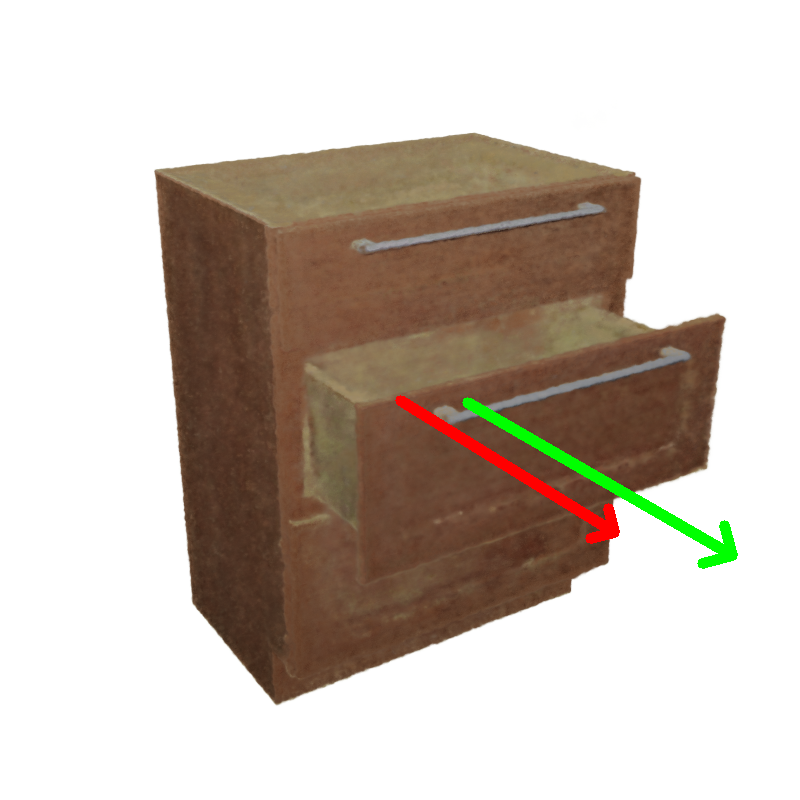}&
        \includegraphics[valign=c, width=0.1\columnwidth]{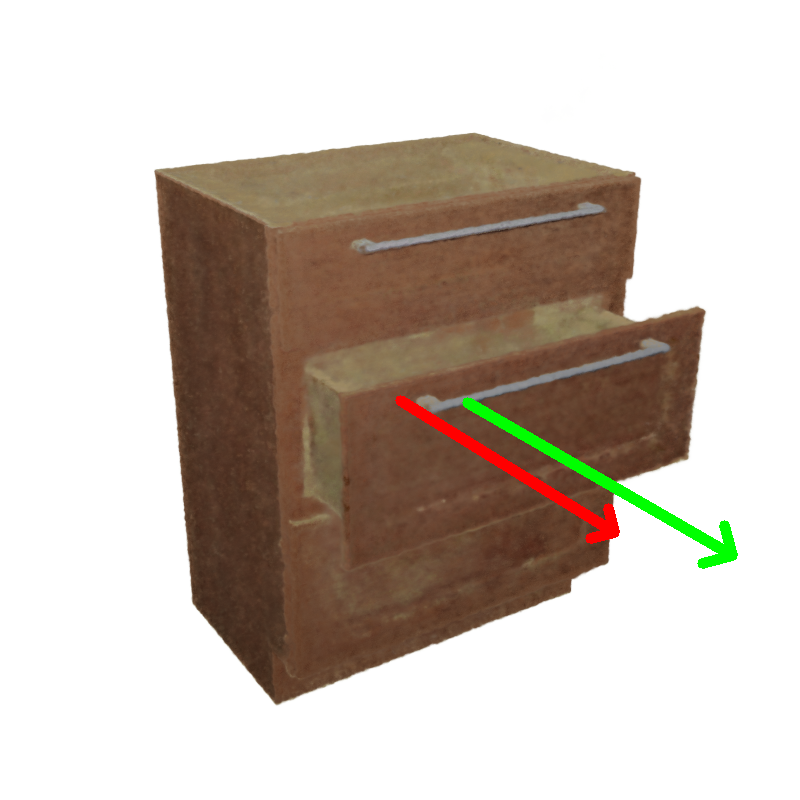}&
        \includegraphics[valign=c, width=0.1\columnwidth]{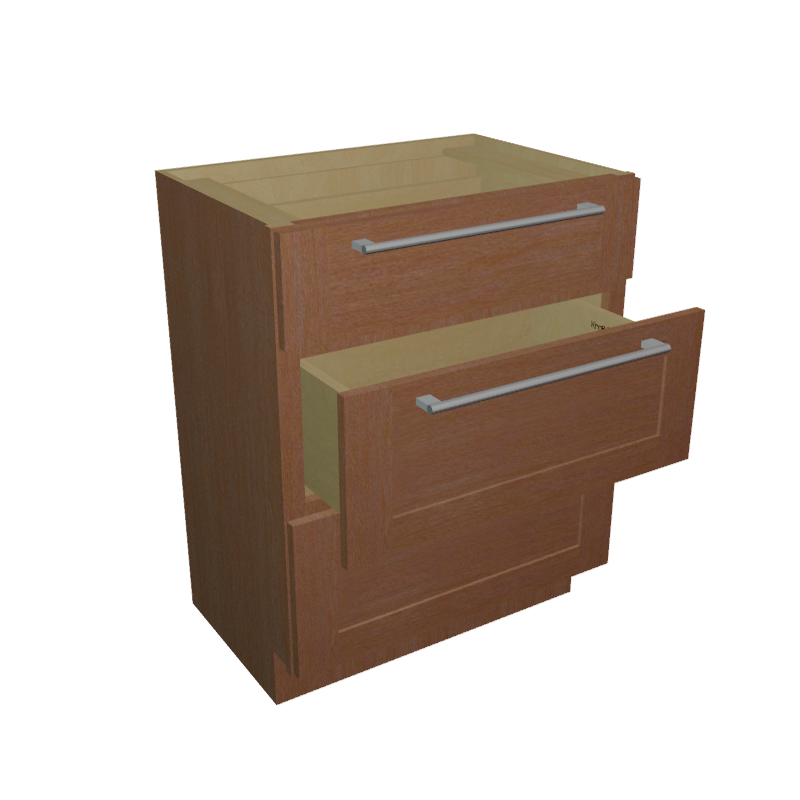}&\\
        Ours &
        \includegraphics[valign=c, width=0.1\columnwidth]{figures/input_state/storage/0033_end.png}&
        \includegraphics[valign=c, width=0.1\columnwidth]{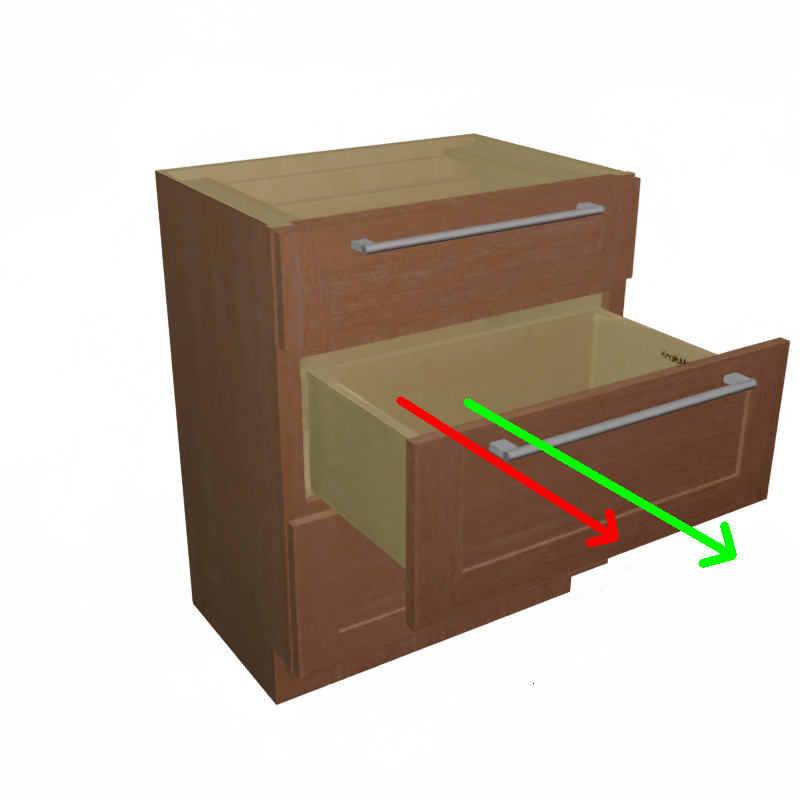}&
        \includegraphics[valign=c, width=0.1\columnwidth]{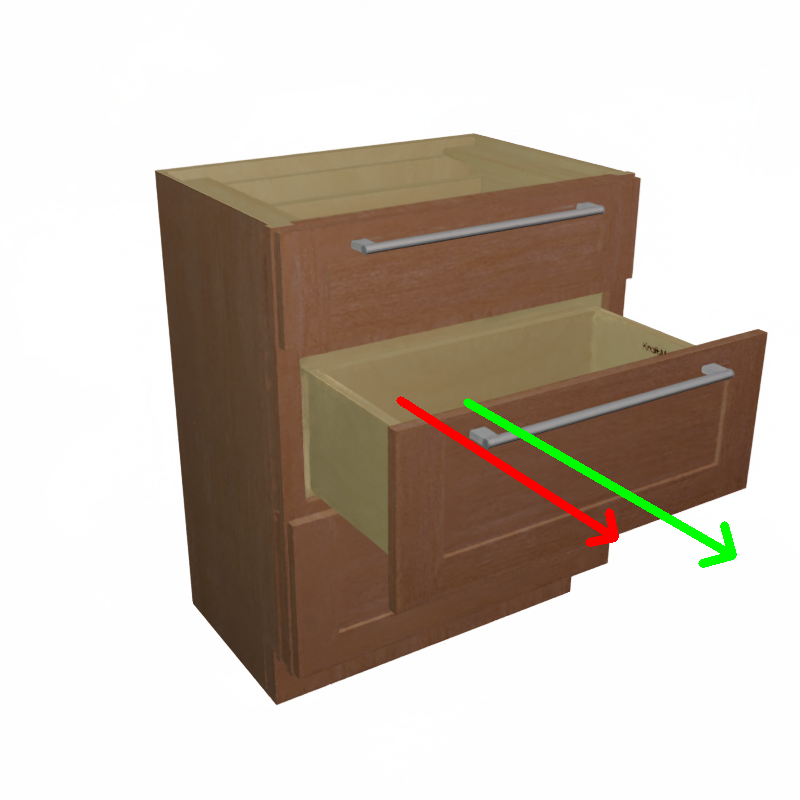}&
        \includegraphics[valign=c, width=0.1\columnwidth]{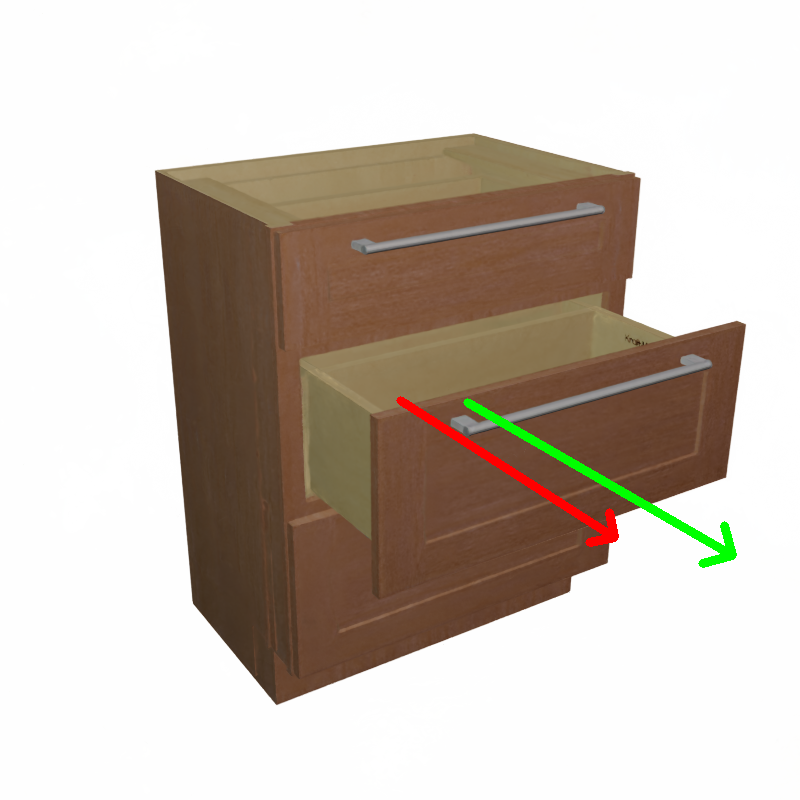}&
        \includegraphics[valign=c, width=0.1\columnwidth]{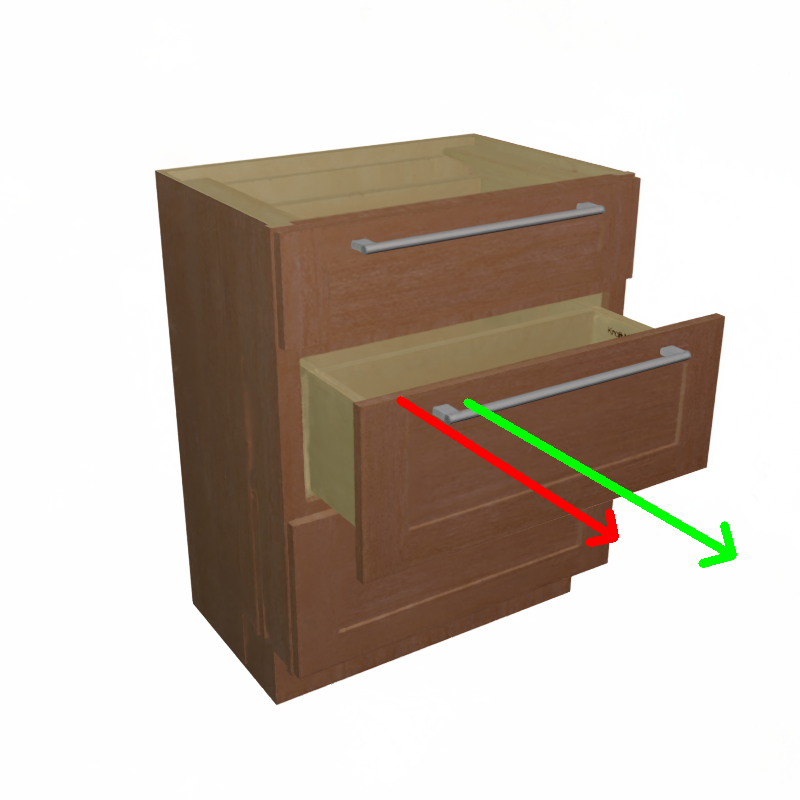}&
        \includegraphics[valign=c, width=0.1\columnwidth]{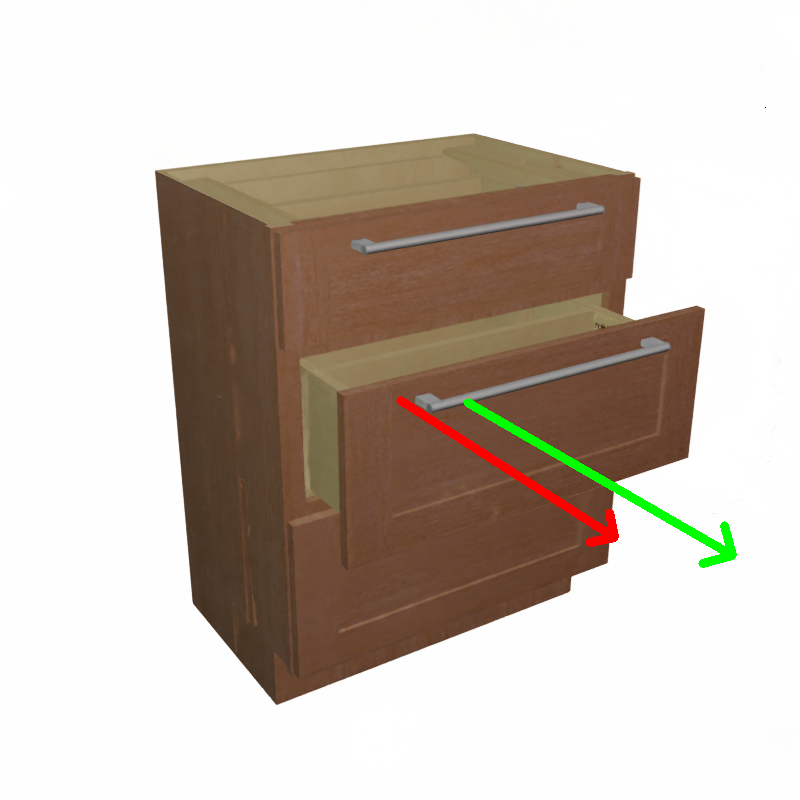}&
        \includegraphics[valign=c, width=0.1\columnwidth]{figures/input_state/storage/0033.png}&\\
        \hline
        PARIS &
        \includegraphics[valign=c, width=0.1\columnwidth]{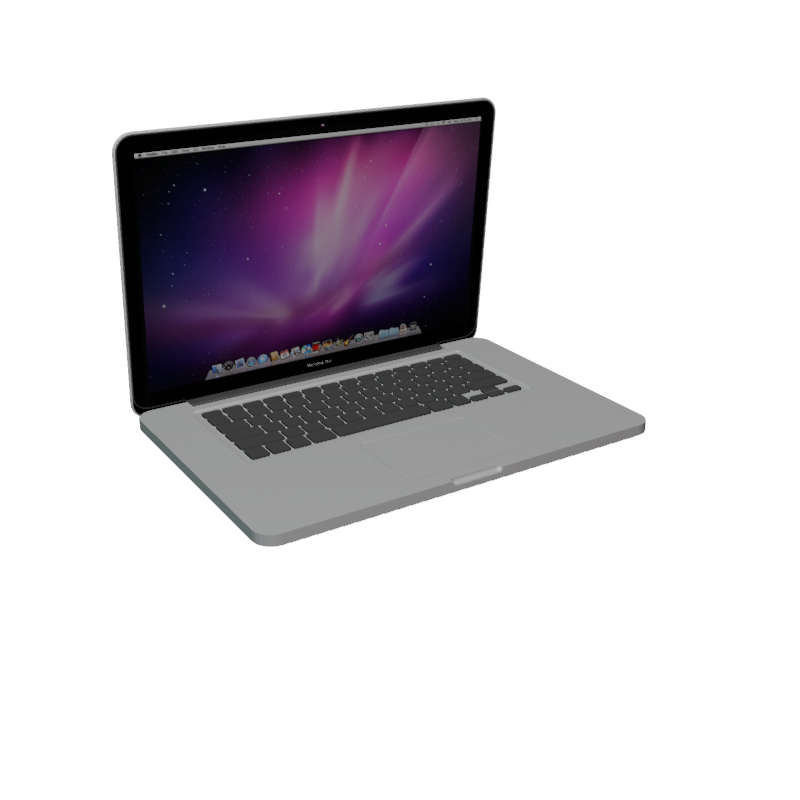}&
        \includegraphics[valign=c, width=0.1\columnwidth]{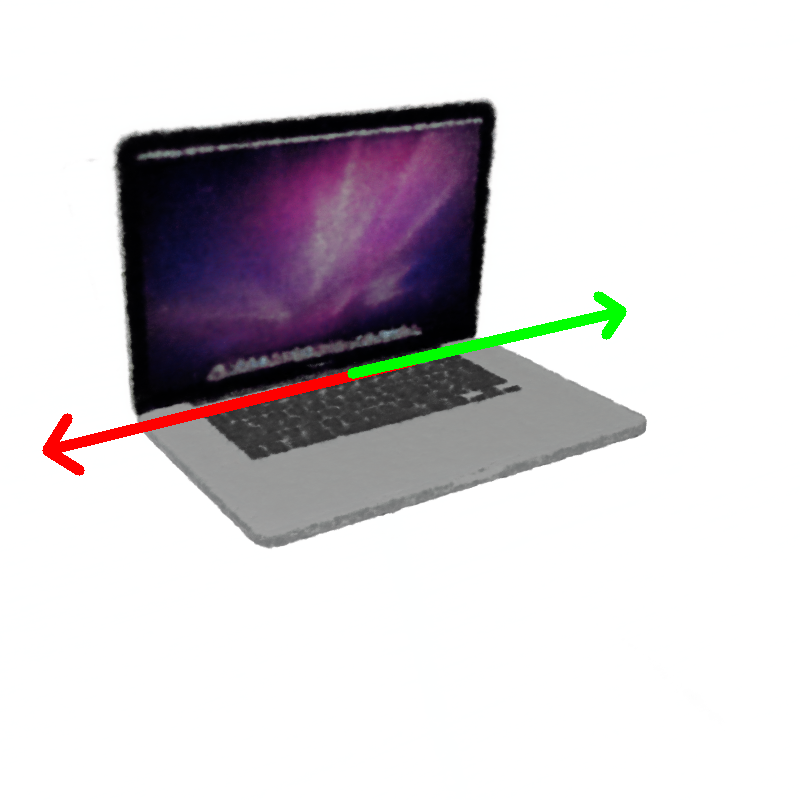}&
        \includegraphics[valign=c, width=0.1\columnwidth]{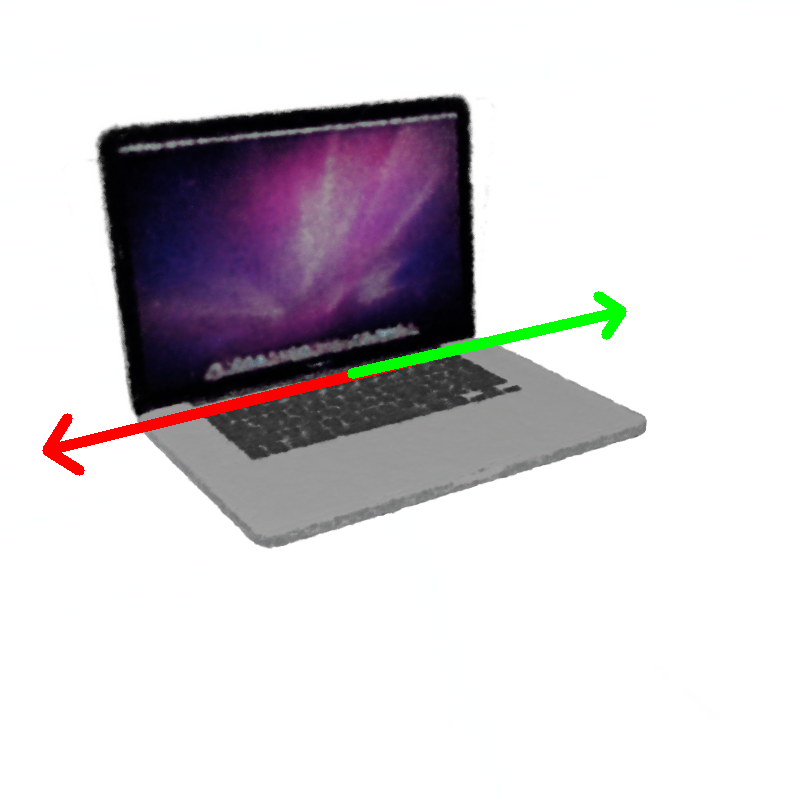}&
        \includegraphics[valign=c, width=0.1\columnwidth]{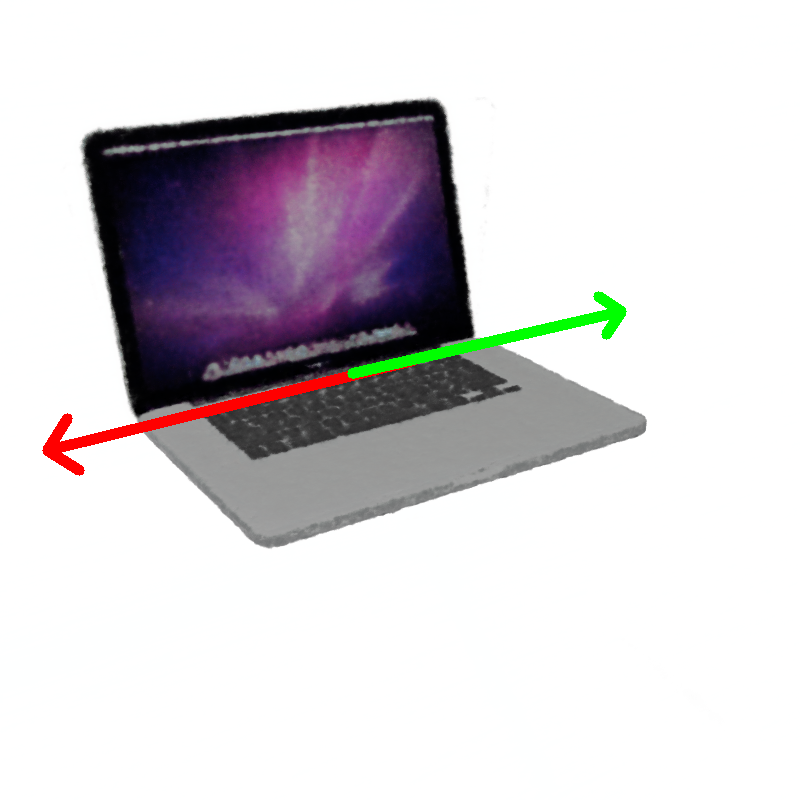}&
        \includegraphics[valign=c, width=0.1\columnwidth]{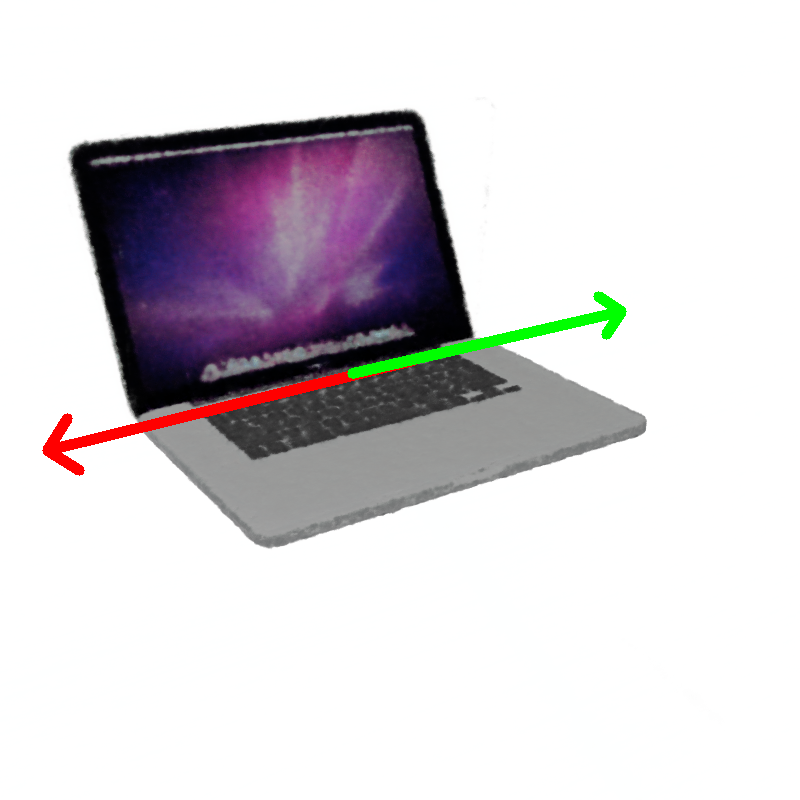}&
        \includegraphics[valign=c, width=0.1\columnwidth]{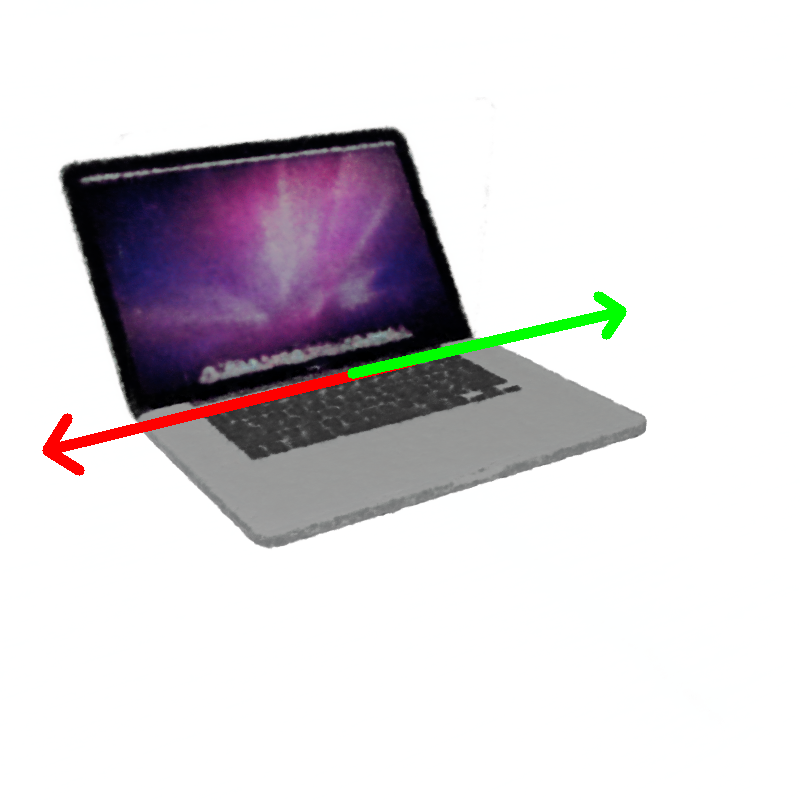}&
        \includegraphics[valign=c, width=0.1\columnwidth]{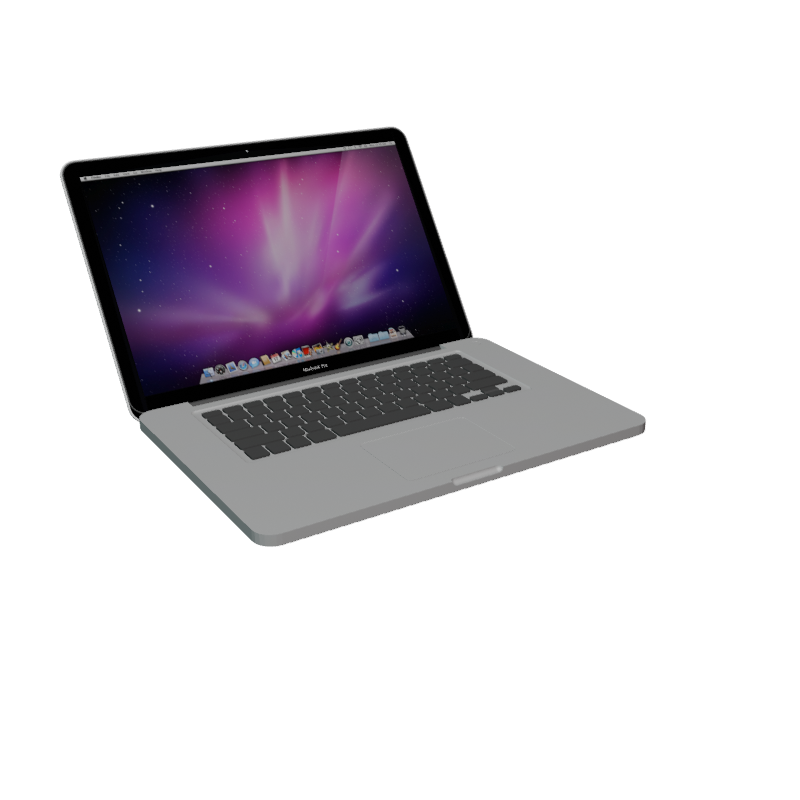}&\\
        Ours &
        \includegraphics[valign=c, width=0.1\columnwidth]{figures/input_state/laptop/start_0033.png}&
        \includegraphics[valign=c, width=0.1\columnwidth]{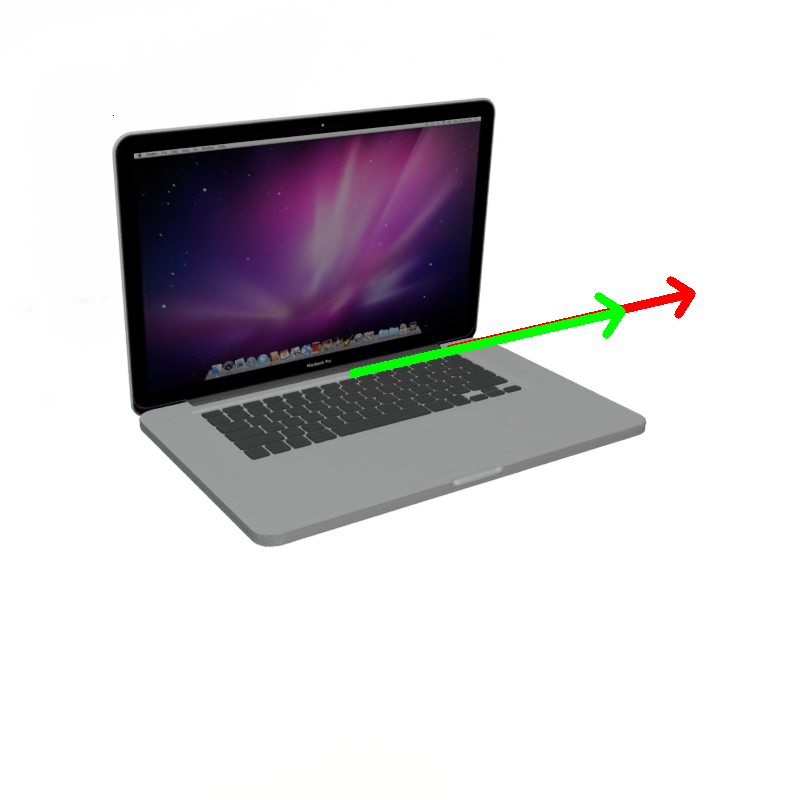}&
        \includegraphics[valign=c, width=0.1\columnwidth]{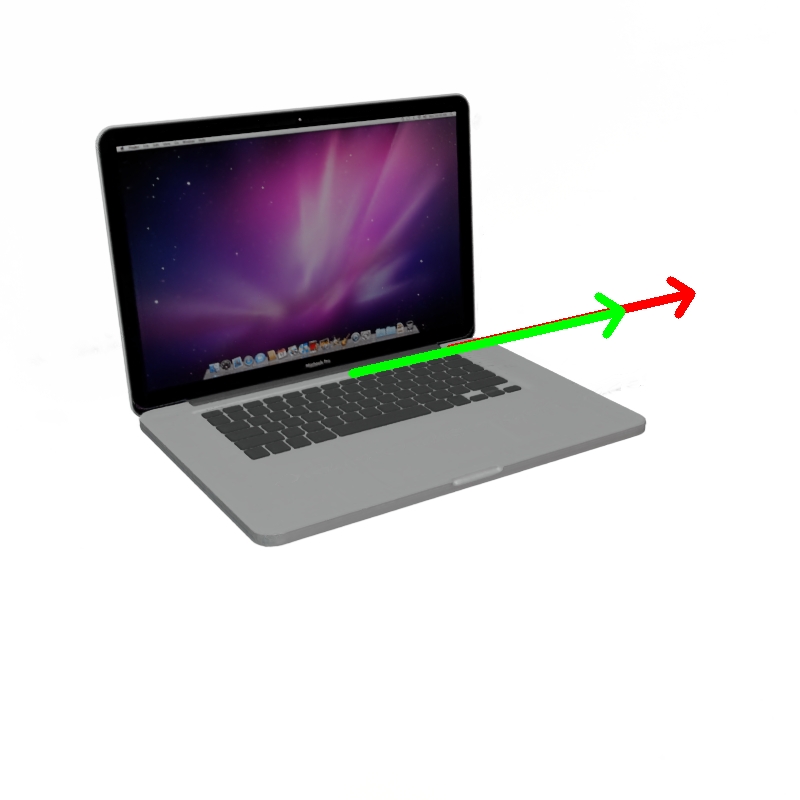}&
        \includegraphics[valign=c, width=0.1\columnwidth]{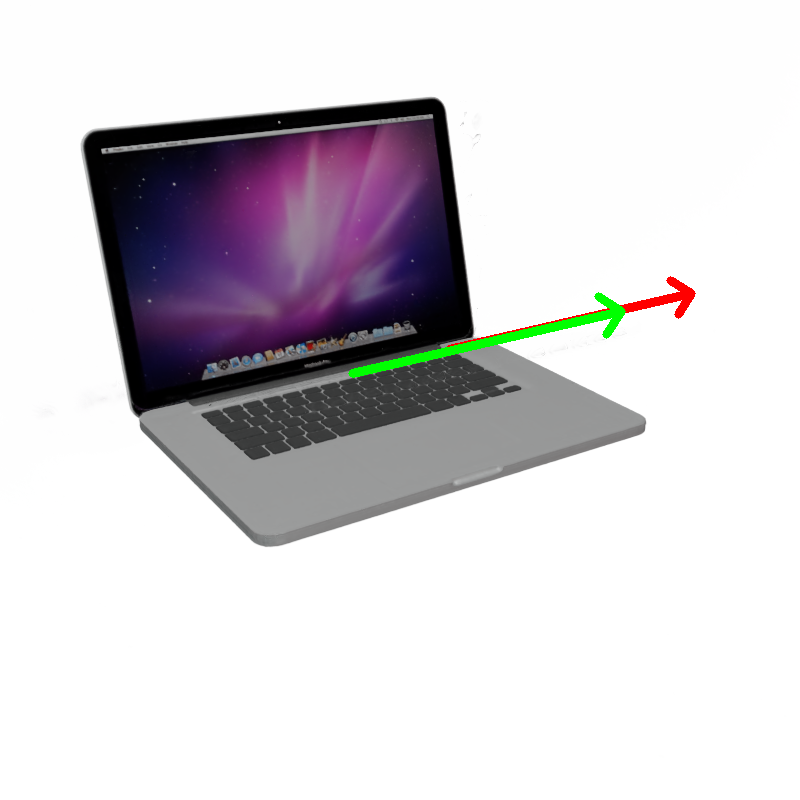}&
        \includegraphics[valign=c, width=0.1\columnwidth]{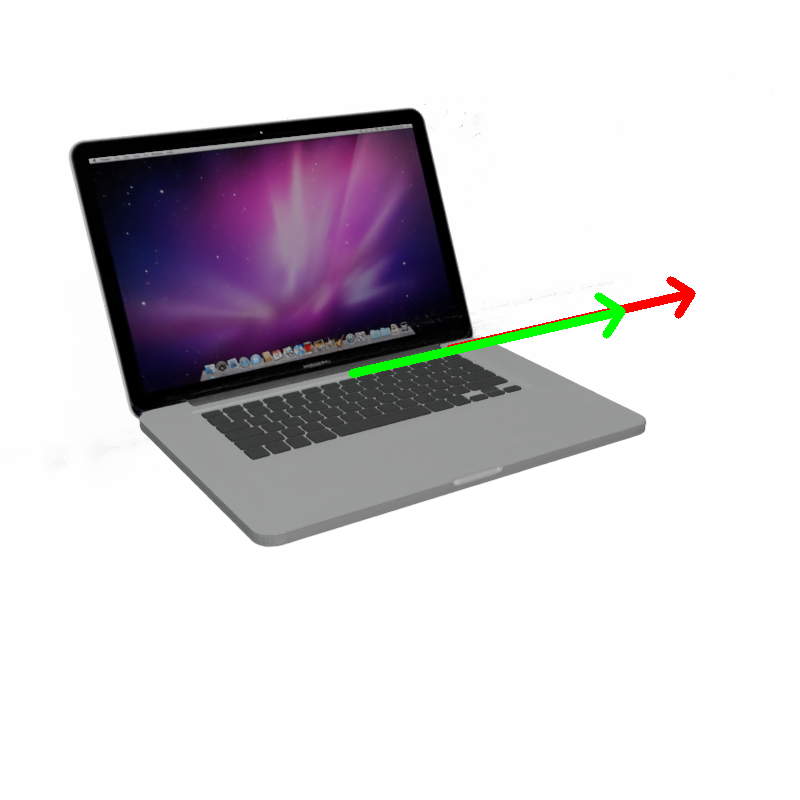}&
        \includegraphics[valign=c, width=0.1\columnwidth]{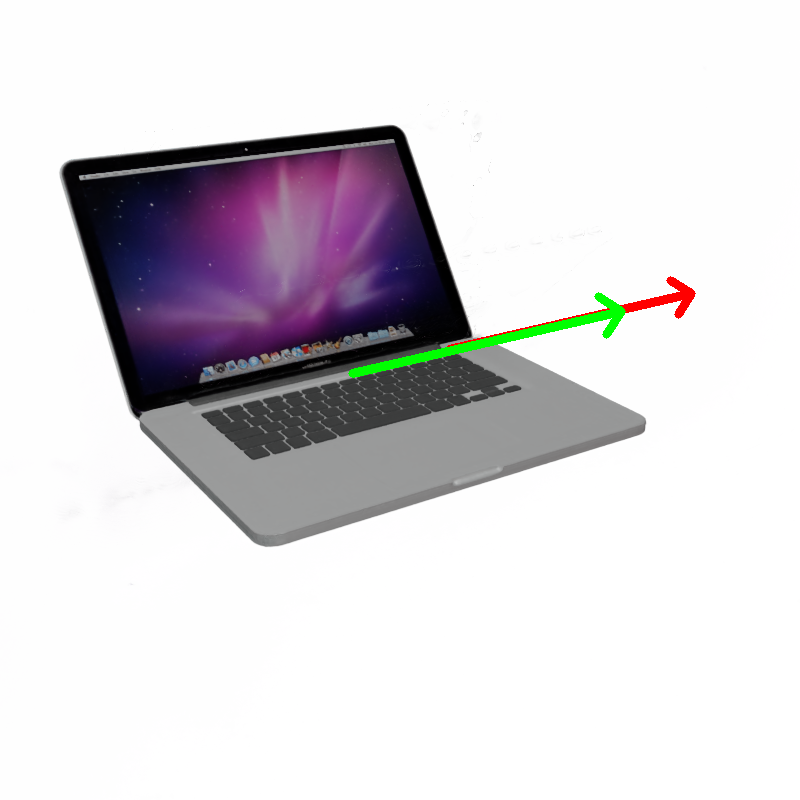}&
        \includegraphics[valign=c, width=0.1\columnwidth]{figures/input_state/laptop/end_0033.png}&\\
        \hline
        PARIS &
        \includegraphics[valign=c, width=0.1\columnwidth]{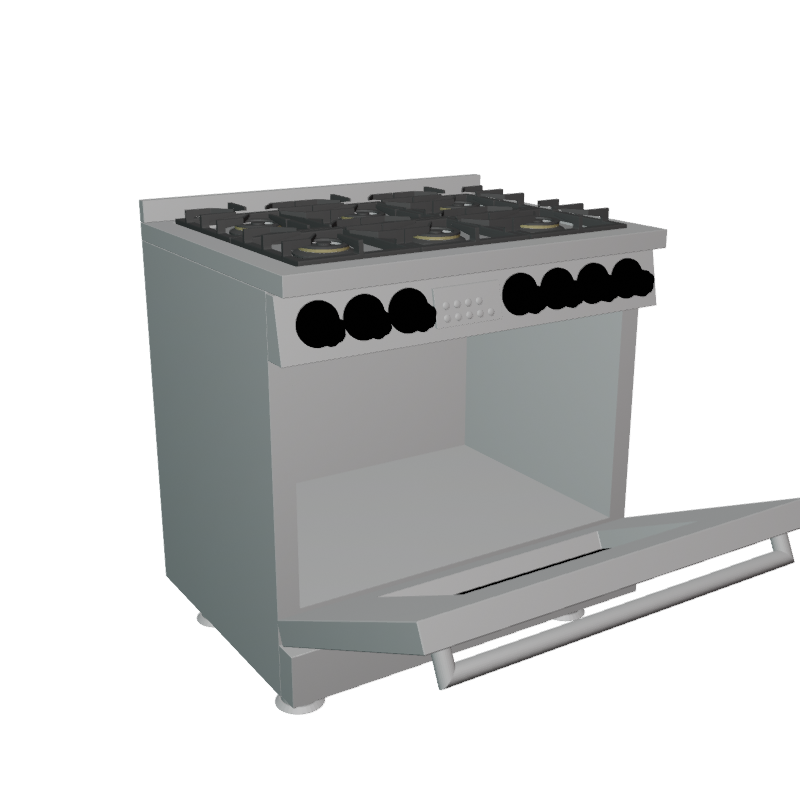}&
        \includegraphics[valign=c, width=0.1\columnwidth]{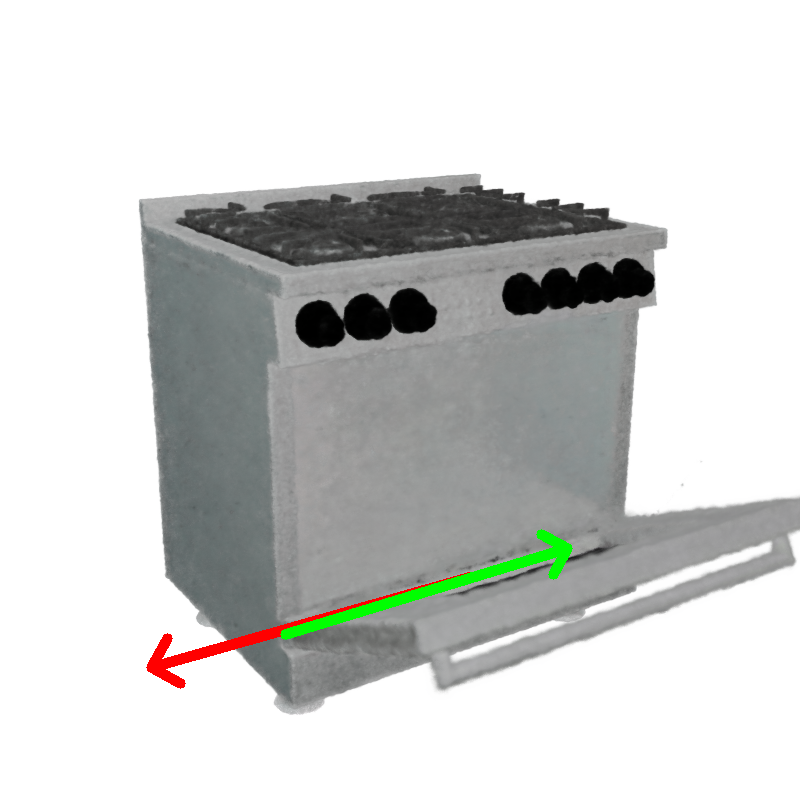}&
        \includegraphics[valign=c, width=0.1\columnwidth]{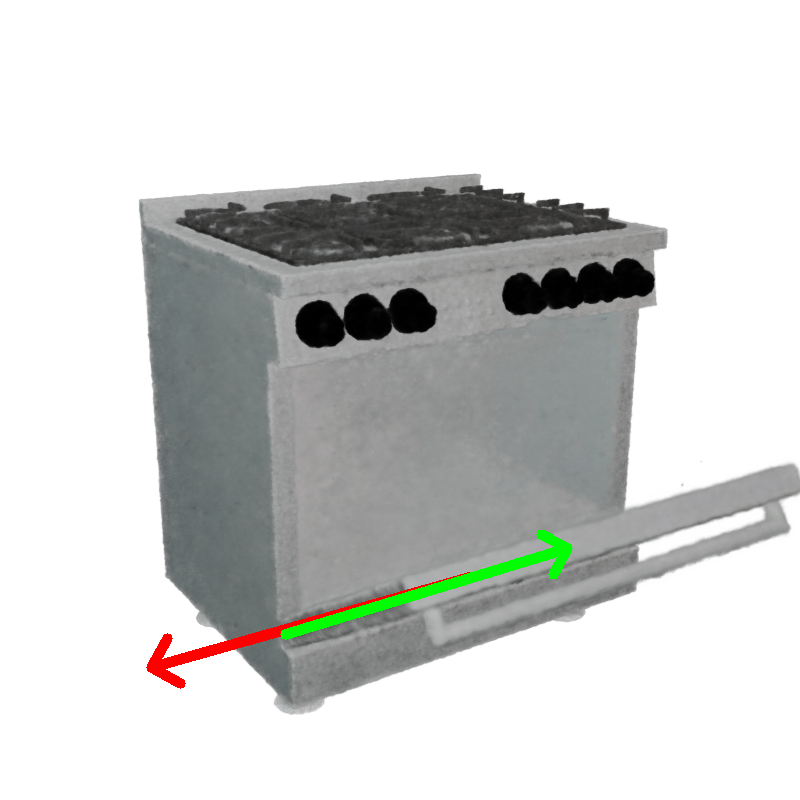}&
        \includegraphics[valign=c, width=0.1\columnwidth]{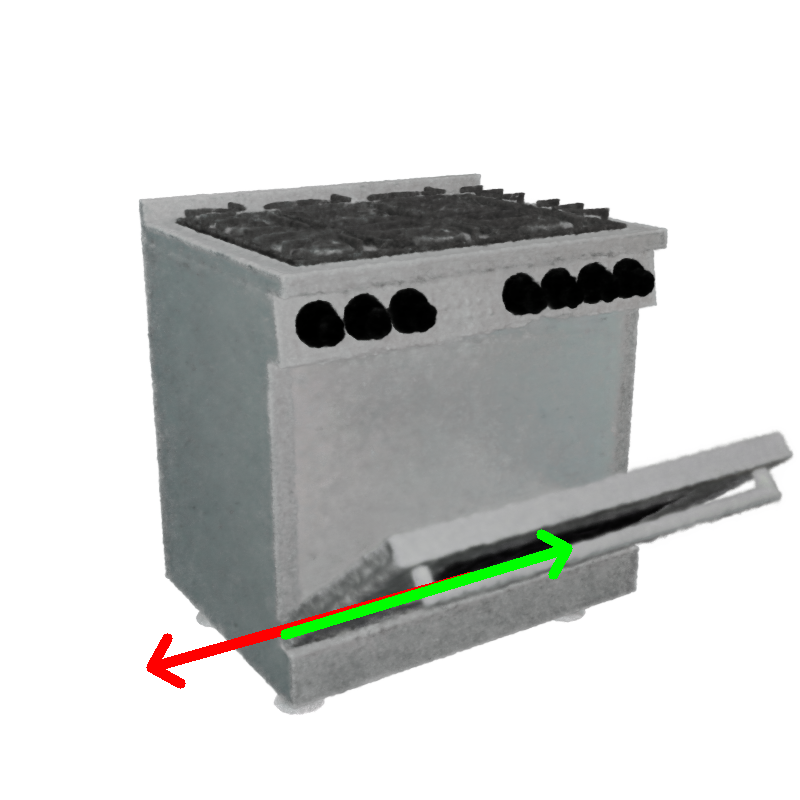}&
        \includegraphics[valign=c, width=0.1\columnwidth]{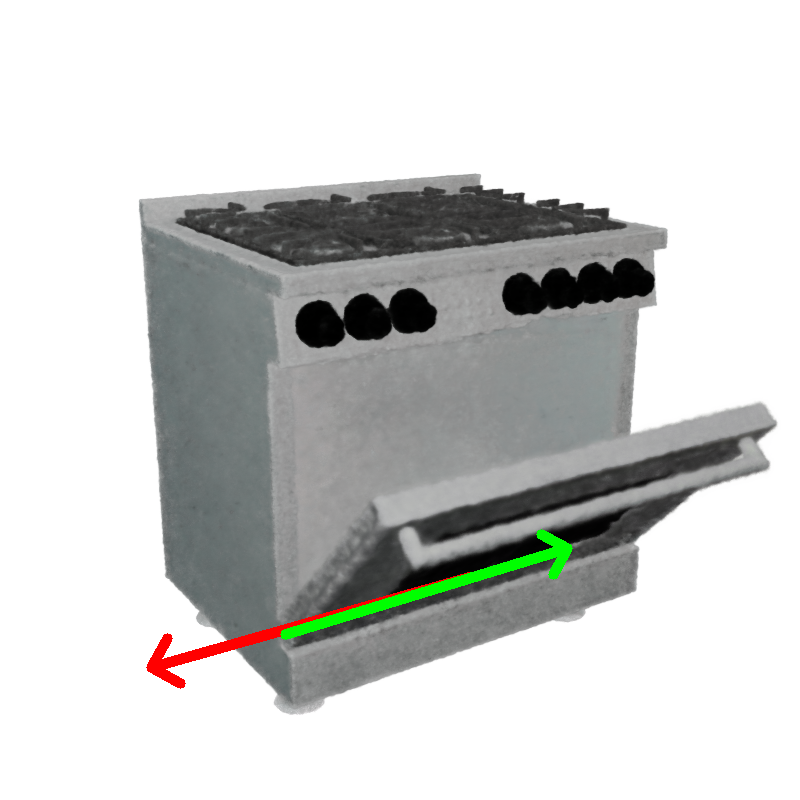}&
        \includegraphics[valign=c, width=0.1\columnwidth]{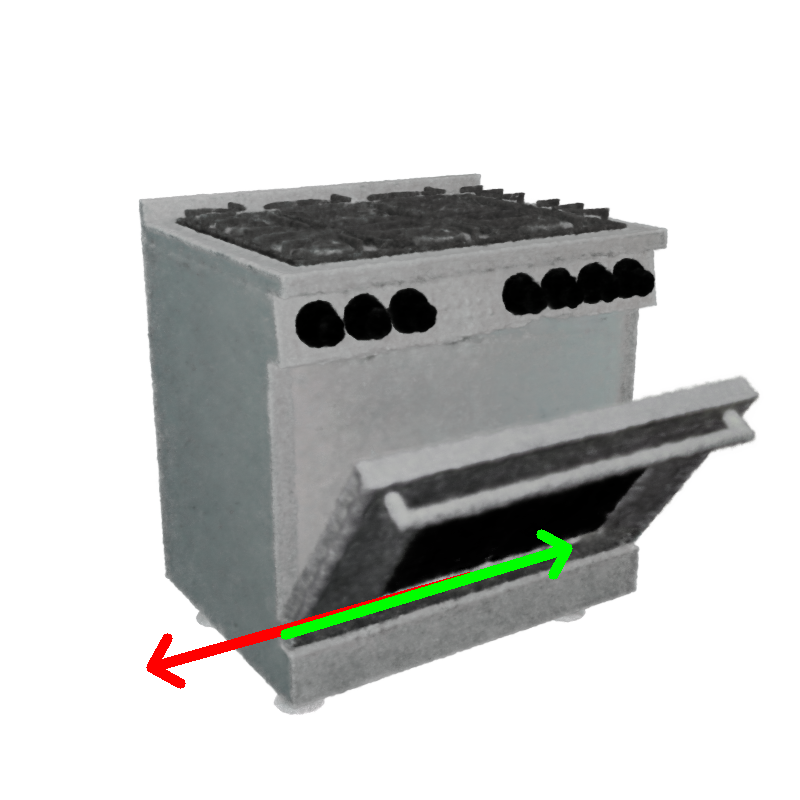}&
        \includegraphics[valign=c, width=0.1\columnwidth]{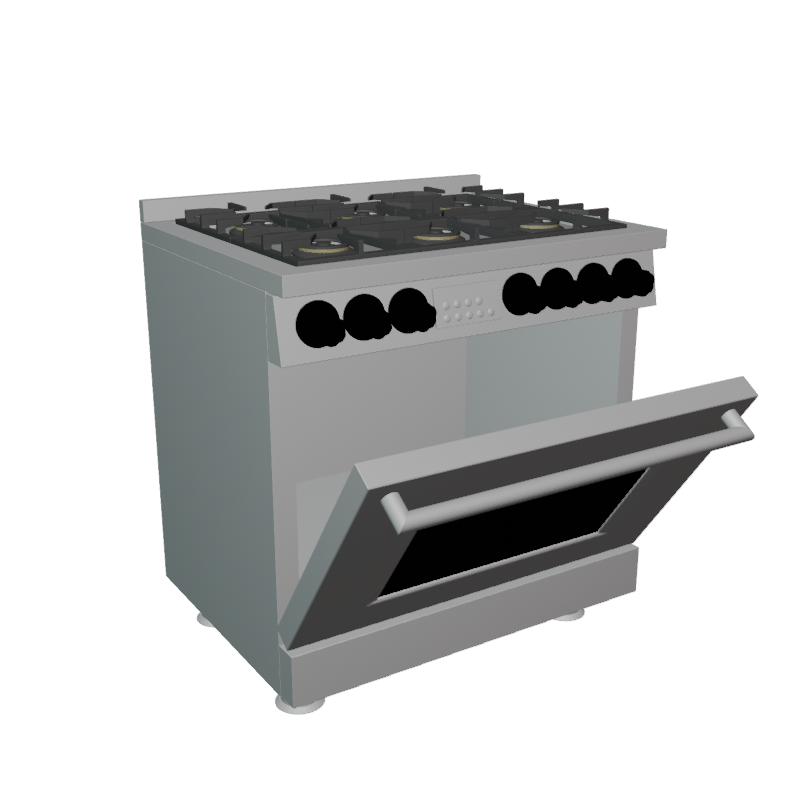}&\\
        Ours &
        \includegraphics[valign=c, width=0.1\columnwidth]{figures/input_state/oven/start_0033.png}&
        \includegraphics[valign=c, width=0.1\columnwidth]{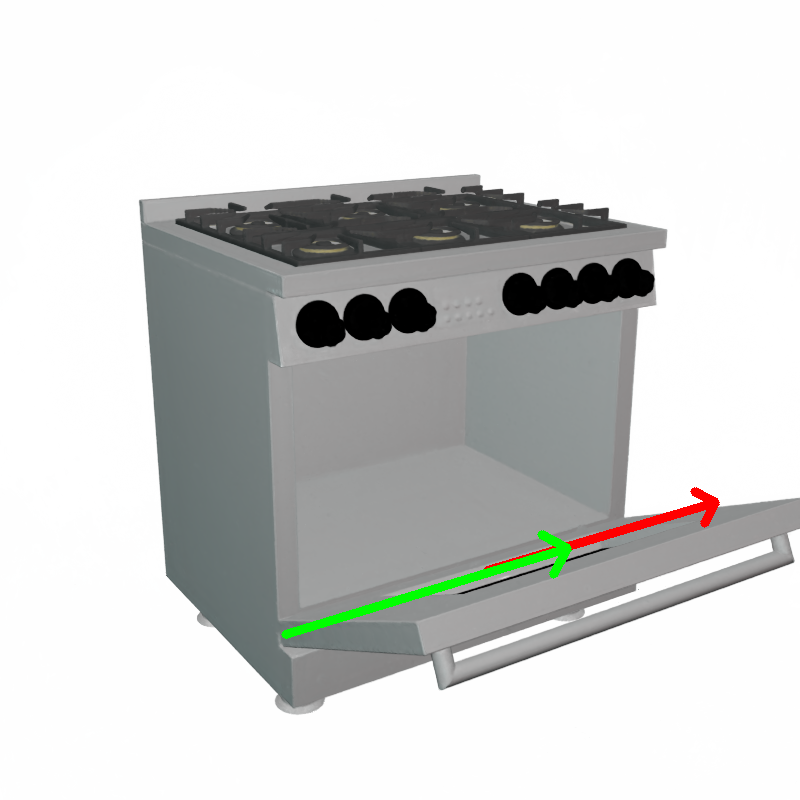}&
        \includegraphics[valign=c, width=0.1\columnwidth]{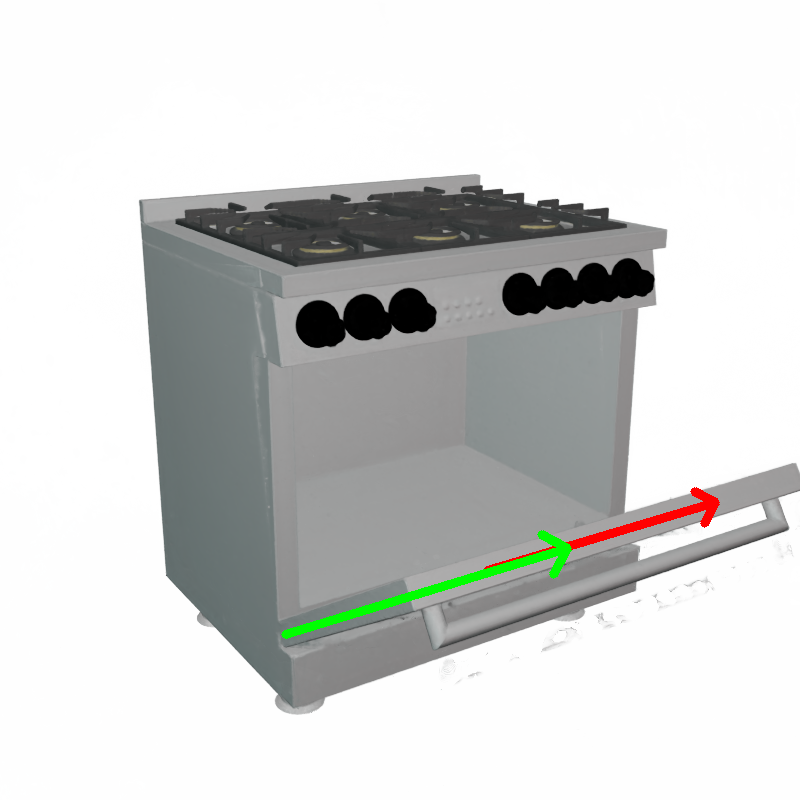}&
        \includegraphics[valign=c, width=0.1\columnwidth]{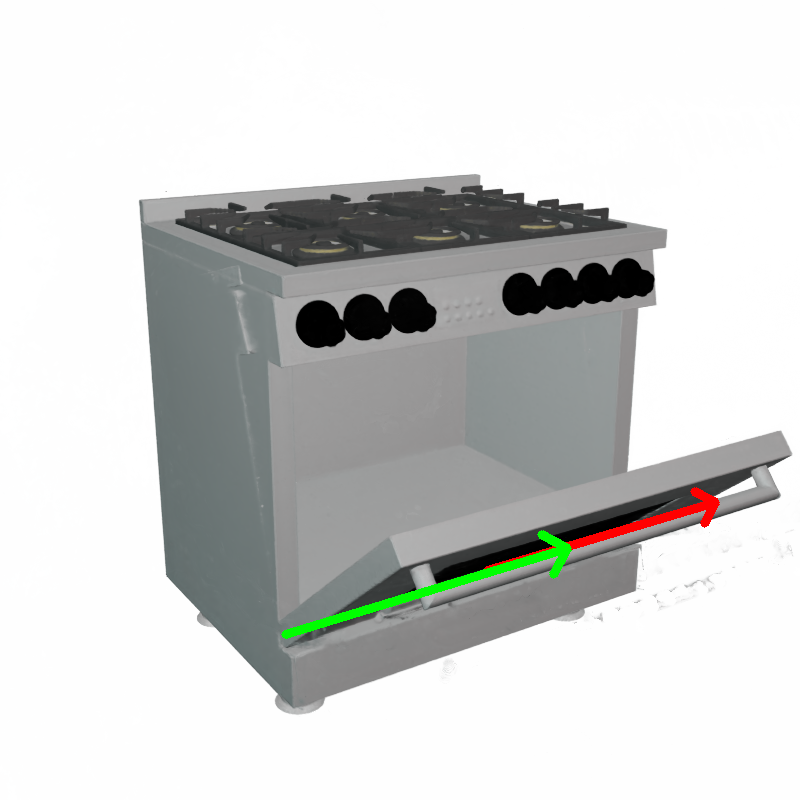}&
        \includegraphics[valign=c, width=0.1\columnwidth]{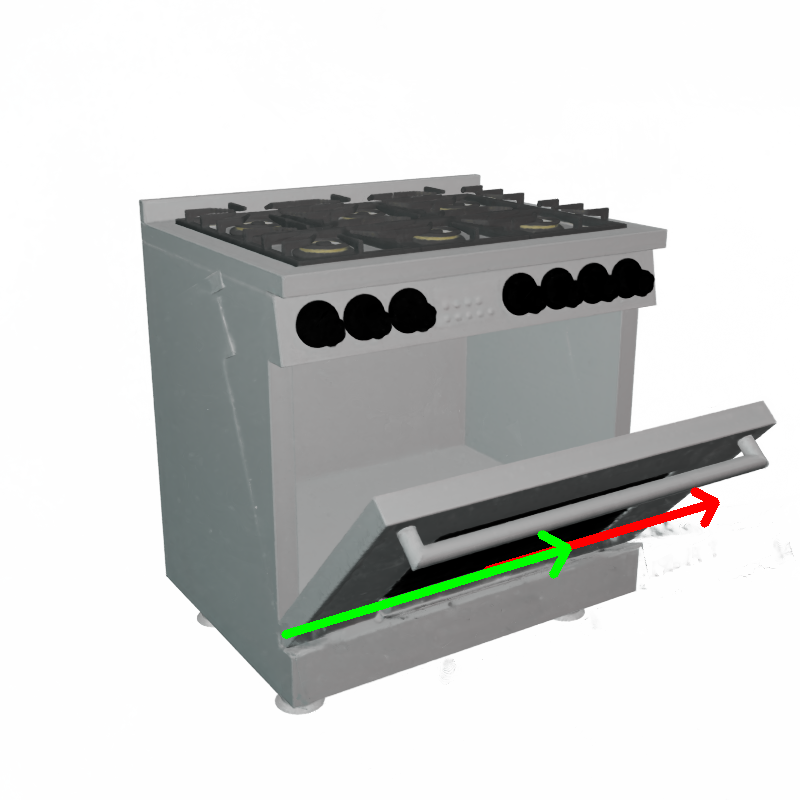}&
        \includegraphics[valign=c, width=0.1\columnwidth]{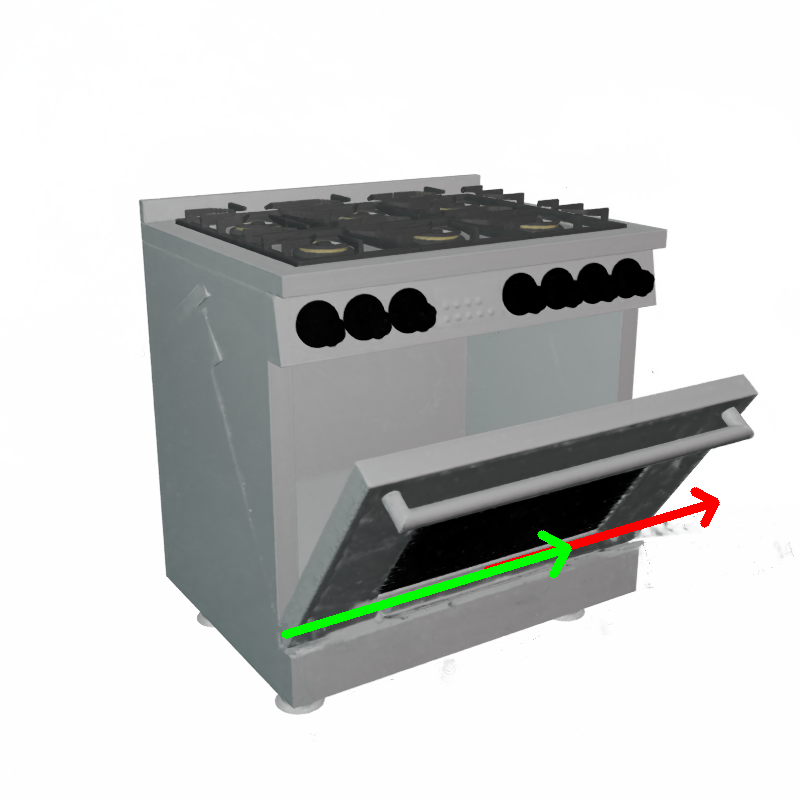}&
        \includegraphics[valign=c, width=0.1\columnwidth]{figures/input_state/oven/end_0033.png}&\\
        \hline
    \end{tblr}
    }
    \caption{\textbf{Articulation interpolation for single moving part objects.}}
    \label{fig:appendix_art_interp_single}
\end{figure}

\begin{figure}
    \centering
    \resizebox{\columnwidth}{!}{%
    \SetTblrInner{rowsep=0pt,colsep=1pt}
    \scriptsize
    \begin{tblr}{c|ccccc|c}
    { GT in $\mathcal{P}$} & \SetCell[c=5]{c}{ Novel articulation synthesis}& &&&& { GT in $\mathcal{P}'$}
        \\
        
        \includegraphics[valign=c, width=0.1\columnwidth]{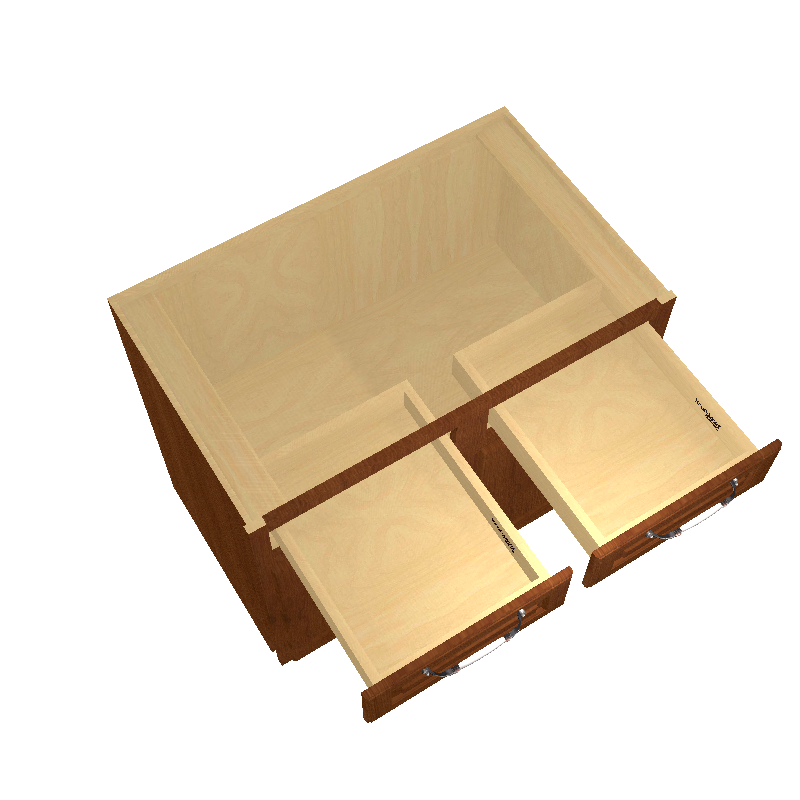}&
        \includegraphics[valign=c, width=0.1\columnwidth]{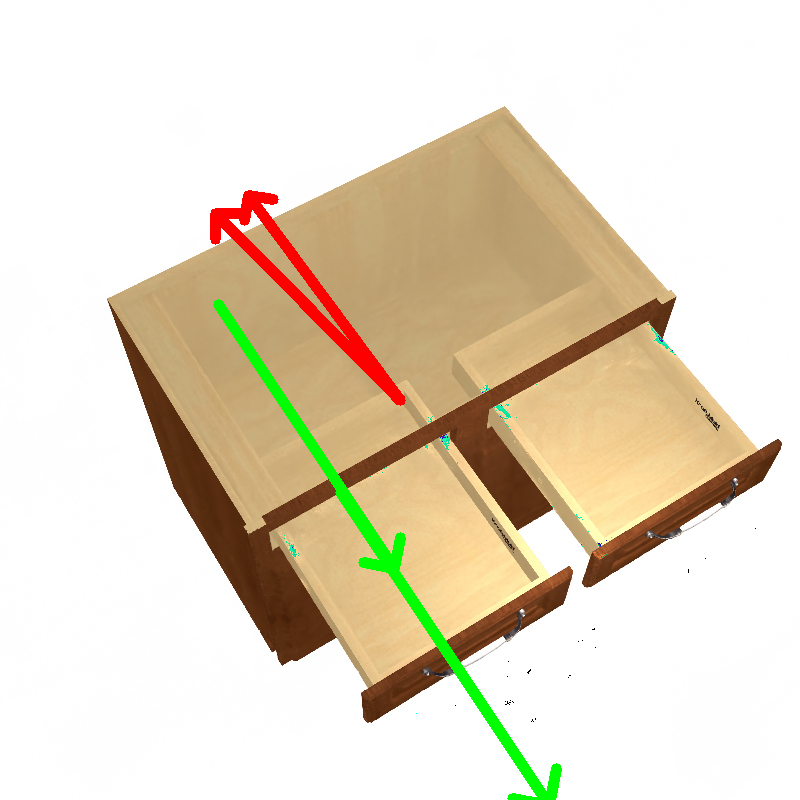}&
        \includegraphics[valign=c, width=0.1\columnwidth]{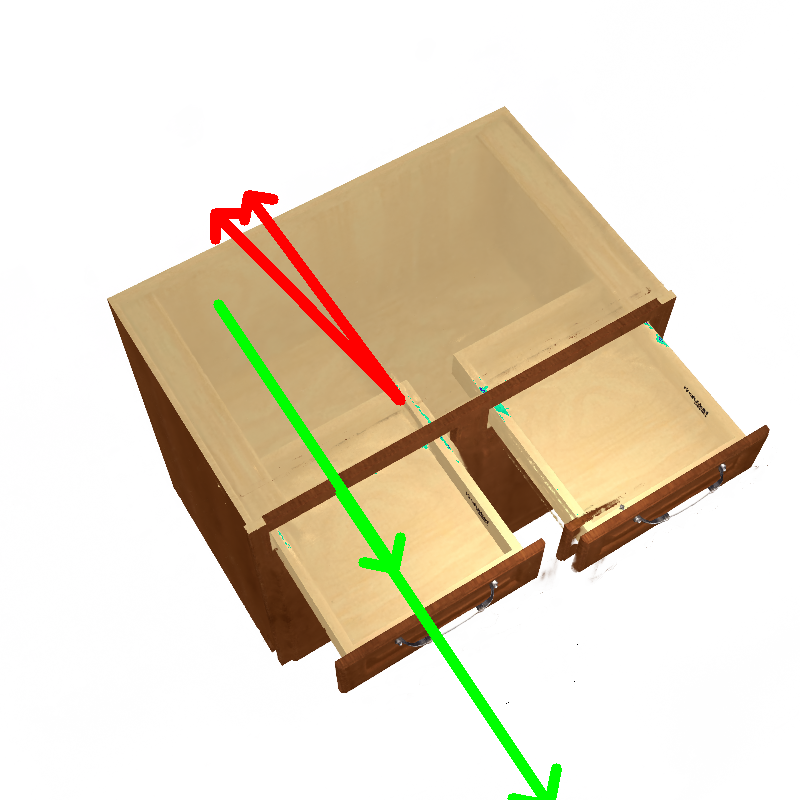}&
        \includegraphics[valign=c, width=0.1\columnwidth]{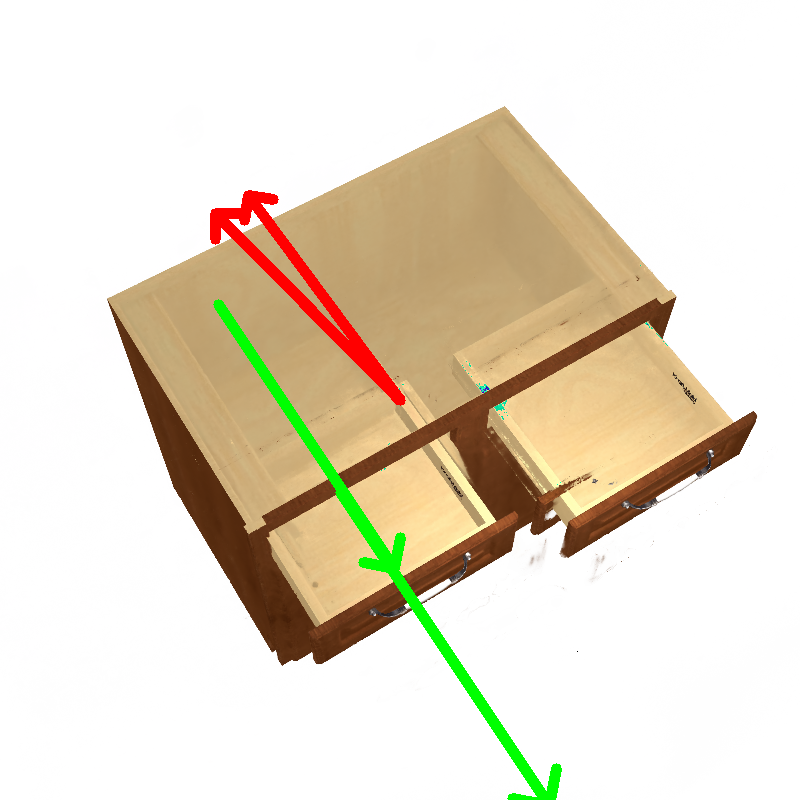}&
        \includegraphics[valign=c, width=0.1\columnwidth]{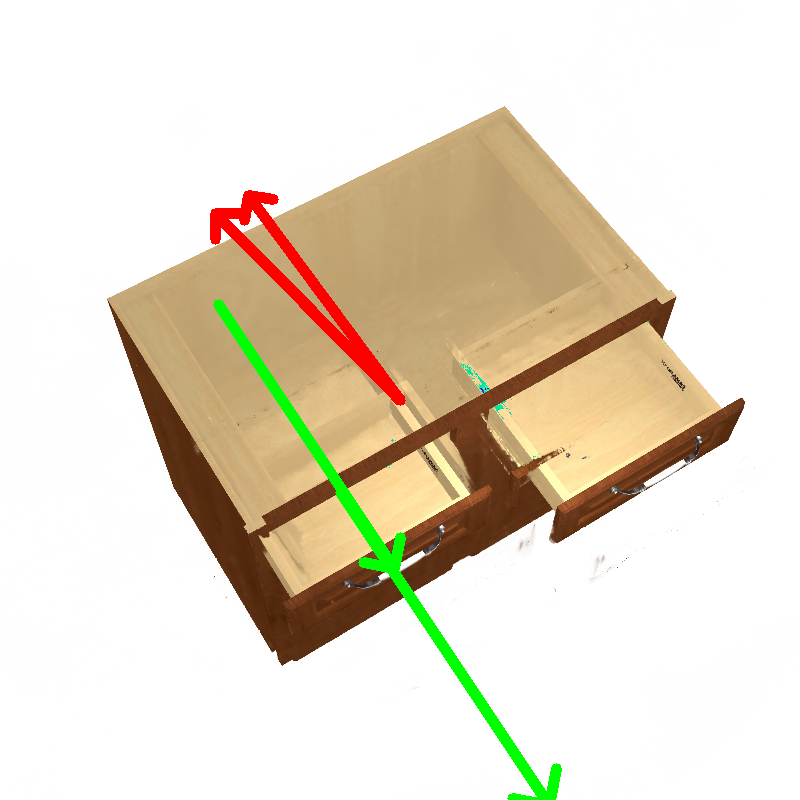}&
        \includegraphics[valign=c, width=0.1\columnwidth]{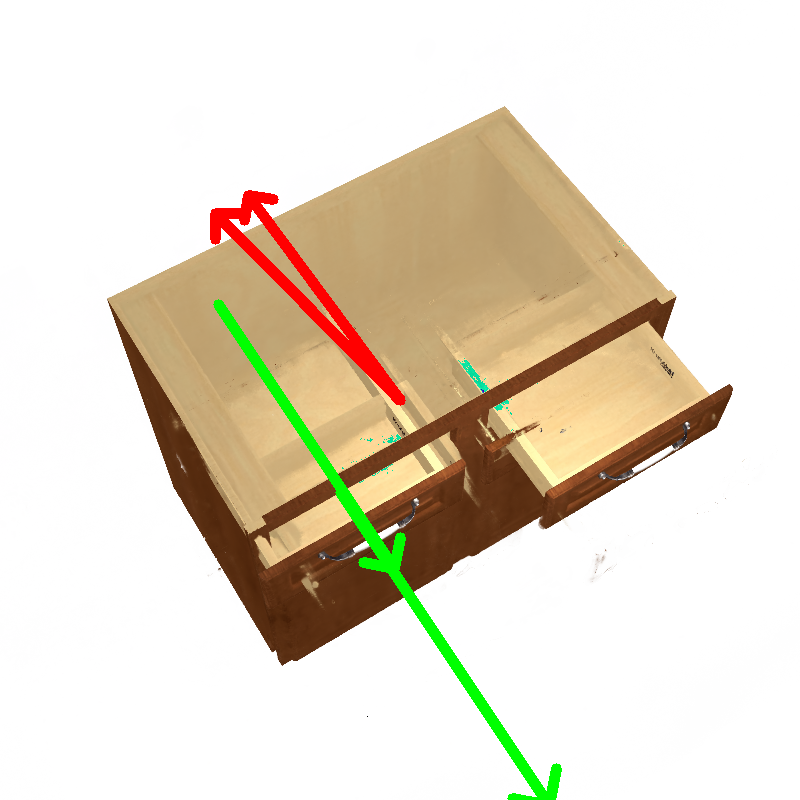}&
        \includegraphics[valign=c, width=0.1\columnwidth]{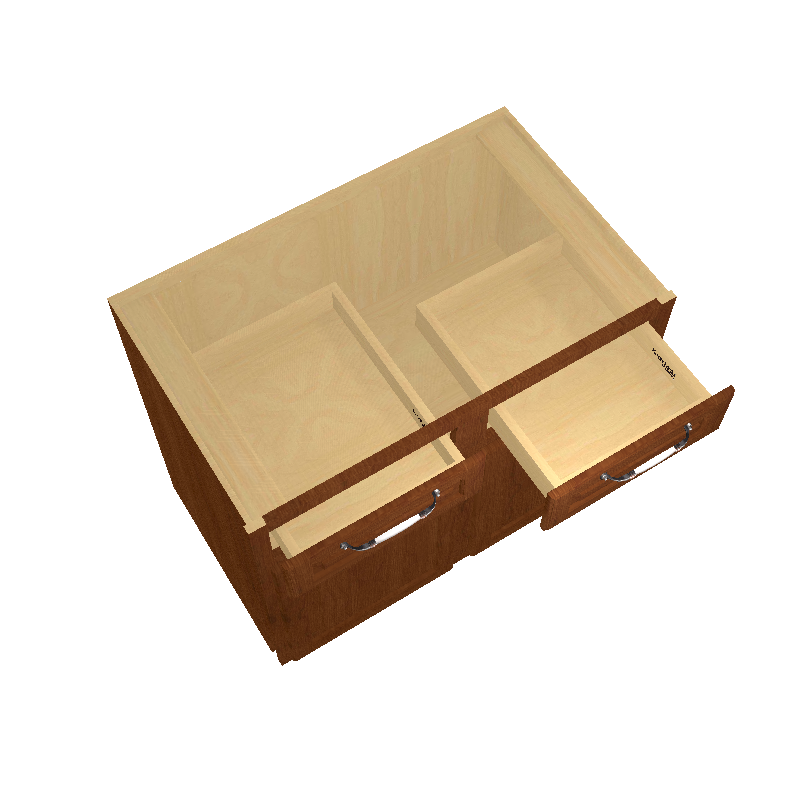}\\
        
        \includegraphics[valign=c, width=0.1\columnwidth]{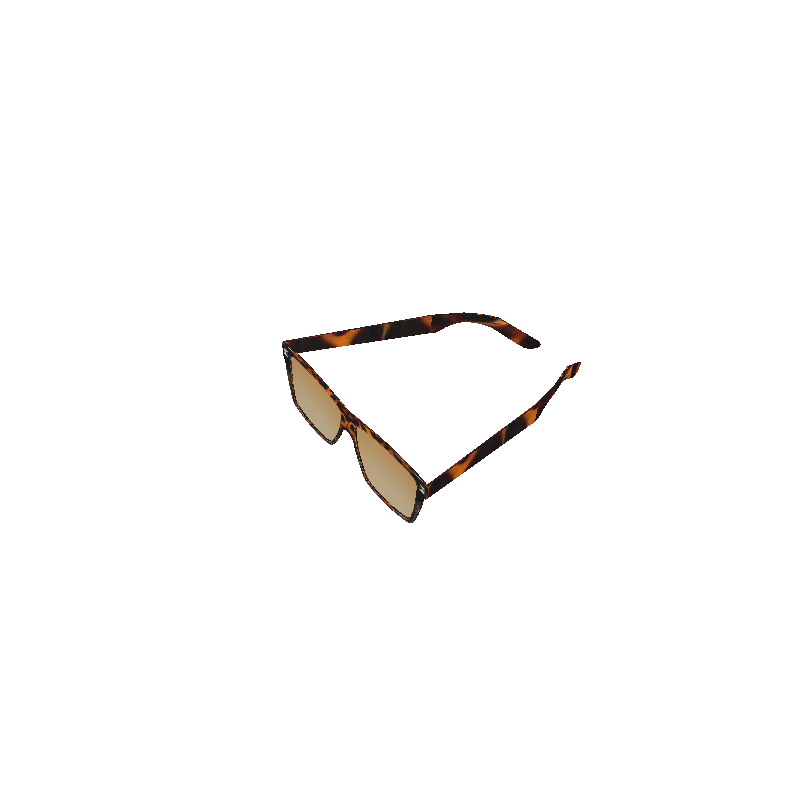}&
        \includegraphics[valign=c, width=0.1\columnwidth]{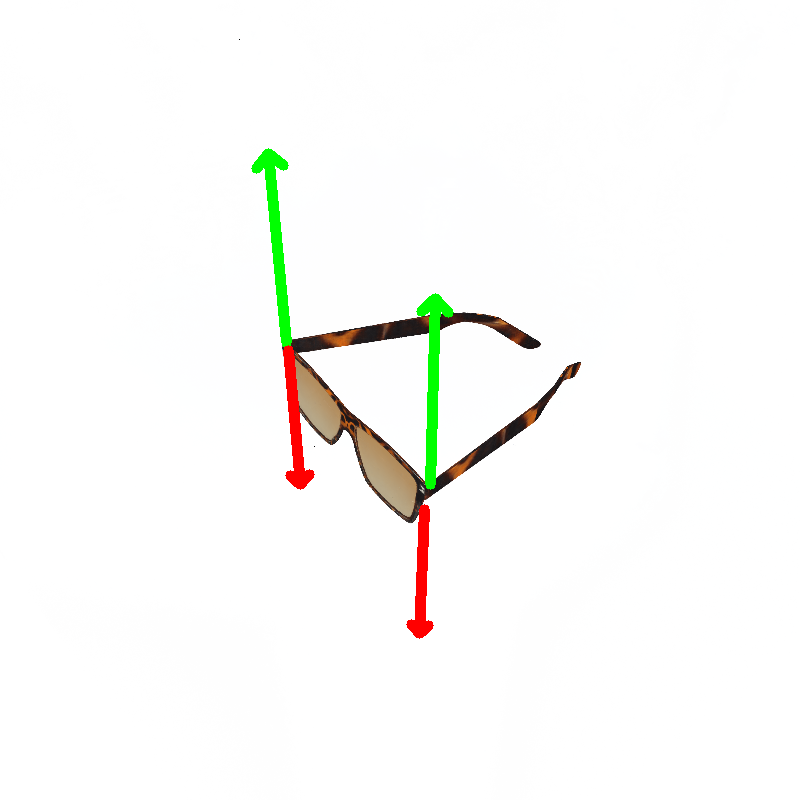}&
        \includegraphics[valign=c, width=0.1\columnwidth]{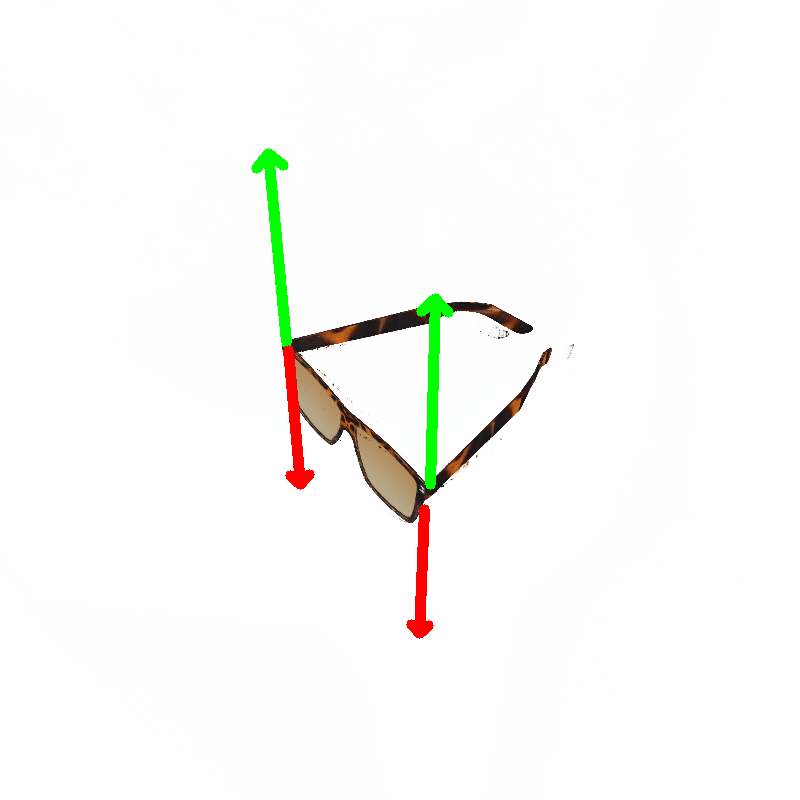}&
        \includegraphics[valign=c, width=0.1\columnwidth]{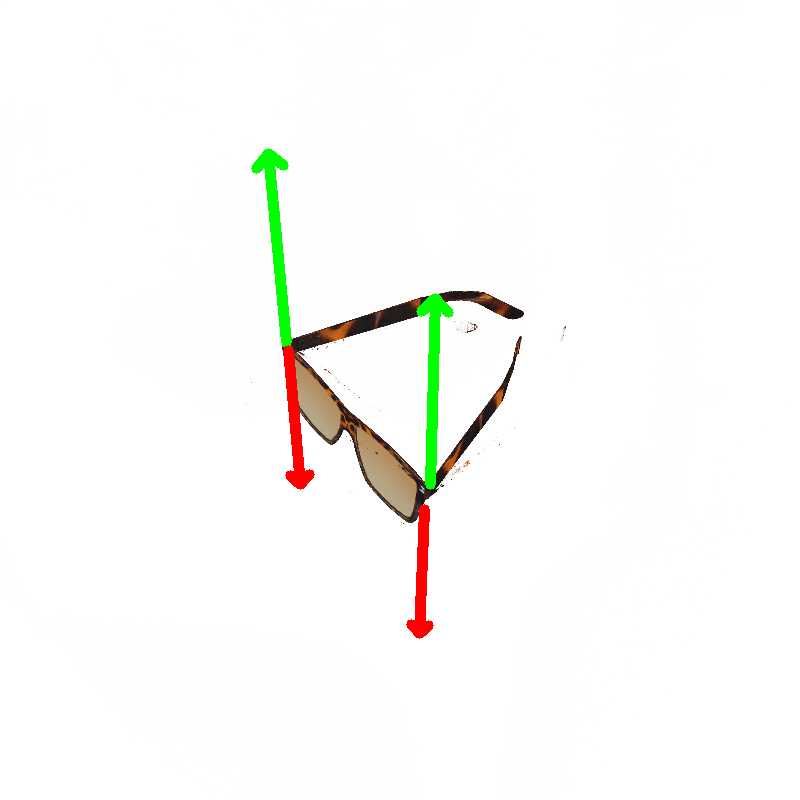}&
        \includegraphics[valign=c, width=0.1\columnwidth]{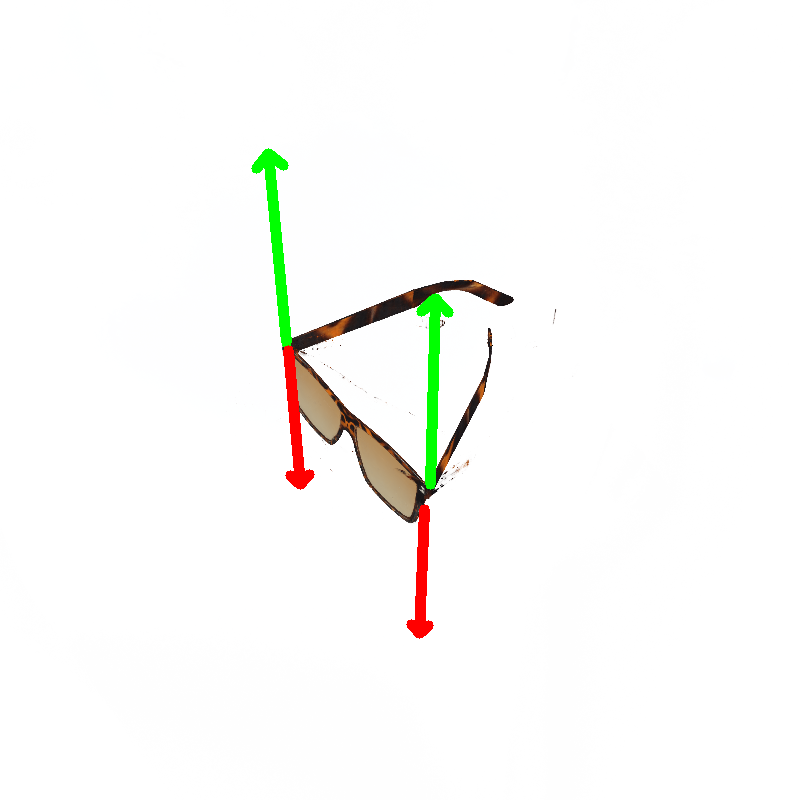}&
        \includegraphics[valign=c, width=0.1\columnwidth]{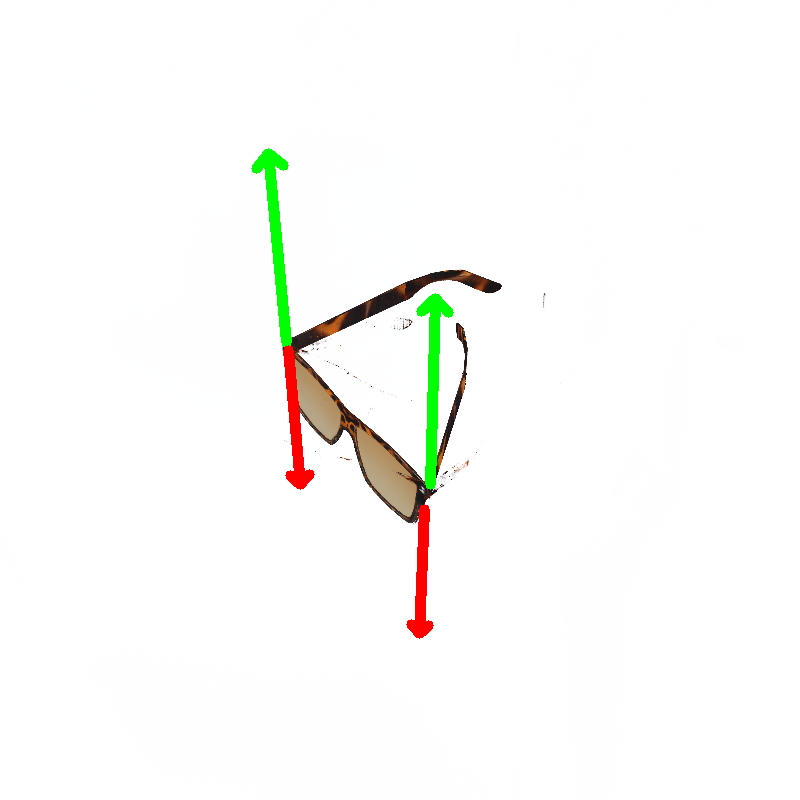}&
        \includegraphics[valign=c, width=0.1\columnwidth]{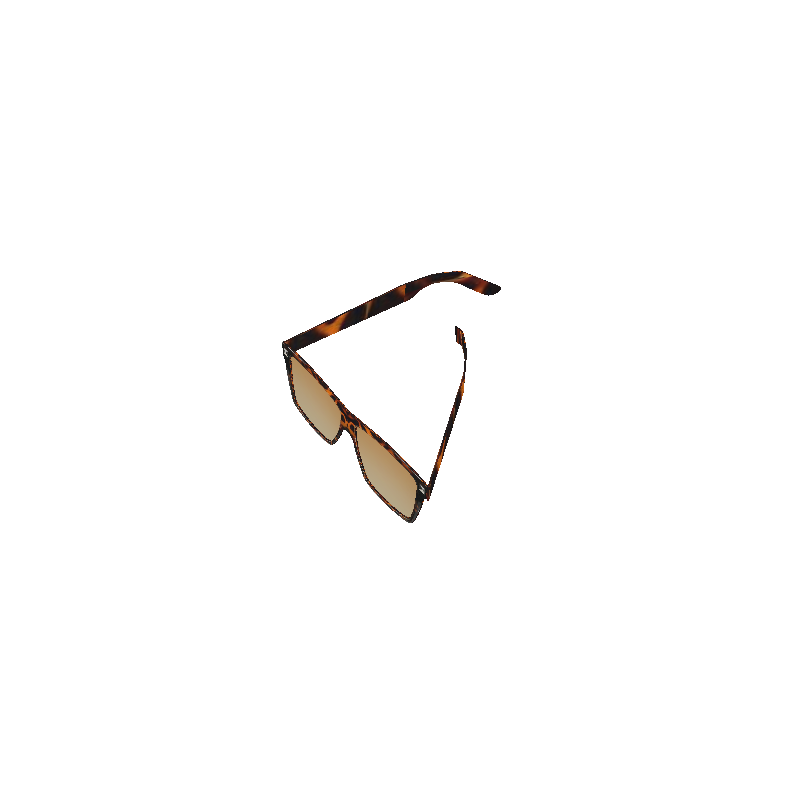}\\
        
        \includegraphics[valign=c, width=0.1\columnwidth]{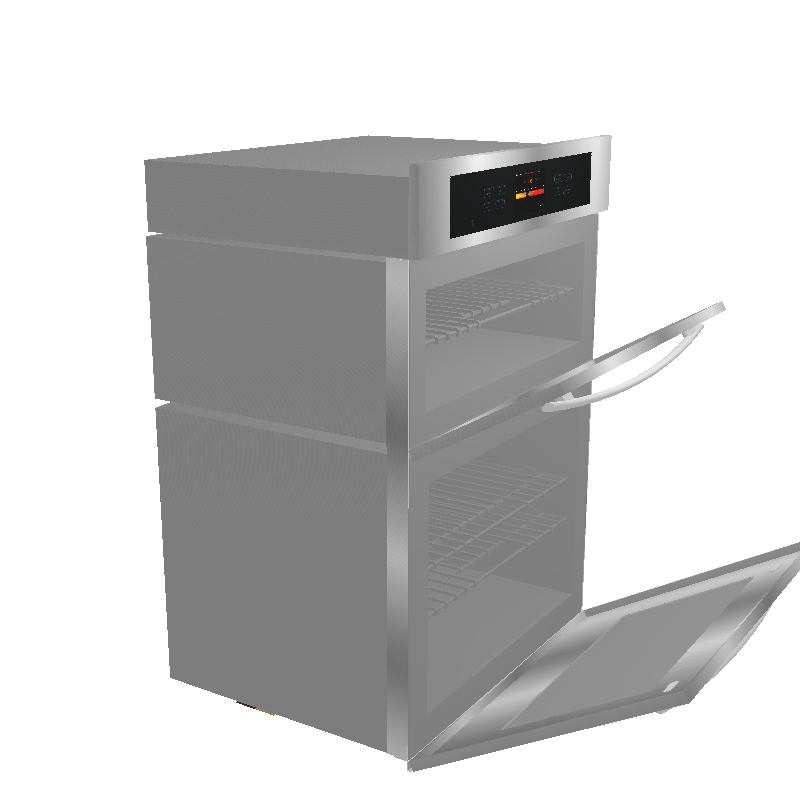}&
        \includegraphics[valign=c, width=0.1\columnwidth]{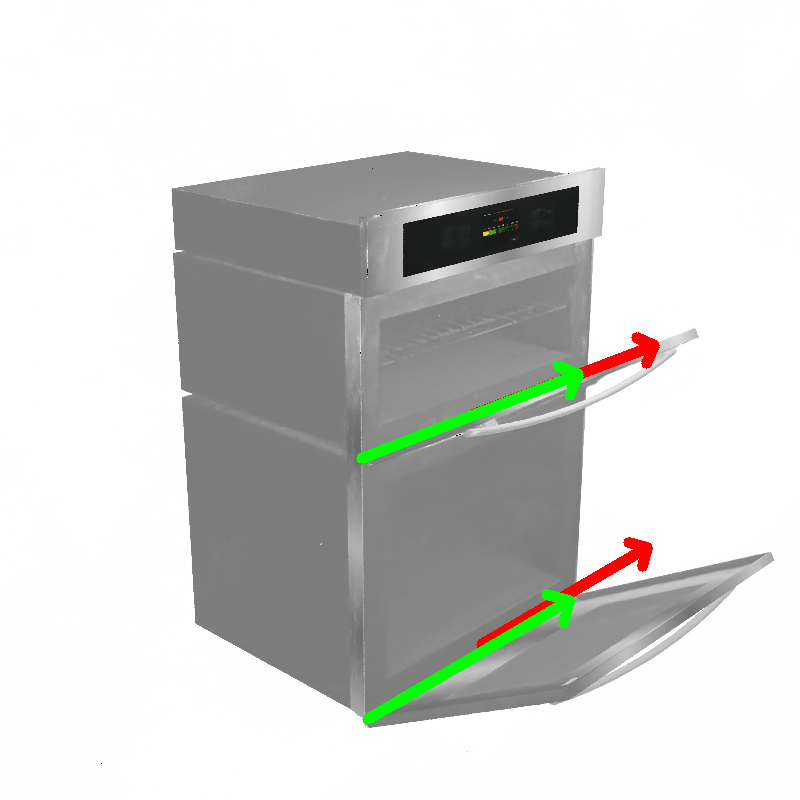}&
        \includegraphics[valign=c, width=0.1\columnwidth]{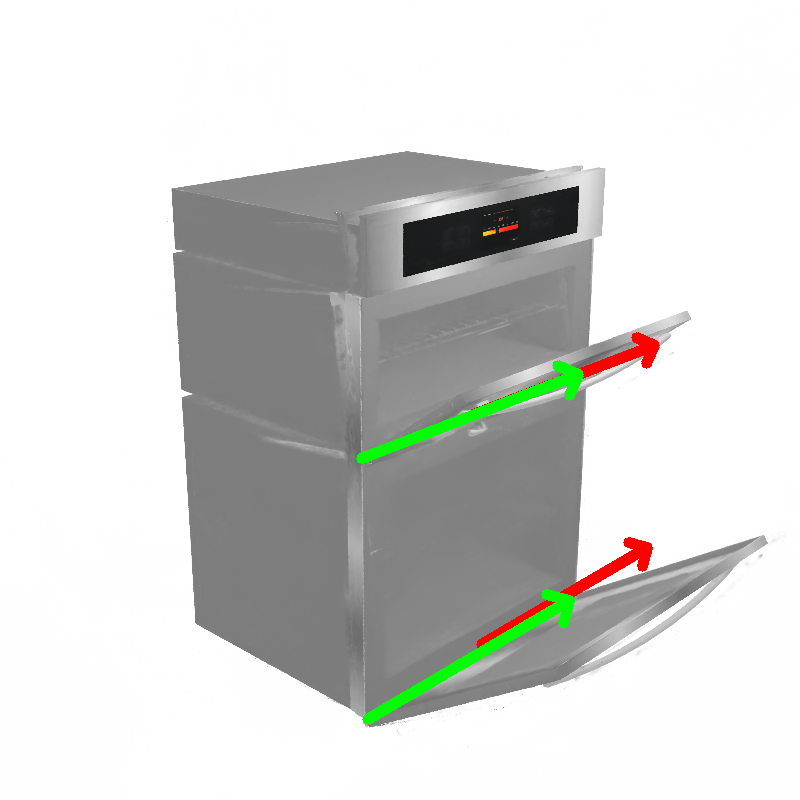}&
        \includegraphics[valign=c, width=0.1\columnwidth]{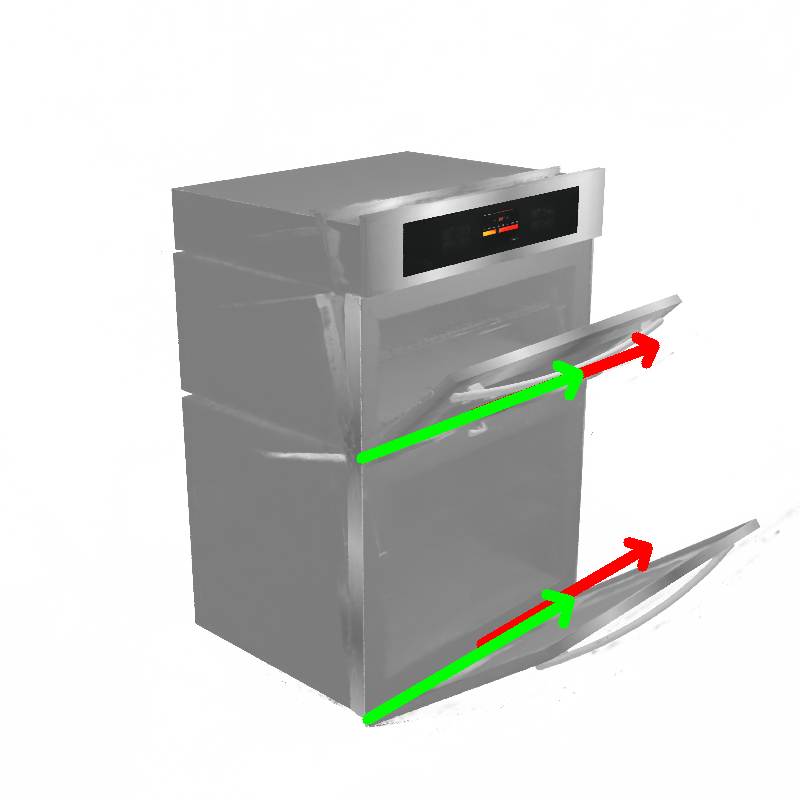}&
        \includegraphics[valign=c, width=0.1\columnwidth]{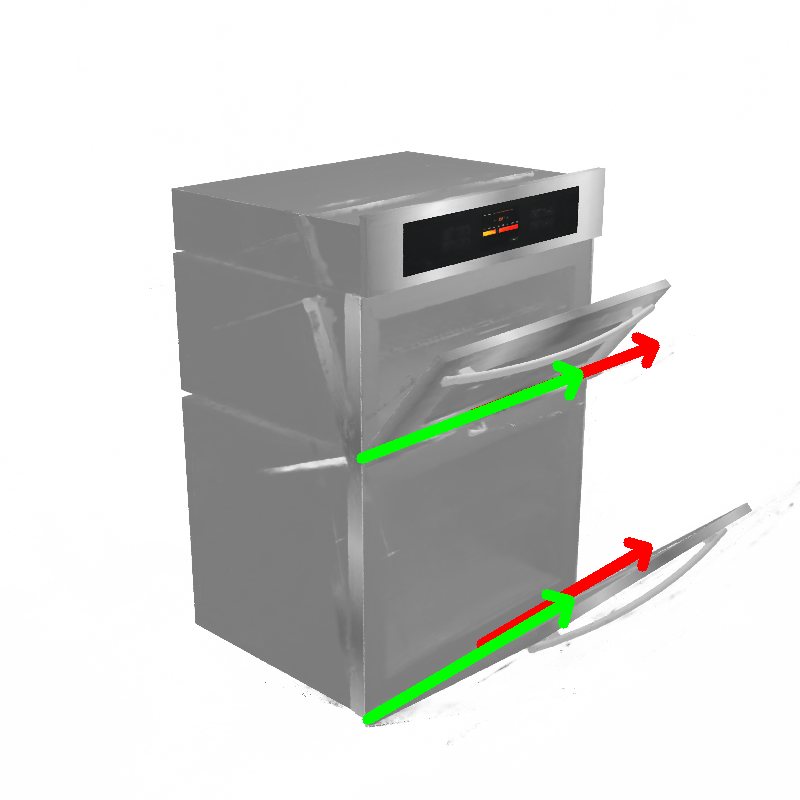}&
        \includegraphics[valign=c, width=0.1\columnwidth]{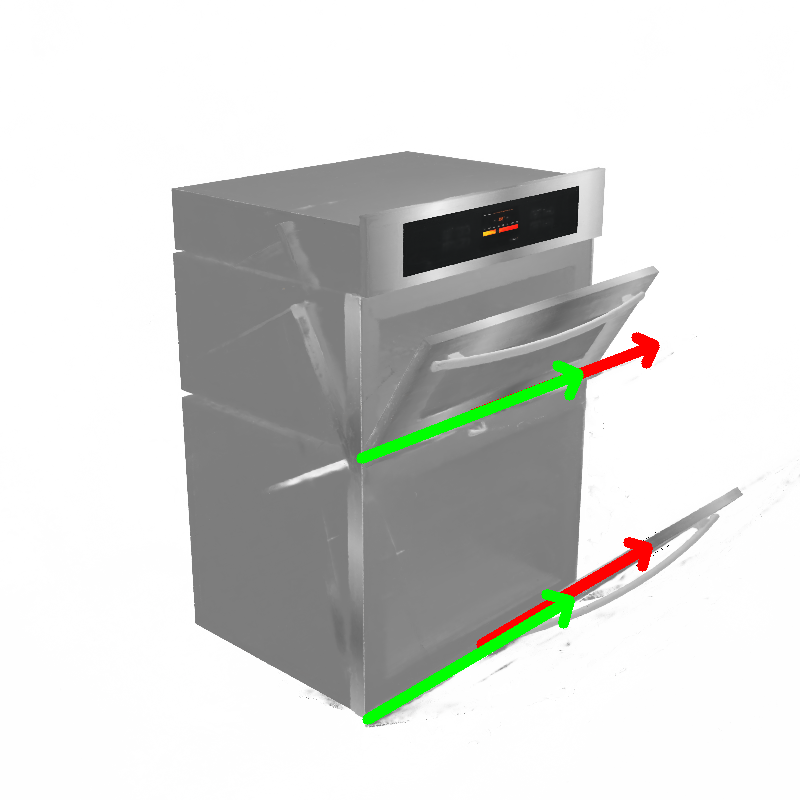}&
        \includegraphics[valign=c, width=0.1\columnwidth]{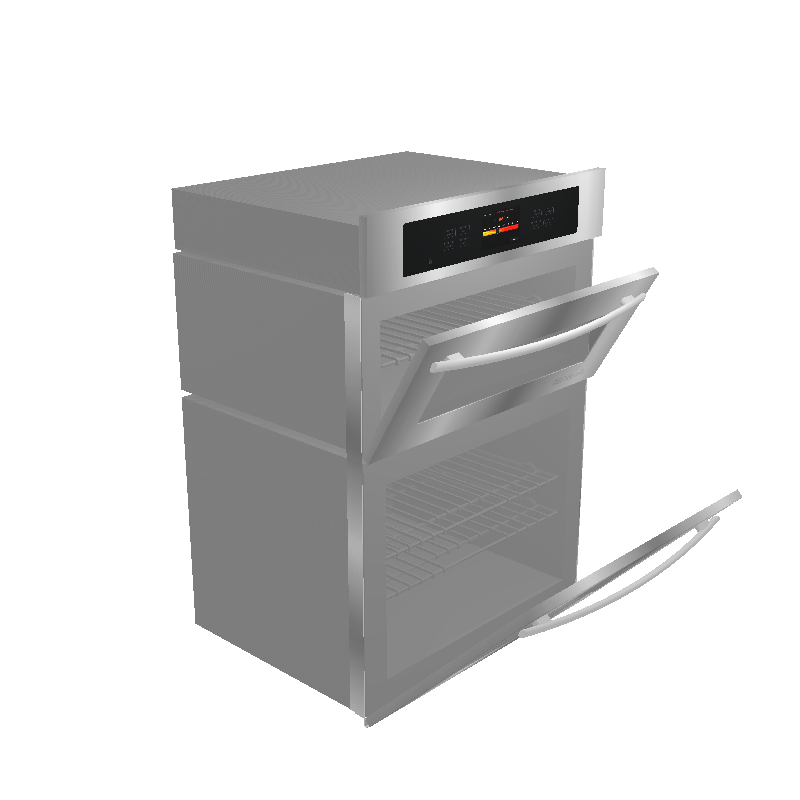}&\\
        \hline
    \end{tblr}
    }
    \caption{\textbf{Articulation interpolation for multiple moving part objects.}}
    \label{fig:appendix_art_interp}
\end{figure}

\begin{figure}
    \centering
    \resizebox{\columnwidth}{!}{%
    \SetTblrInner{rowsep=0pt,colsep=1pt}
    \scriptsize
    \begin{tblr}{cccccc}
   \SetCell[c=6]{c}{ Novel articulation synthesis}& &&&&
        \\
        
        \includegraphics[valign=c, width=0.1\columnwidth]{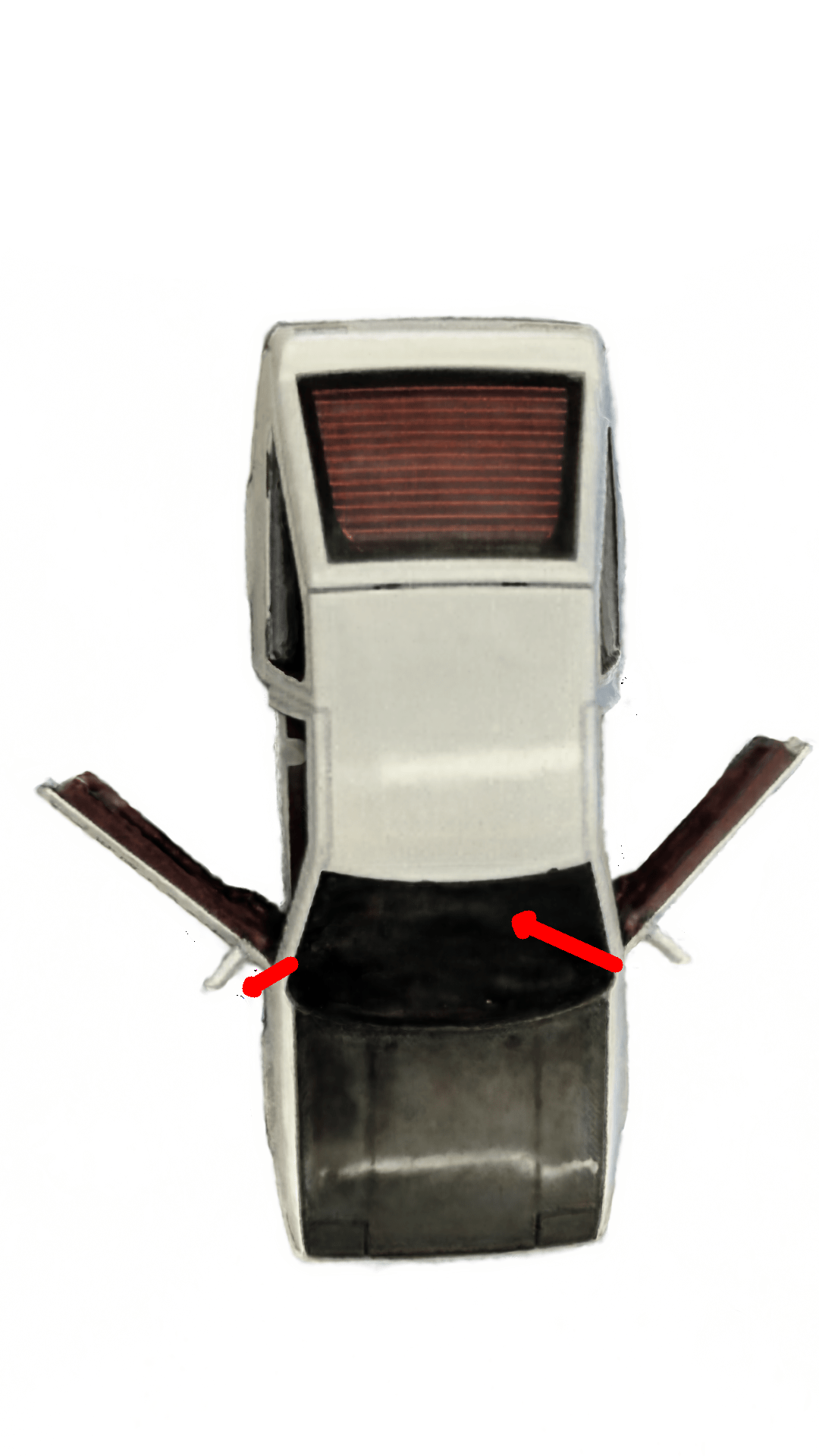}&
        \includegraphics[valign=c, width=0.1\columnwidth]{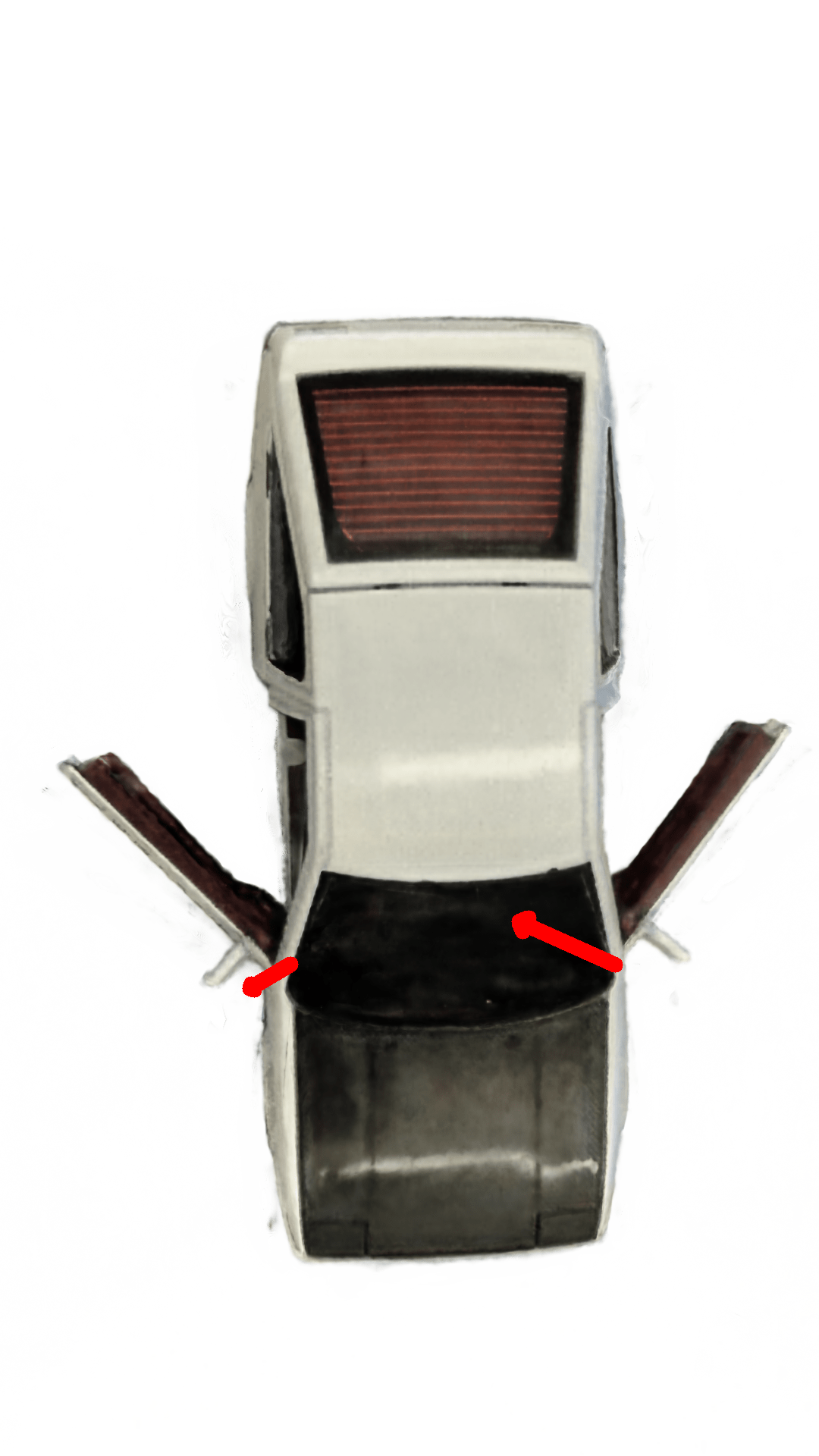}&
        \includegraphics[valign=c, width=0.1\columnwidth]{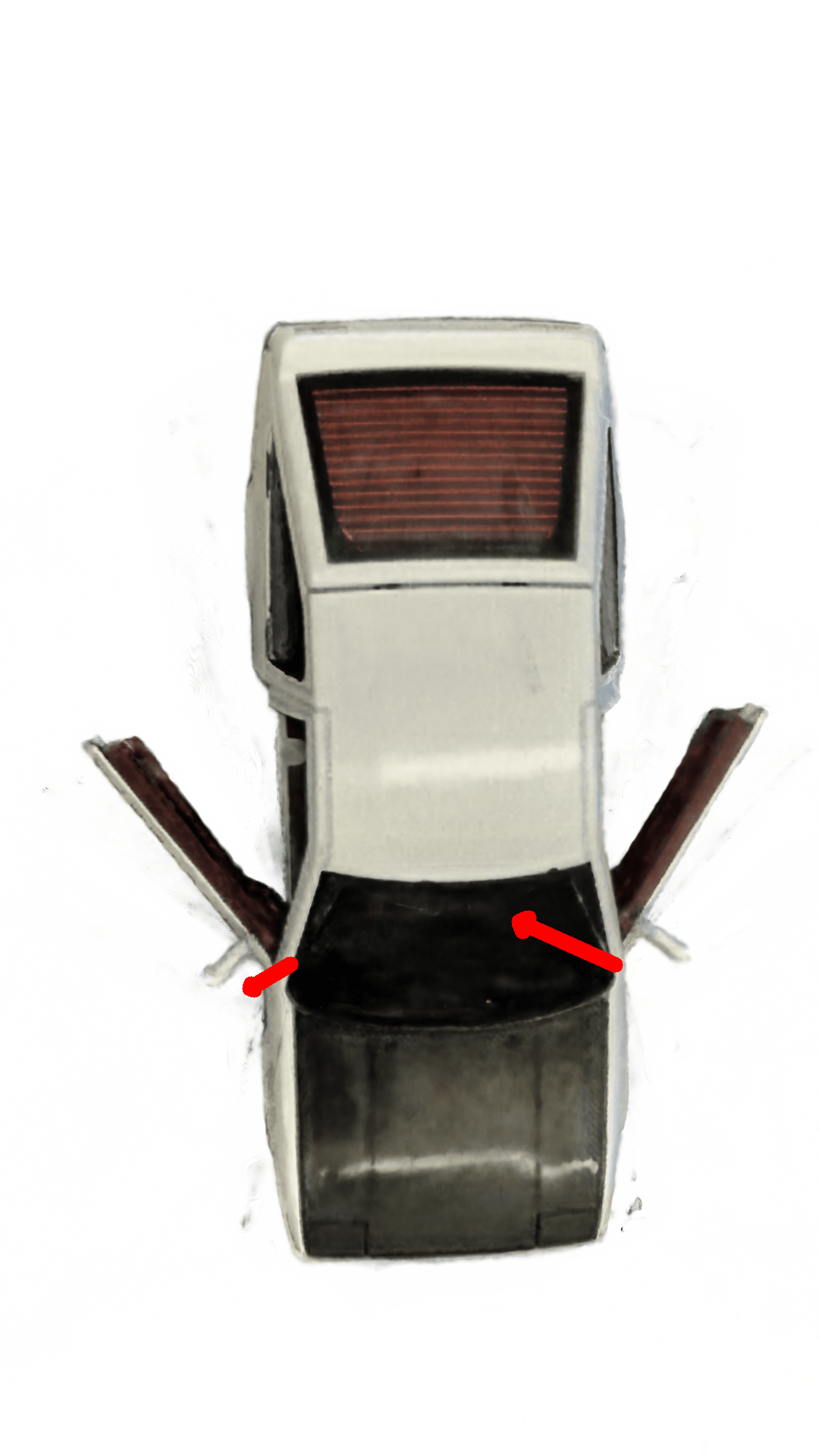}&
        \includegraphics[valign=c, width=0.1\columnwidth]{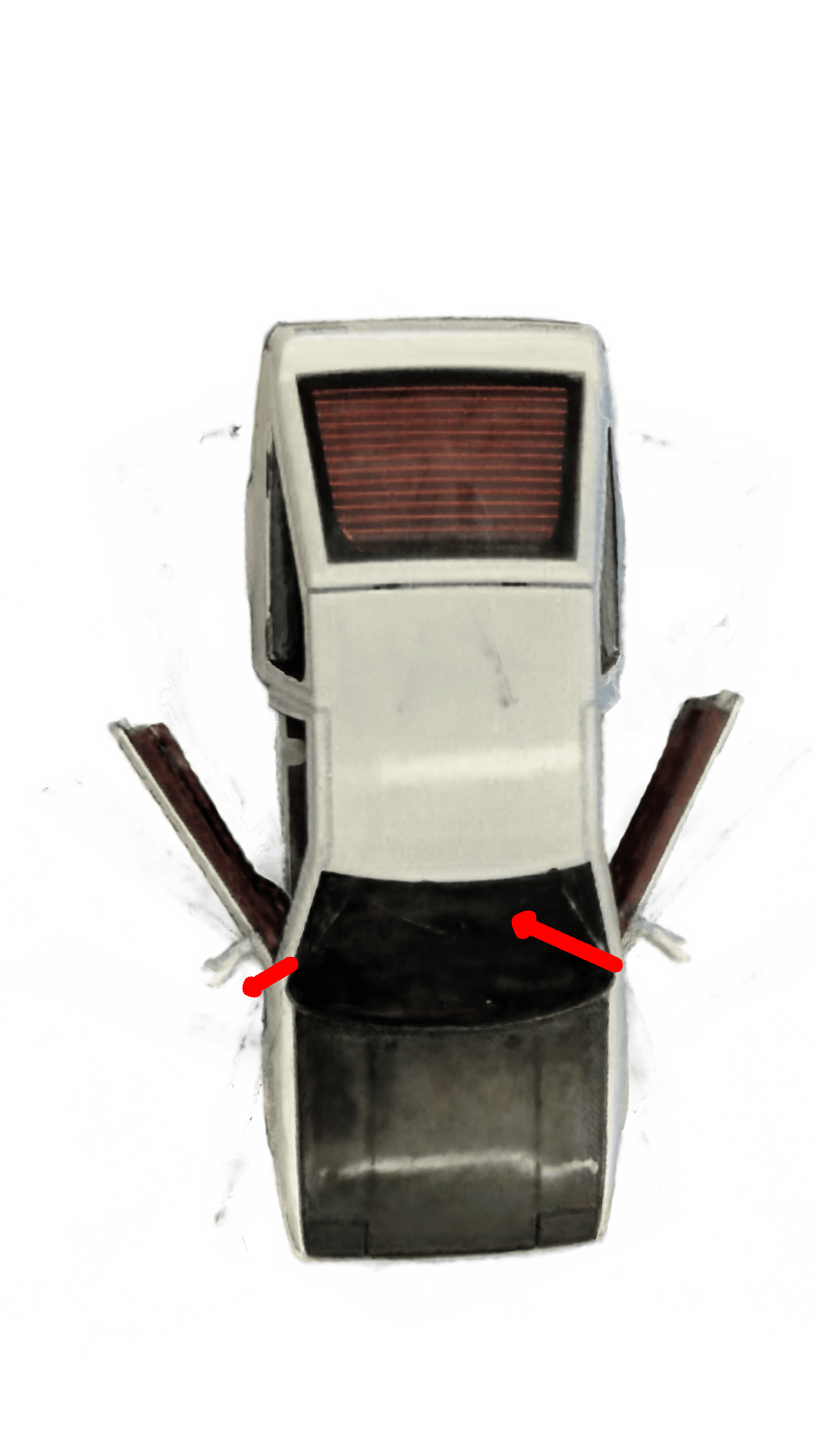}&
        \includegraphics[valign=c, width=0.1\columnwidth]{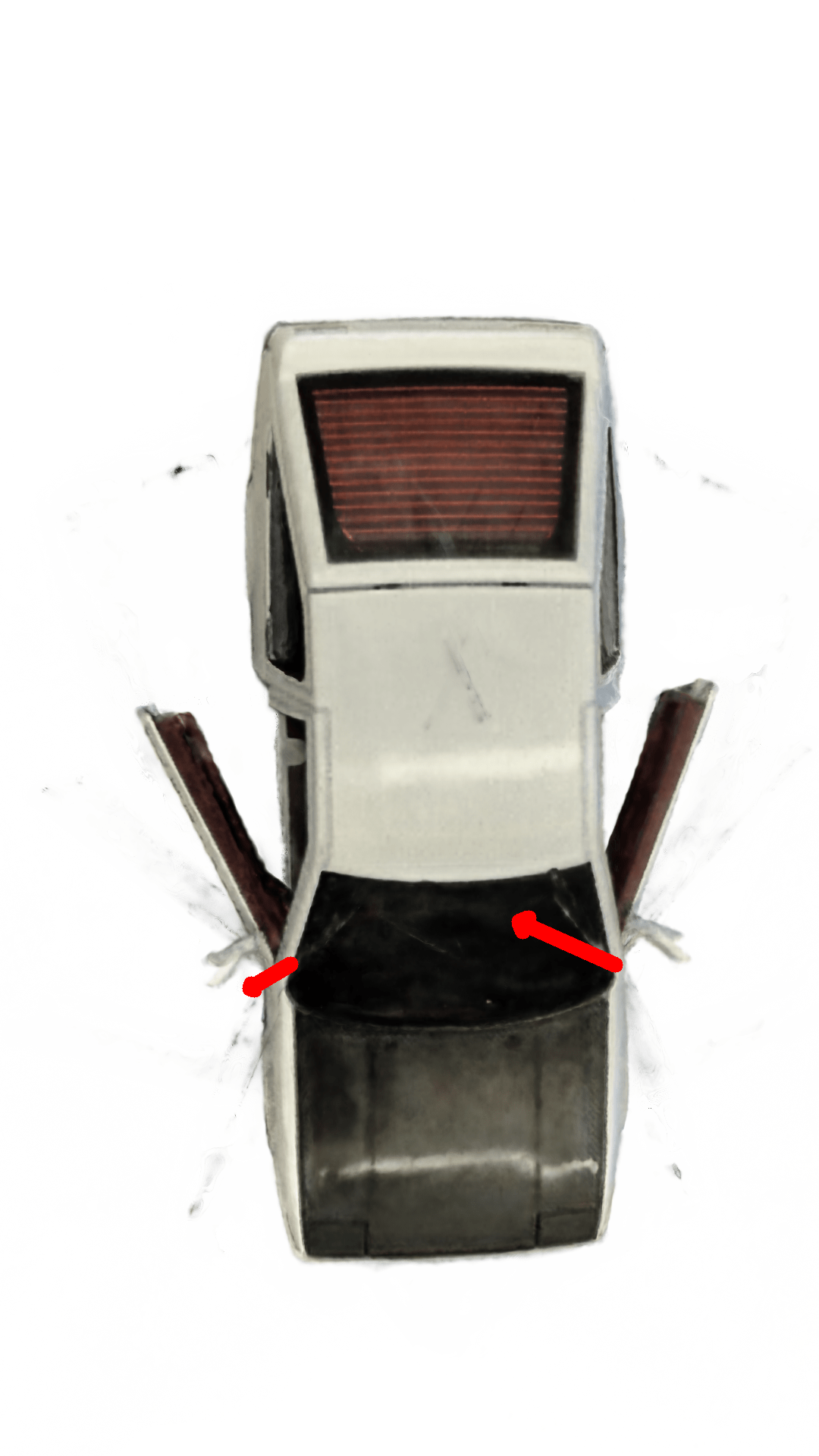}&
        \includegraphics[valign=c, width=0.1\columnwidth]{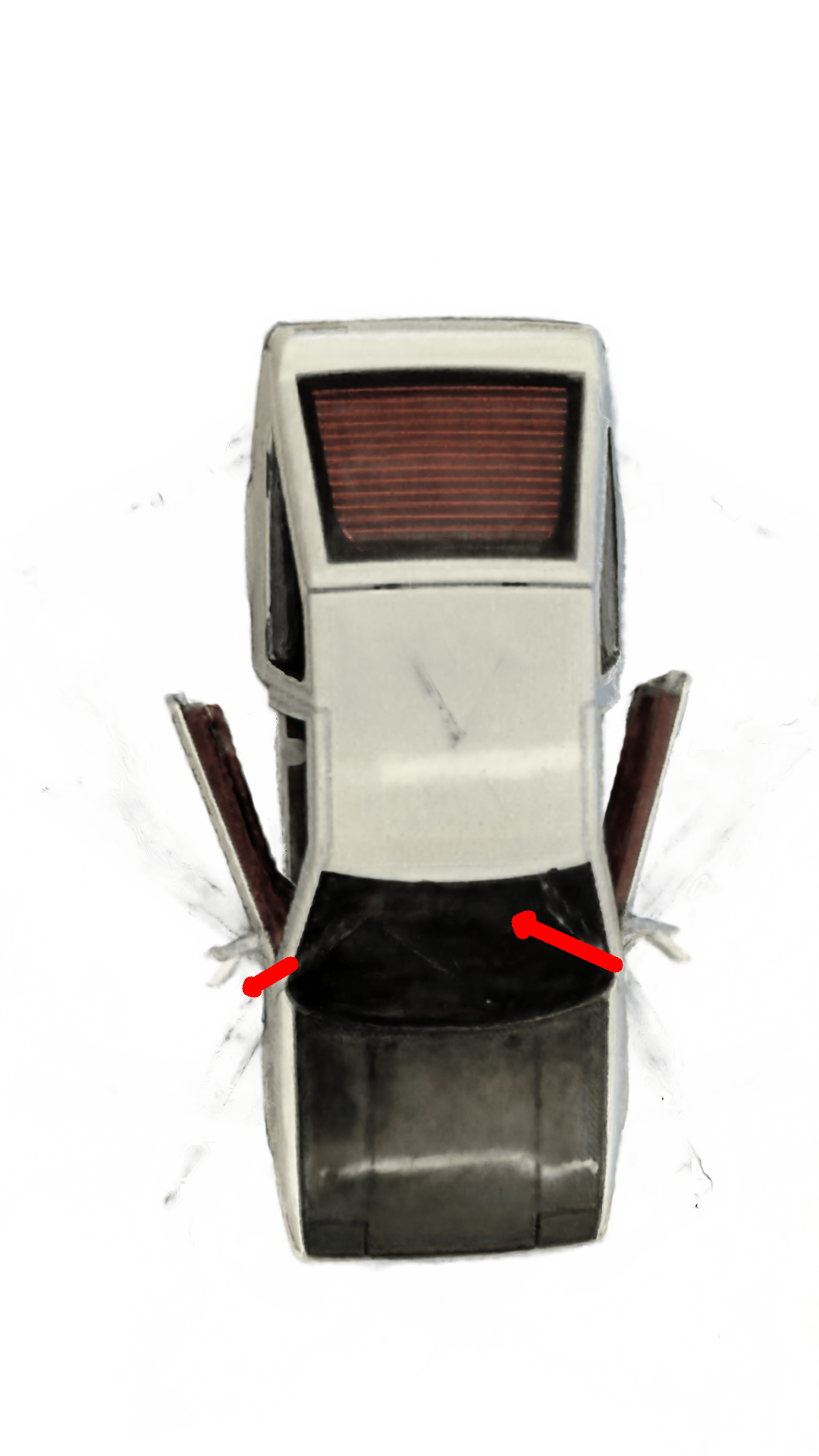}\\
        
        \includegraphics[valign=c, width=0.1\columnwidth]{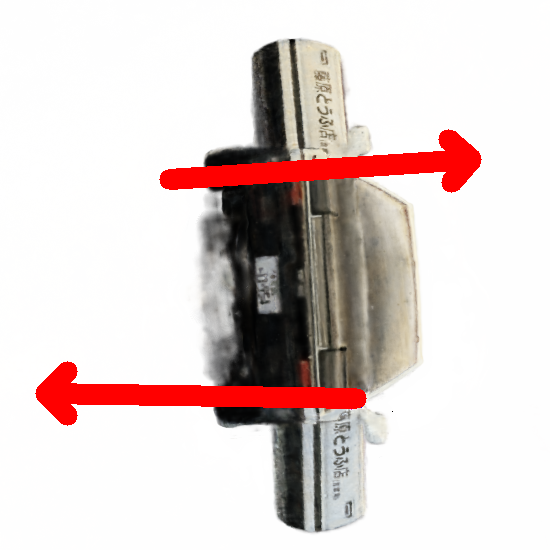}&
        \includegraphics[valign=c, width=0.1\columnwidth]{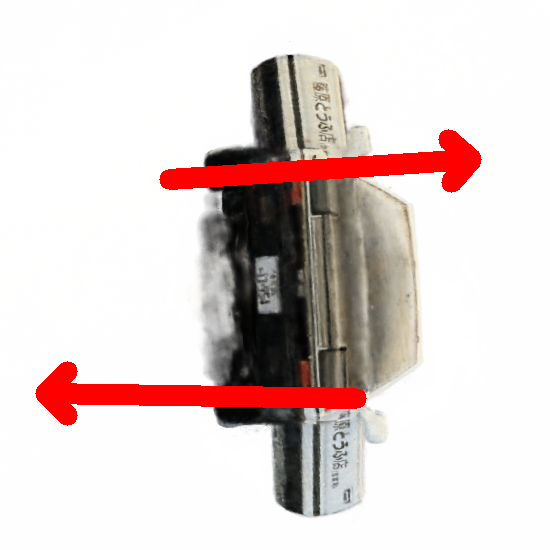}&
        \includegraphics[valign=c, width=0.1\columnwidth]{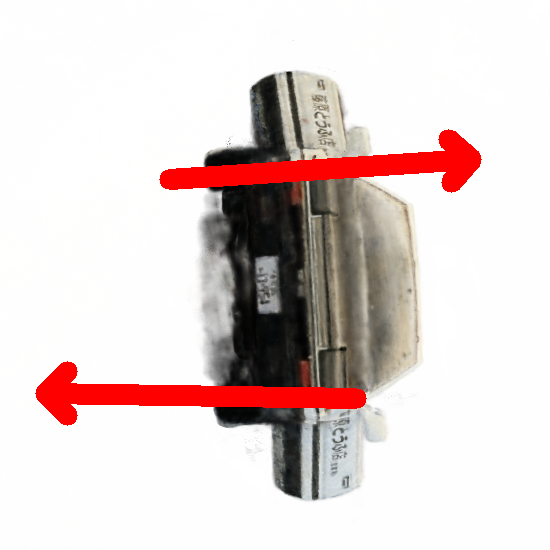}&
        \includegraphics[valign=c, width=0.1\columnwidth]{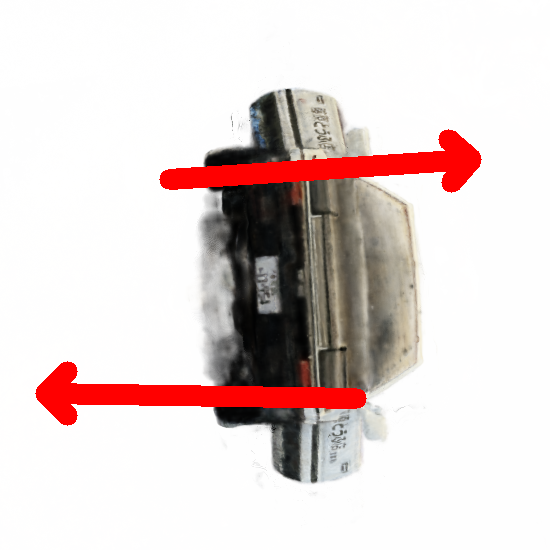}&
        \includegraphics[valign=c, width=0.1\columnwidth]{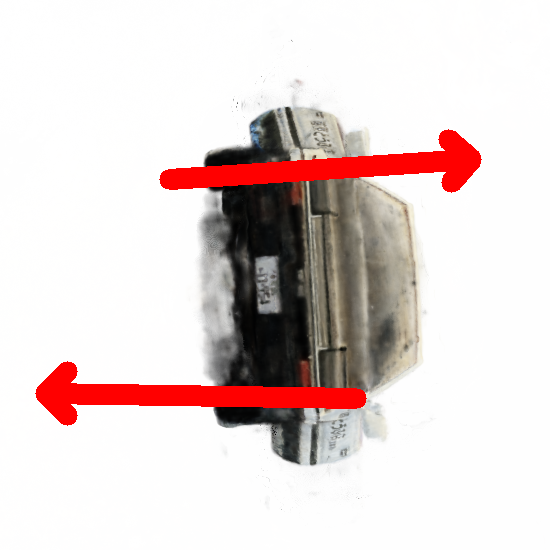}&
        \includegraphics[valign=c, width=0.1\columnwidth]{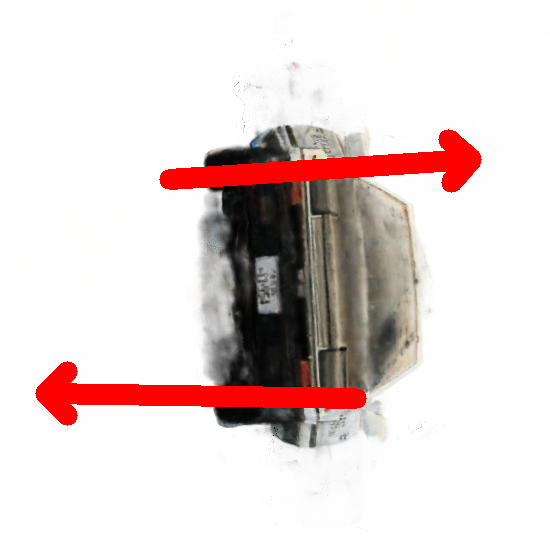}\\
        
        \includegraphics[valign=c, width=0.1\columnwidth]{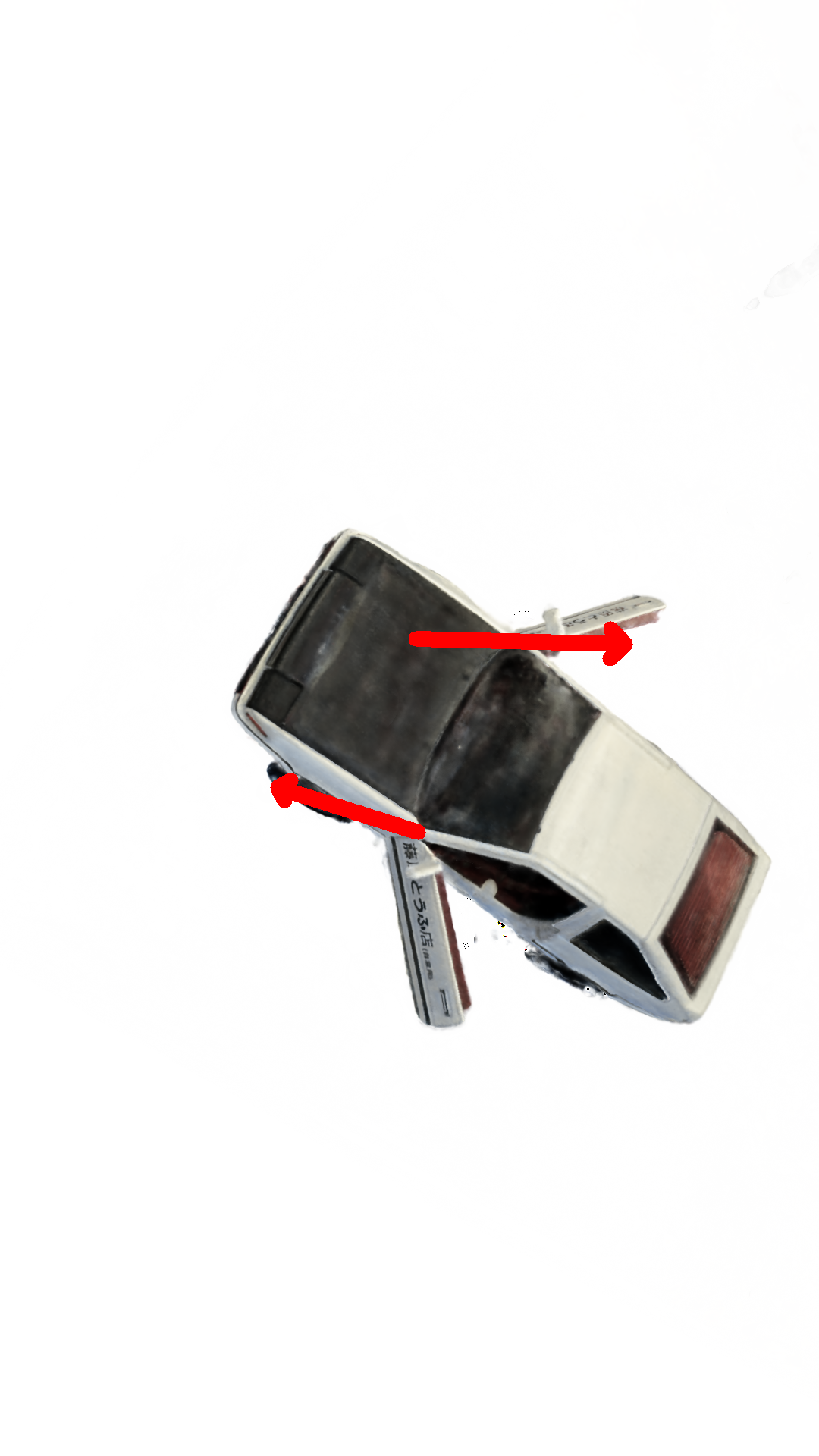}&
        \includegraphics[valign=c, width=0.1\columnwidth]{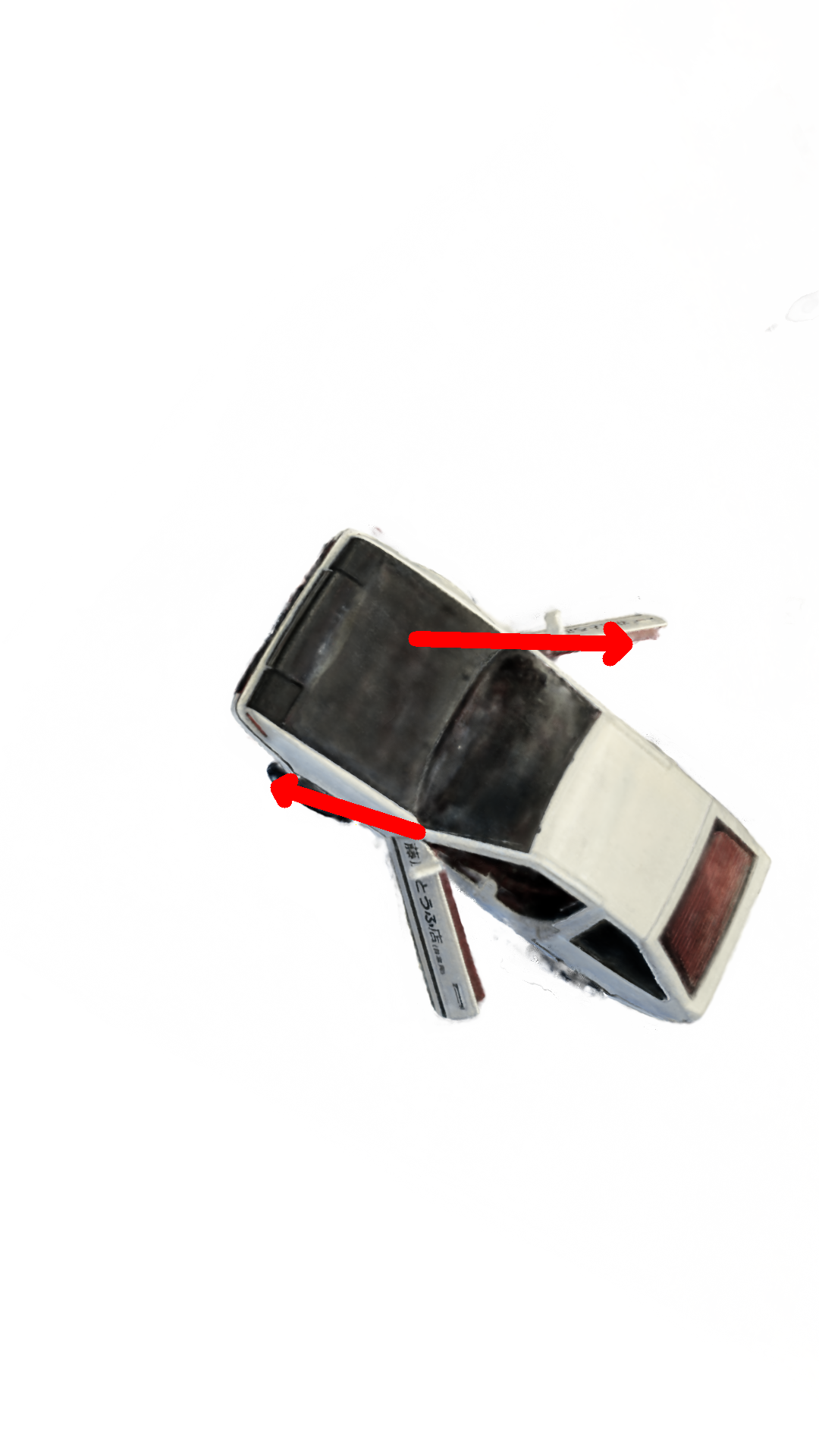}&
        \includegraphics[valign=c, width=0.1\columnwidth]{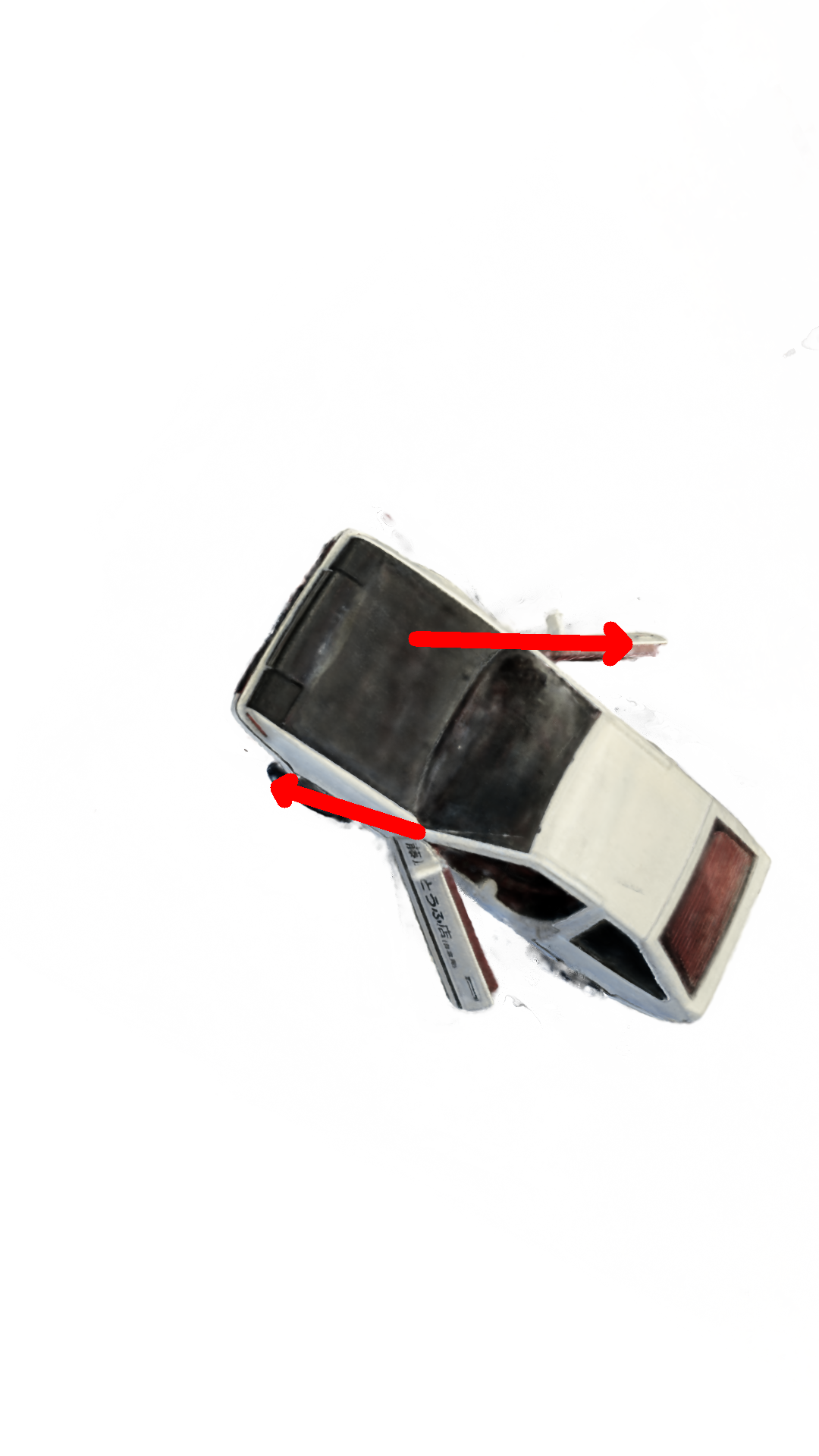}&
        \includegraphics[valign=c, width=0.1\columnwidth]{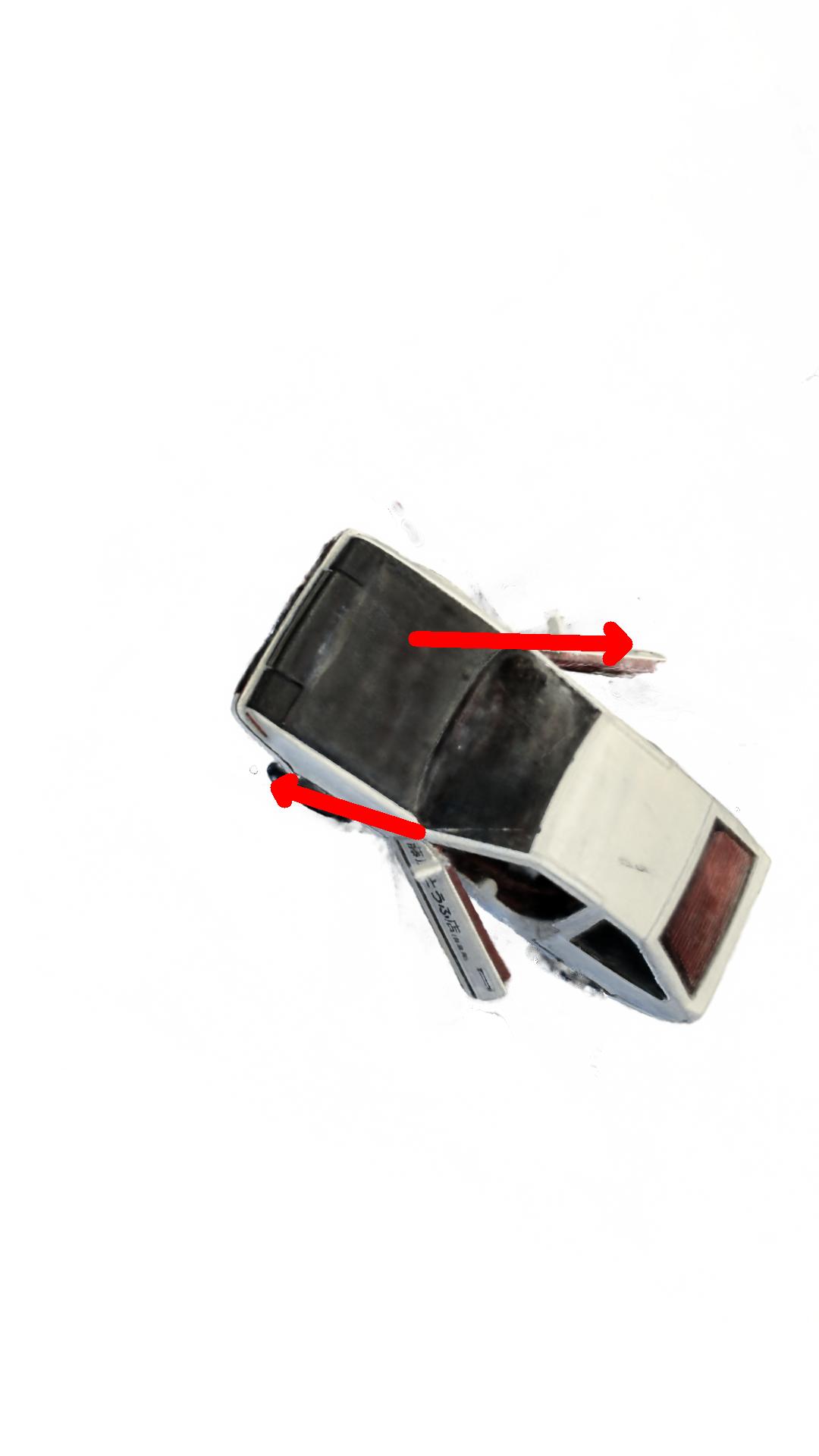}&
        \includegraphics[valign=c, width=0.1\columnwidth]{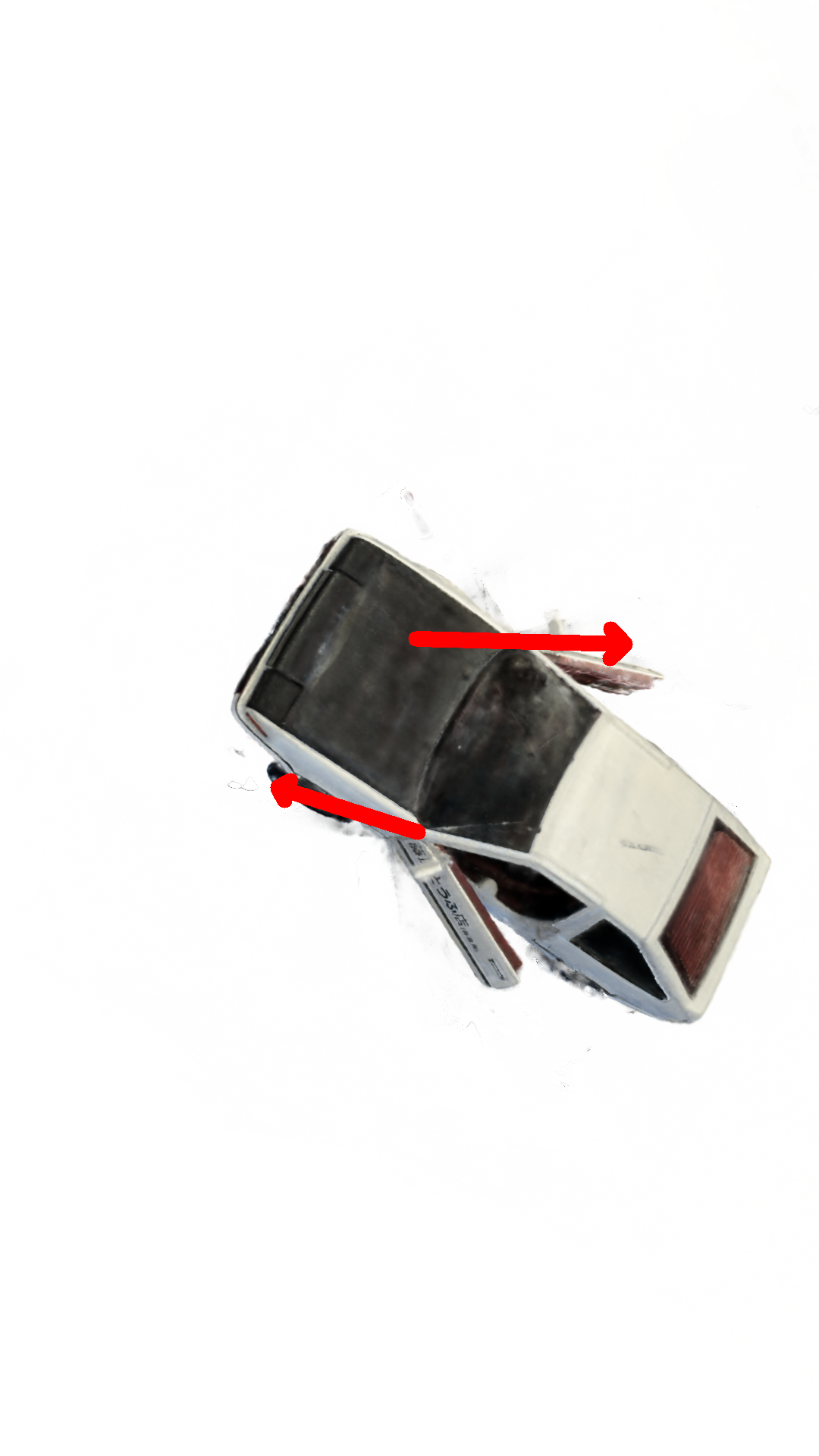}&
        \includegraphics[valign=c, width=0.1\columnwidth]{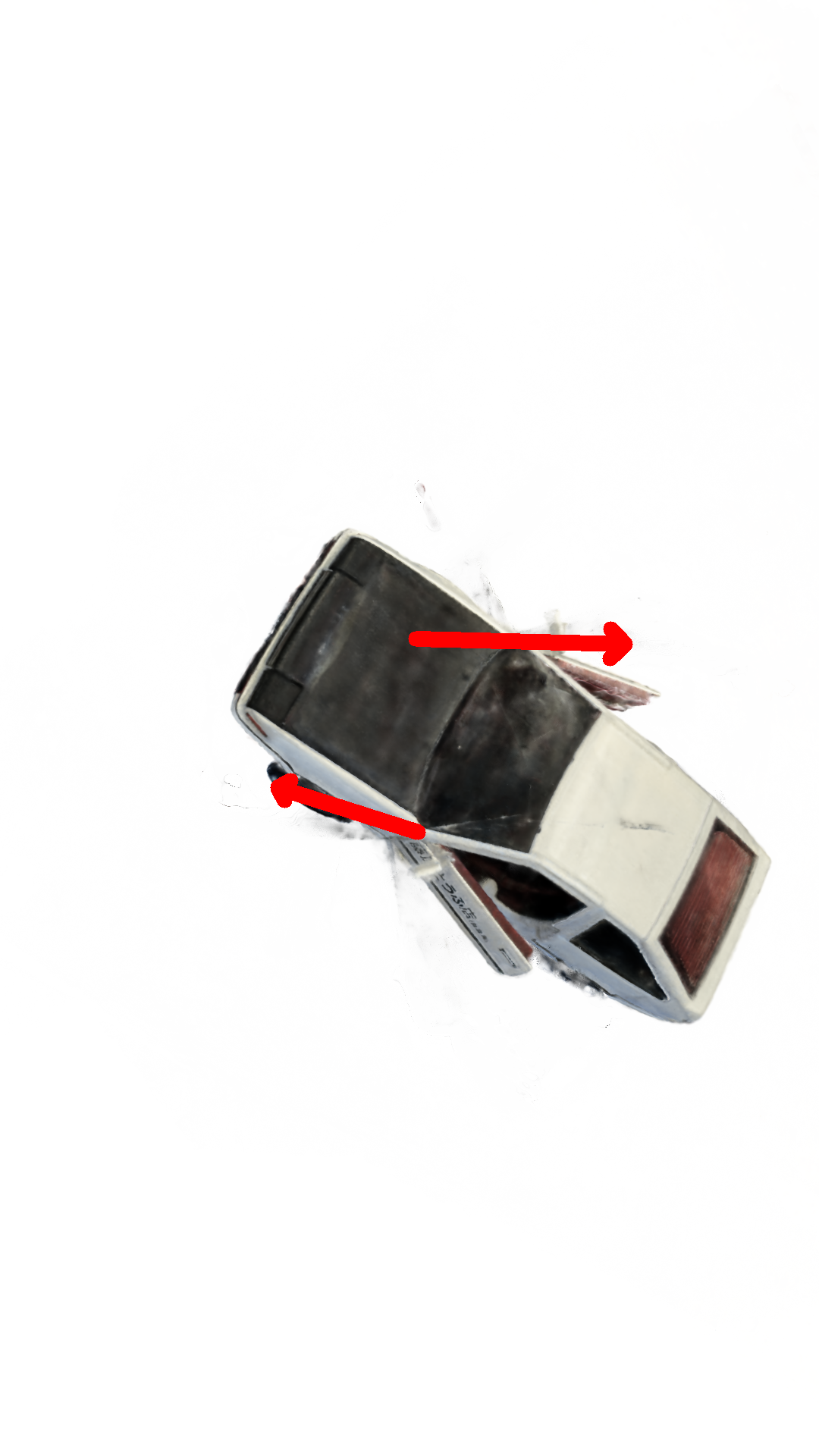}\\
        \hline
        \SetCell[c=6]{c}{Segmentation}\\
        \includegraphics[valign=c, width=0.1\columnwidth]{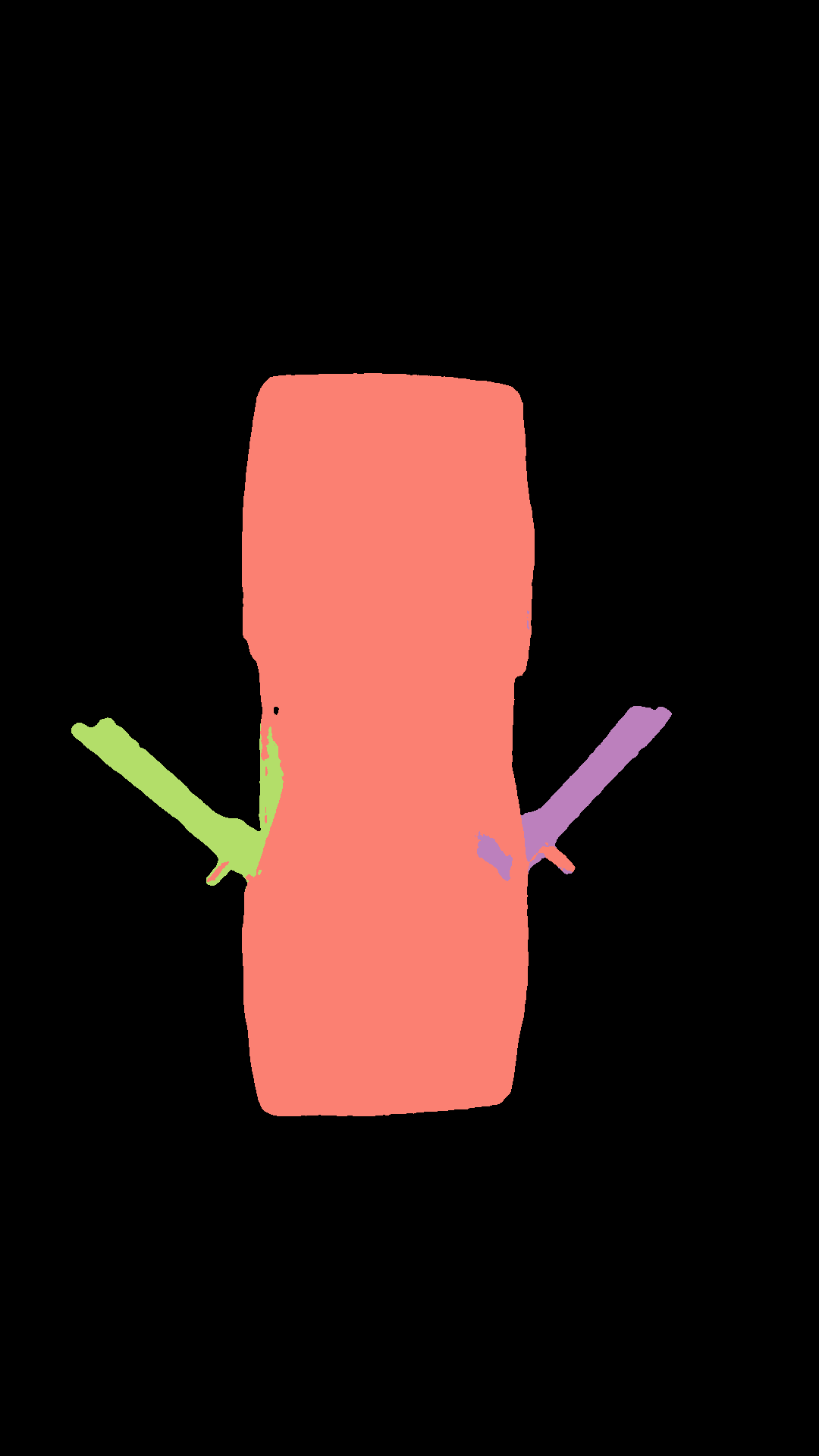}&
        \includegraphics[valign=c, width=0.1\columnwidth]{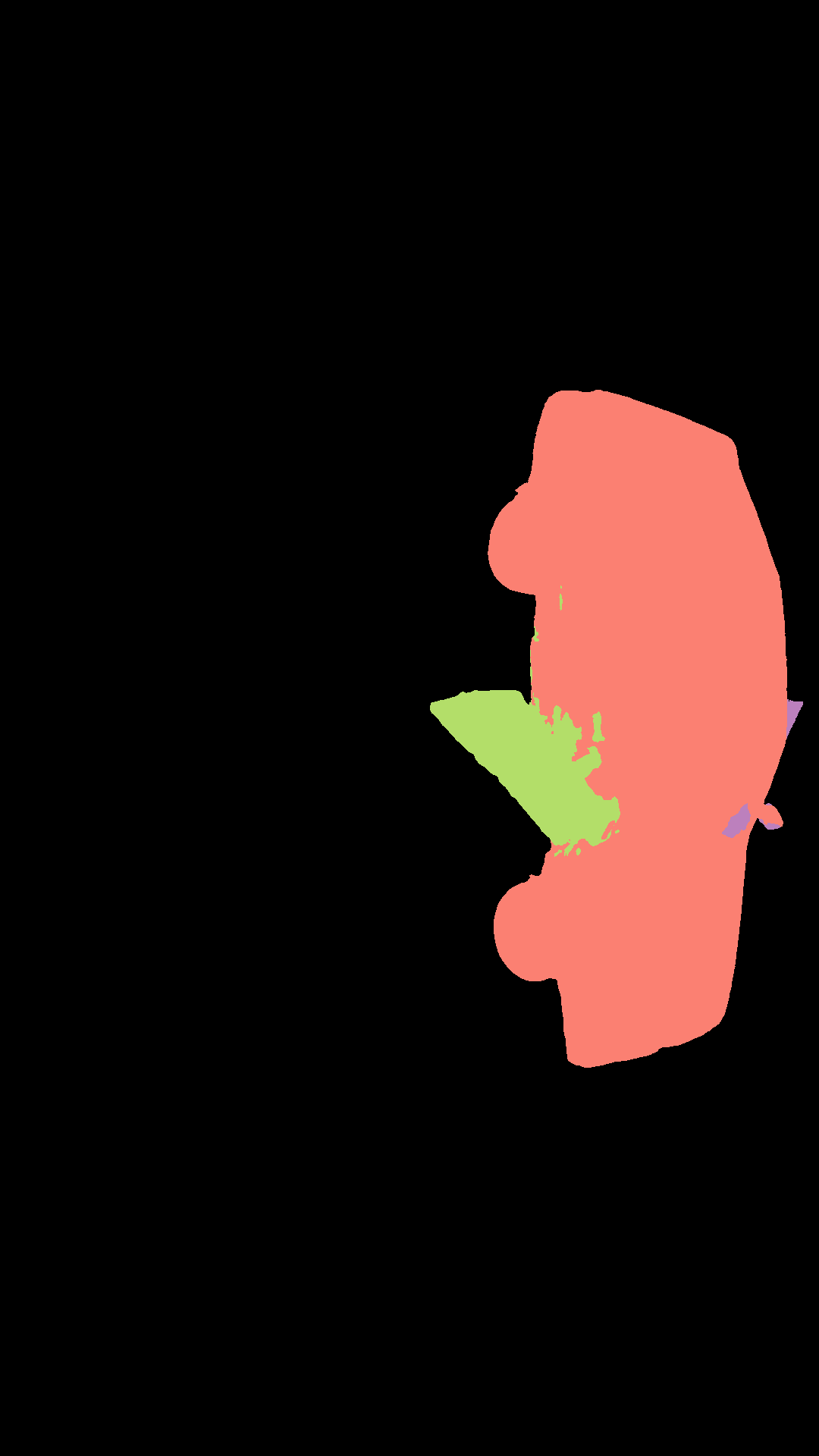}&
        \includegraphics[valign=c, width=0.1\columnwidth]{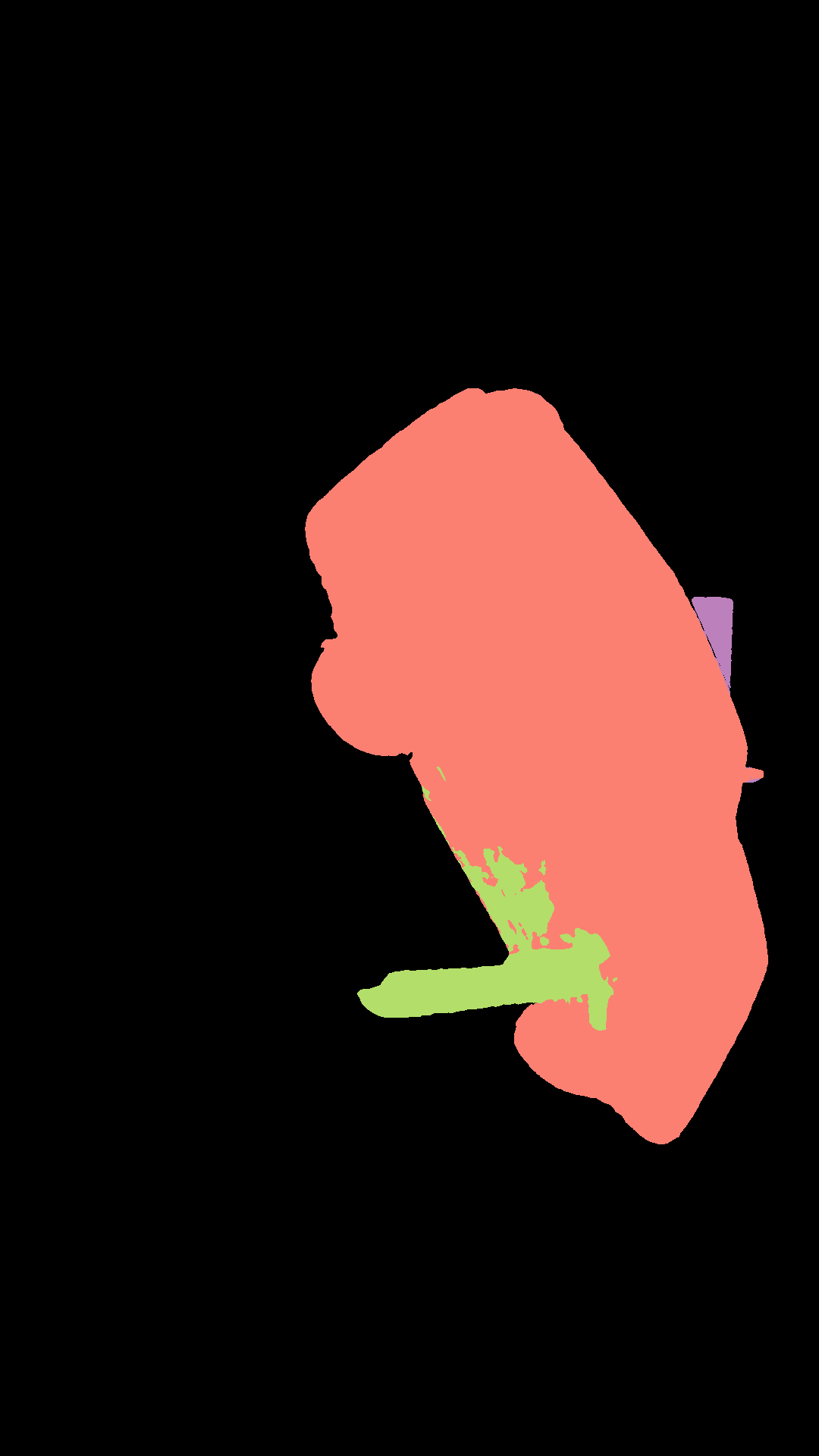}&
        \includegraphics[valign=c, width=0.1\columnwidth]{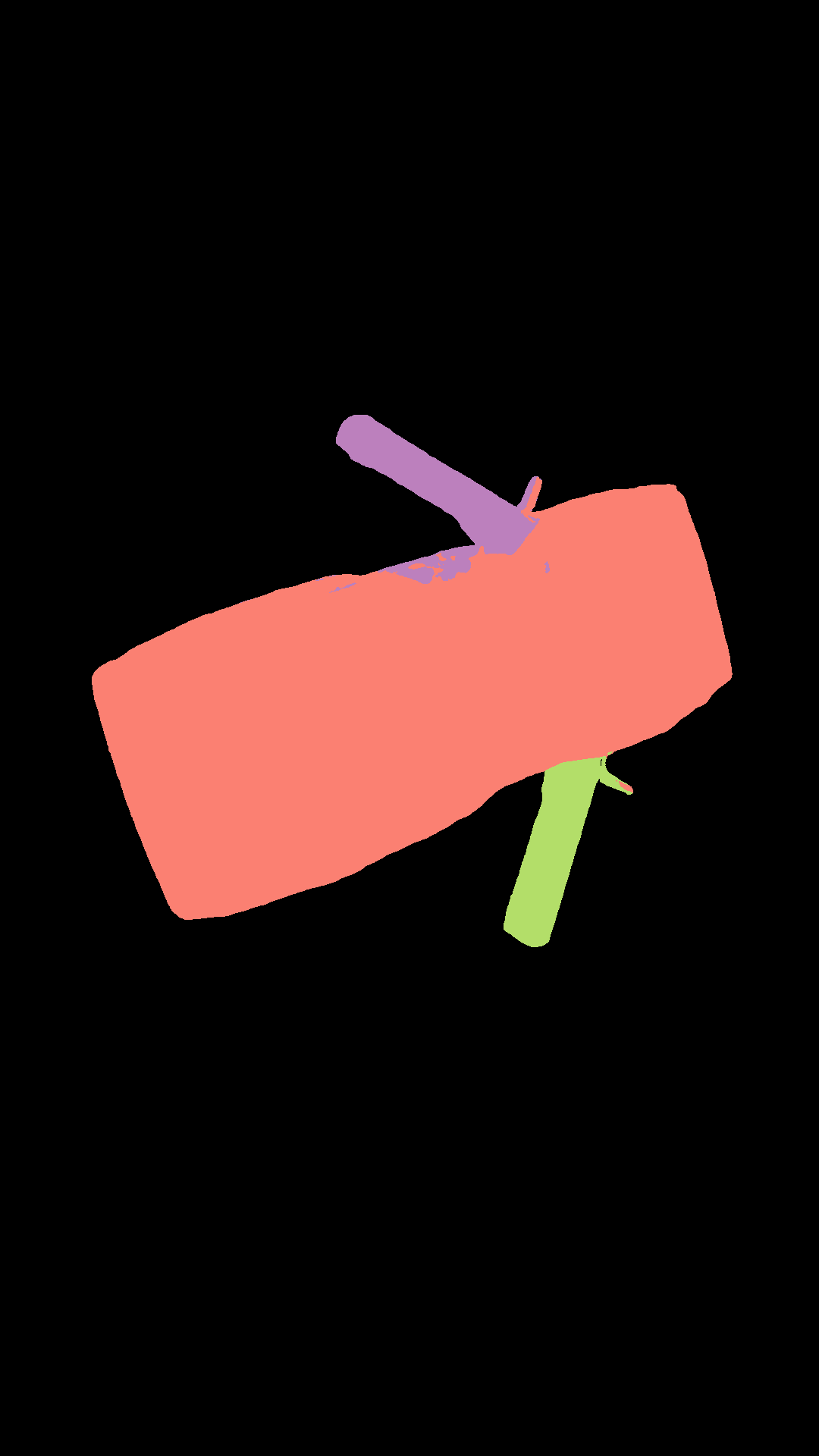}&
        \includegraphics[valign=c, width=0.1\columnwidth]{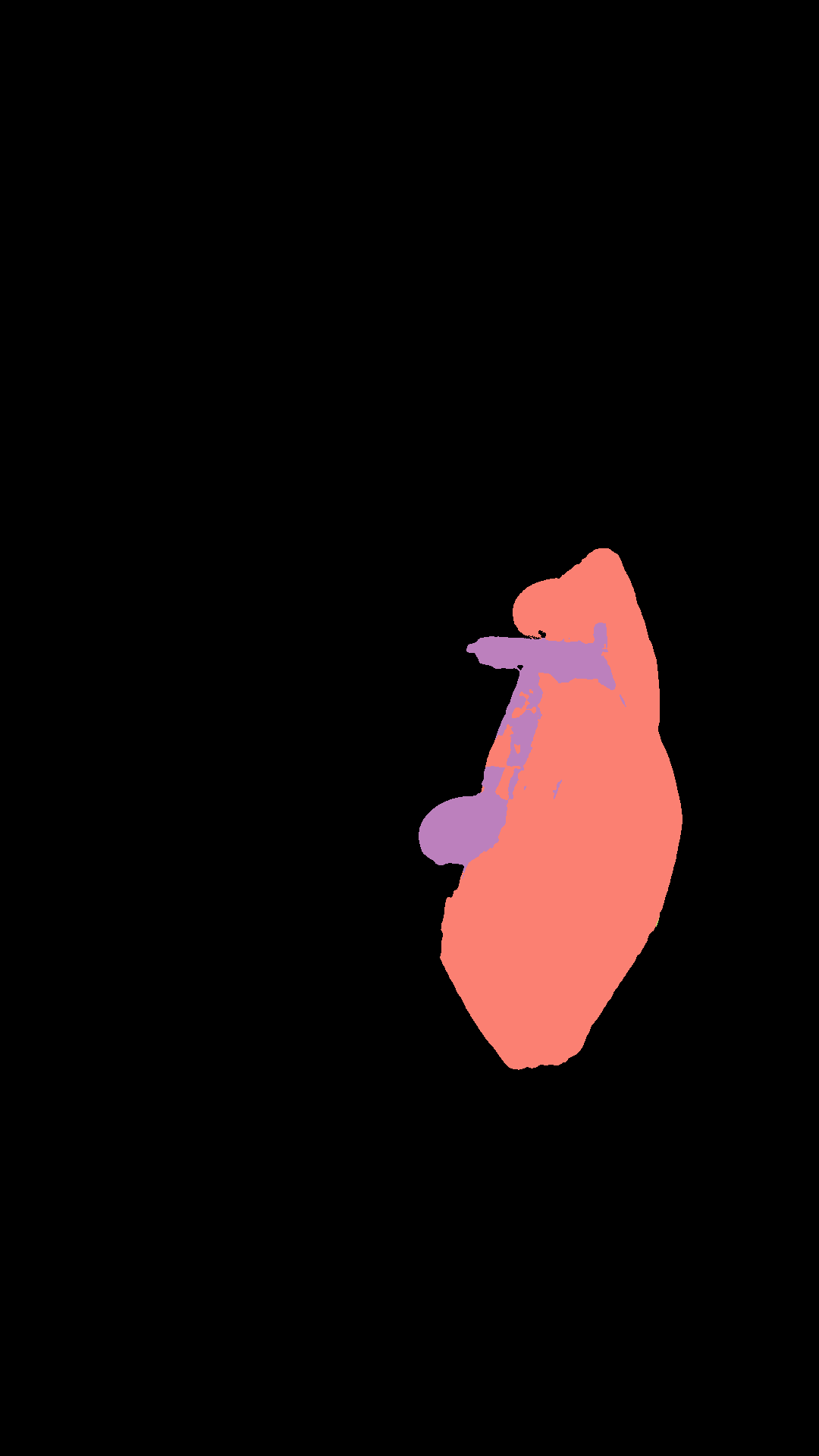}&
        \includegraphics[valign=c, width=0.1\columnwidth]{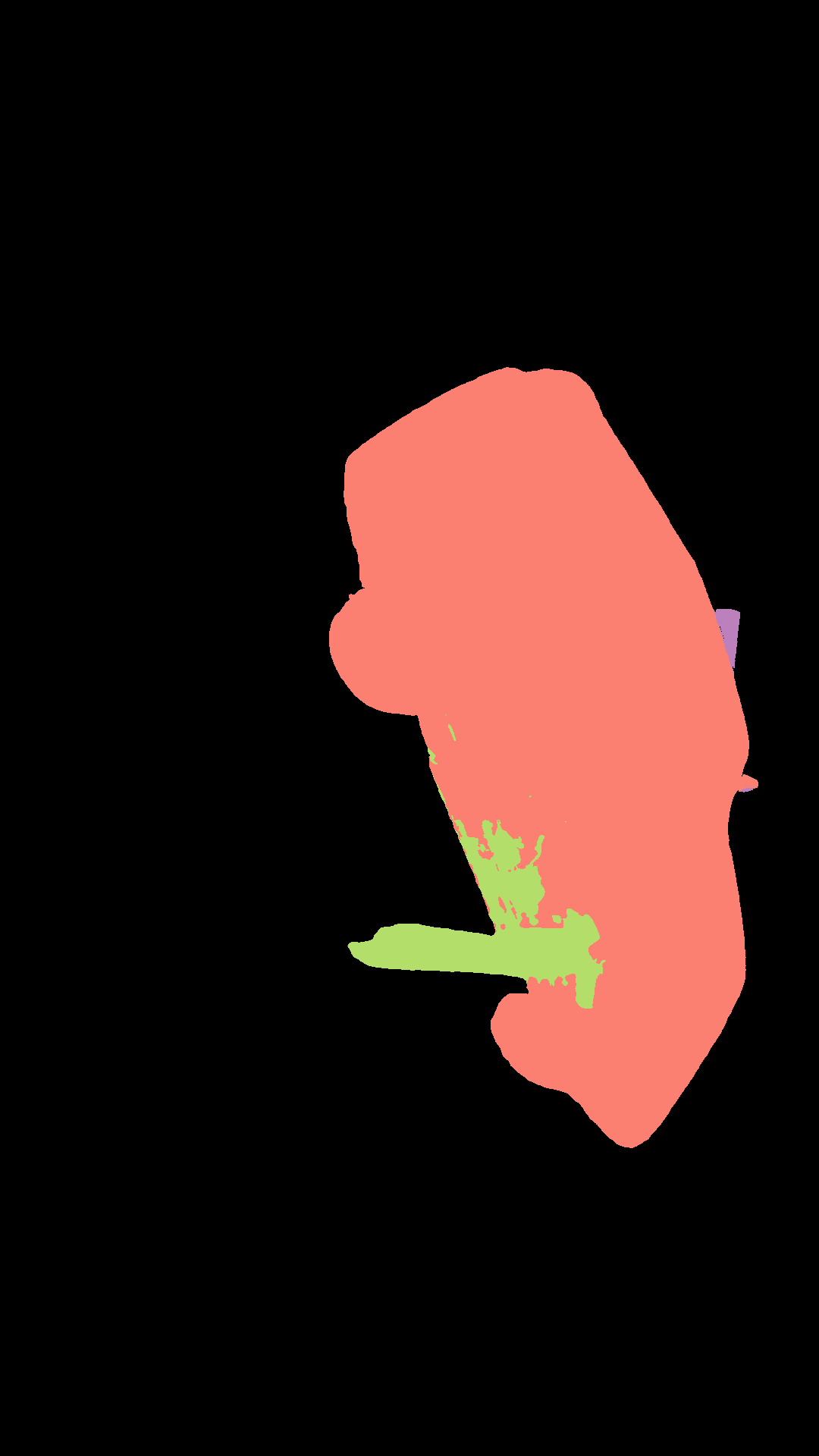}\\
        \\ \hline
    \end{tblr}
    }
    \caption{\textbf{More Qualitative evaluation on real-world object.}}
    \label{fig:appendix_real_vis}
\end{figure}